\documentclass[11pt, a4paper]{article}

\usepackage[margin=1in]{geometry} %
\usepackage[utf8]{inputenc}
\usepackage[english]{babel}
\usepackage{authblk} %
\usepackage{setspace}
\usepackage{amsmath,amssymb,amsthm}
\usepackage{tikz}
\usepackage{pgfplots}
\usepackage{amsfonts} %
\usetikzlibrary{calc}
\pgfplotsset{compat=1.17}
\usepgfplotslibrary{groupplots,fillbetween}
\usetikzlibrary{arrows.meta}
\pgfplotsset{compat=1.18}

\usepackage[numbers,sort&compress]{natbib} %
\usepackage{xcolor}

\newcommand{\bla}{\color{black}}

\usepackage{algorithm}
\usepackage{algpseudocode}
\usepackage{bbm}
\usepackage{subcaption}
\usepackage{graphicx}
\usepackage{booktabs}
\usepackage{xcolor}

\usepackage{natbib}
 \bibpunct[, ]{(}{)}{,}{a}{}{,}%
 \def\bibsep{\smallskipamount}%

\usepackage{algorithm}
\usepackage{algpseudocode}
\usepackage{bbm}
\usepackage{geometry}
\usepackage{subcaption}
\usepackage{graphicx}
\usepackage{booktabs}

\newcommand{\E}{{\rm I\!E}}

\def\EE{\mathbb{E}}
\def\E{\mathbb{E}}

\usepackage{hyperref}

\hypersetup{
	unicode=false,
	pdftoolbar=true,
	pdfmenubar=true,
	pdffitwindow=false,     %
	pdfstartview={FitH},    %
	pdfkeywords={}, %
	pdfnewwindow=true,      %
	colorlinks=true,       %
	linkcolor=black,          %
	citecolor=blue,        %
	filecolor=pink,      %
	urlcolor=cyan           %
}

\newcommand\independent{\protect\mathpalette{\protect\independenT}{\perp}}
\def\independenT#1#2{\mathrel{\rlap{$#1#2$}\mkern2mu{#1#2}}}

\def\given{\, | \,}

\newtheorem{theorem}{Theorem}
\newtheorem{proposition}{Proposition}
\newtheorem{lemma}{Lemma}

\newtheorem{assumption}{Assumption}
\newtheorem{corollary}{Corollary}
\newtheorem{example}{Example}

\newtheorem{definition}{Definition}

\def\begar{$$\begin{array}{lll}}
	\def\endar{\end{array}$$}
\def\begarlab{\begin{equation} \begin{array}{lll} \label}
		\def\endarlab{\end{array} \end{equation}}

\def\ds1{{\mathrm{1 \hspace{-2.6pt} I}}}

\def\calP{{\cal P}}

\def\calS{{\cal S}}

\def\calX{{\cal X}}

\begin{document}

\title{Beyond Demand Estimation: Consumer Surplus Evaluation via Cumulative Propensity Weights}

\author[1]{Zeyu Bian\thanks{Alphabetical Order.} \thanks{Email: \texttt{zeyu.bian@fsu.edu}}}
\author[2]{Max Biggs\thanks{Email: \texttt{biggsm@darden.virginia.edu}}}
\author[3]{Ruijiang Gao\thanks{Email: \texttt{ruijiang.gao@utdallas.edu}}}
\author[4]{Zhengling Qi\thanks{Email: \texttt{qizhengling@email.gwu.edu}}}

\affil[1]{Florida State University}
\affil[2]{University of Virginia}
\affil[3]{University of Texas at Dallas}
\affil[4]{George Washington University}

\date{} %

\maketitle

\begin{abstract}
This paper develops a practical framework for using observational data to audit the consumer surplus effects of AI-driven decisions, specifically in targeted pricing and algorithmic lending. Traditional approaches first estimate demand functions and then integrate to compute consumer surplus, but these methods can be challenging to implement in practice due to model misspecification in parametric demand forms and the large data requirements and slow convergence of flexible nonparametric or machine learning approaches. Instead, we exploit the randomness inherent in modern algorithmic pricing, arising from the need to balance exploration and exploitation, and introduce an estimator that avoids explicit estimation and numerical integration of the demand function.  
Each observed purchase outcome at a randomized price is an unbiased estimate of demand and by carefully reweighting purchase outcomes using novel cumulative propensity weights (CPW),  we are able to reconstruct the integral, or area under demand curve, when these outcomes are aggregated. 
Building on this idea, we introduce a doubly robust variant named the augmented cumulative propensity weighting (ACPW) estimator that only requires one of either the demand model or the historical pricing policy distribution to be correctly specified. Furthermore, this approach facilitates the use of flexible machine learning methods for estimating consumer surplus, since it achieves fast convergence rates by incorporating an estimate of demand, even when the machine learning estimate has slower convergence rates. Neither of these estimators is a standard application of off-policy evaluation techniques, since the target estimand, consumer surplus, is typically unobserved. To address algorithmic fairness, we extend this framework to an inequality-aware surplus measure, allowing regulators and firms to quantify the trade-off between firm profit and equity. Finally, we conduct a comprehensive numerical study to validate the theoretical properties of our proposed methods.
\end{abstract}

\section{Introduction}

With the proliferation of data on consumer behavior and the development of sophisticated Artificial Intelligence (AI), firms are increasingly adopting autonomous, targeted algorithmic strategies to price goods and offer personalized loans. For example, many e-commerce firms will dynamically adjust prices based on an individual's purchasing history, demographics, or even browsing behavior (e.g., \cite{hannak2014measuring}), while financial firms employ similar targeted algorithms to set interest rates for auto loans, mortgages, and credit cards. Indeed, the setting of interest rates is widely recognized as a form of targeted pricing, with recent studies utilizing auto loan datasets to empirically demonstrate the efficacy of algorithmic pricing strategies \citep{phillips2015effectiveness,ban2021personalized,elmachtoub2023balanced,miao2023personalized}. Although these prescriptive AI technologies promise to enhance market efficiency, boost firms' revenues, and broaden inclusion through personalized service, there are also concerns regarding their potential adverse impact on consumers.

A primary concern is that increased firm revenues due to targeted algorithms come at the expense of consumer surplus, with recent models from the academic literature suggesting that this burden can be unfairly distributed \citep{kallus2021fairness,cohen2022price}. In addition, there is a growing concern that marginalized groups might be disproportionately affected by these pricing practices, facing steeper prices and lower surplus. For example, ride-hailing services have been found to charge higher fares in neighborhoods with larger non-white populations and higher poverty levels, indicating that these communities face larger price hikes due to algorithmic pricing strategies \citep{pandey2021disparate}.
These concerns have drawn significant attention from policymakers and the media \citep{NYT2023}. For example, a White House report \citep{WhiteHouse2015BigData} highlighted the risks associated with algorithmic decision-making in pricing, the Federal Trade Commission (FTC) recently initiated investigations into potential discriminatory pricing practices by mandating transparency from companies employing personalized pricing schemes \citep{KLGates2024,sisco2024}, and lawmakers have called for scrutiny of major retailers' pricing practices \citep{warren2024}. 
Although there is significant interest, it is not yet clear what the impact of more targeted algorithmic pricing policies will be \citep{WhiteHouse2015BigData}.

These developments underscore the need for robust auditing tools capable of evaluating the impact of algorithmic pricing on consumer surplus and its distribution across customers. For regulators like the FTC, such tools can help ensure firms are abiding by relevant laws and regulations, such as the Gender Tax Repeal Act of 1995 (commonly known as the ``Pink Tax''), which prohibits gender-based pricing discrimination \citep{GenderTaxRepealAct1995}, or the Equal Credit Opportunity Act (ECOA) that prohibits creditors from discriminating against demographic factors. They can also help policymakers understand the impacts of various types of algorithmic pricing and who is adversely affected, leading to better guidelines and rules. Such tools can also help firms assess regulatory or reputational risks of existing algorithms, and simulate outcomes to ensure future compliance. They also provide the opportunity to strike a better balance between short-term profit and long-term customer relationships by ensuring consumer surplus remains at sustainable levels for all customer types.

Despite this need, there are issues regarding the efficacy of existing consumer surplus estimation methods. Typically, estimation techniques rely on first estimating demand and then integrating demand to calculate surplus (e.g., \cite{bhattacharya2024nonparametric}), so the accuracy of the surplus estimate is highly dependent on the accuracy of the demand estimate. This approach is inherently indirect and requires estimating demand as a potentially complex function of price, even when the surplus target can be a simple scalar. Classical demand models depend on parametric assumptions \citep{mcfadden1981econometric,SmallRosen1981, dagsvik2005compensating,shiller2013first} and can be biased when behavior deviates from the assumed form \citep{jagabathula2017nonparametric,bhattacharya2024nonparametric}. More flexible nonparametric and machine learning methods, such as neural networks \citep{farrell2020deep, farrell2021deep}, relax functional assumptions but often require very large datasets and careful tuning to achieve reliable accuracy, and typically have slower convergence rates. Given these shortcomings, it is useful to explore more direct alternatives for consumer surplus estimation that are less dependent on accurate demand estimation, and to investigate how to incorporate machine learning estimates while still achieving fast convergence rates.

One key feature of many successful modern pricing algorithms that can be leveraged for new methods of surplus estimation is that they involve a degree of price experimentation. In practice, market conditions, competitor actions, and consumer preferences evolve over time, so pricing algorithms must continuously explore the demand curve rather than solely exploit current prices to maintain optimality \citep{besbes2009dynamic,rana2014real}. Recent empirical work \citep{bray2024observational} has observed severe bias in the estimation of price sensitivity at a large supermarket chain when using relatively static historic pricing data compared to experimental data, while \cite{dube2023personalized} have demonstrated that prices set after a period of price experimentation drastically improved profitability at ZipRecruiter. These examples also indicate an increased willingness among firms to engage in experimentation, which has become more ubiquitous as e-commerce platforms enable continuous price updates. While primarily intended to support profit maximization, this inherent randomization also provides a rich source of quasi-experimental variation that can be leveraged to estimate the consumer surplus.

One natural approach to utilizing such data and potentially avoiding explicit demand estimation is to adapt inverse-propensity-score weighting (IPW) from causal inference \citep{rosenbaum1983central, beygelzimer2009offset}. However, the availability of quasi-experimental data alone does not solve the fundamental difficulty of measuring consumer welfare. Unlike revenue, which is entirely determined by the price and the observed purchase decision, consumer surplus relies on the customer's valuation, a variable that is inherently latent. In the observational data, we do not observe the surplus. Instead, we only observe a binary purchase decision, which serves as a coarse approximation indicating that the valuation exceeded the price. Because the true outcome of interest is never directly seen, standard causal inference techniques such as weighting observed outcomes cannot be directly applied. %

Motivated by these challenges, we propose the new cumulative propensity weight (CPW) estimator, which directly estimates consumer surplus and bypasses the need for explicit demand modeling. This approach leverages the randomness already present in the observational data due to the need to balance exploration and exploitation in many modern algorithmic pricing strategies.
Intuitively, each observed purchase outcome at a randomized price is an unbiased estimate of demand. By carefully reweighting that observation according to how often a target pricing policy would offer a price at or below the one observed for similar customers, relative to how frequently that price level appeared historically, we are able to reconstruct the consumer surplus, i.e., the integral or the area under the demand curve, when these weighted outcomes are aggregated. 
When the historical pricing policy is known or well documented, the approach is straightforward to implement and is entirely model-free. When it is unknown, the cumulative weights can be estimated from transaction data.

Building on the cumulative-weights idea, we introduce a doubly robust estimator with cross-fitting that delivers reliable surplus estimates even when either the demand model or the historical pricing distribution is misspecified. Specifically, when the historical pricing policy is known or can be estimated accurately, consistency holds even with a biased demand model. In contrast, when demand is asymptotically unbiased, consistency holds even with a misspecified historical pricing model. Furthermore, mirroring standard doubly robust estimators in causal inference, our estimator achieves efficiency under weak rate conditions without imposing the Donsker condition on either demand or cumulative-weight estimation. This flexibility enables the use of modern machine learning techniques and complex nonparametric models. We formally establish that our proposed estimators are asymptotically equivalent to the efficient influence function, and therefore achieve the lowest possible asymptotic variance. Finally, we prove the asymptotic normality of the proposed estimators, enabling the construction of valid confidence intervals, enabling policymakers to assess the reliability of these estimates.

To address fairness, we extend our objective from estimating the standard consumer surplus to estimating an inequality-aware surplus. This new target captures both the magnitude of surplus a policy creates and its distribution across different customer types via a single pre-specified parameter. At its baseline, this parameter replicates the standard arithmetic average of customers' surplus, but as it decreases, it places progressively greater weight on outcomes for worse-off groups. This approach is based on generalized mean aggregates as studied by \cite{bergson1954concept} and \cite{atkinson1970measurement}. To achieve this objective, we derive the efficient influence function of this new target and adapt our cumulative weights approach to construct an efficient estimator. Due to the nonlinearity of the objective, this estimator does not inherit full double robustness. Instead, it is singly robust with respect to the demand model, in that consistency holds as long as the demand is correctly specified, even if the cumulative weights are misspecified. This alters the theoretical requirements for establishing the asymptotic properties compared to the previous case: while incorporating cumulative weights still has the benefit of relaxed rate conditions, i.e., allowing flexible machine learning models, demand estimation requires stricter control, specifically satisfying standard nonparametric rates. Nevertheless, we derive valid asymptotic properties for this new estimator and show it has the minimal asymptotic variance among all regular estimators, which enables auditors to report valid confidence intervals for both aggregate and equity-sensitive surplus.

Our experiments confirm the reliability and robustness of the proposed framework. For the standard aggregate surplus, our doubly robust estimator accurately recovers the true surplus even when either the demand or pricing model is misspecified, achieving the best convergence rate when both are well specified. We further evaluate our inequality-aware surplus, studying its estimation error and confidence interval coverage. Despite the theoretical shift to single robustness for this nonlinear objective, our results demonstrate that the estimator remains highly effective, producing accurate point estimates and valid confidence intervals across varying equity settings. %

To further demonstrate the power of this framework, we apply it to a large-scale financial dataset of U.S. automobile loans. The global automotive market size was estimated at USD 2.75 trillion in 2025 and is projected to reach USD 3.26 trillion by 2030 \citep{mordor2025automotive}. It is an important driver of household financial stability and is increasingly dominated by algorithmic underwriting. We compare the welfare outcomes of a historical pricing policy against a trained AI pricing agent. Our analysis reveals an aggregate consumer surplus–equity tradeoff: personalized pricing reduces total consumer surplus while narrowing disparities across credit and political groups, demonstrating the framework’s value for auditing and regulatory evaluation.

\section{Related Work}
\label{sec:related_work}

\subsection{Consumer Surplus Estimation}

There is a substantial literature on consumer surplus evaluation in both discrete choice environments, where consumers typically choose one alternative from a set (e.g., \cite{mcfadden1972conditional,small1981applied, bhattacharya2015nonparametric}), and continuous demand settings (e.g., \cite{hausman1981exact, vartia1983efficient, HausmanNewey1995,hausman2017nonparametric}), such as gasoline purchases \citep{poterba2017gasoline}. Our work is more closely aligned with discrete choice models, but we focus on a single-item setting in which we observe individual customer characteristics and a binary outcome indicating whether they purchase the item. Initially, parametric models of demand were used for surplus estimation \citep{mcfadden1981econometric,herriges1999nonlinear, dagsvik2005compensating}, for example, the widely used logsum formula \citep{small1981applied}. This relies on strong assumptions about preferences and customer heterogeneity, such as additive extreme-value error distributions, which may lead to misleading welfare conclusions if the model is misspecified. 

To overcome such restrictive assumptions and make inferences in a broader range of settings, many semiparametric and nonparametric methods for estimating demand have been developed \citep{HausmanNewey1995, matzkin2016independence, berry2021foundations, bhattacharya2015nonparametric}. These typically involve flexible nonparametric regression to fit the demand function directly, followed by an integration to calculate the surplus. For example, in the continuous setting, \citet{HausmanNewey1995} fit demand using series and kernel estimators, and use it to solve a differential equation based on Shephard's Lemma. Subsequent research has focused on incorporating unobserved consumer heterogeneity into these models \citep{hausman2016individual, lewbel2017unobserved}. Similarly, in the discrete choice setting, \cite{bhattacharya2015nonparametric, bhattacharya2018empirical} estimates a conditional nonparametric probability of purchase for each item (demand in this setting), then integrates to estimate consumer surplus. Recent machine learning approaches have used neural networks to flexibly estimate demand and surplus with minimal functional-form assumptions \citep{farrell2020deep, farrell2021deep}, including in discrete choice models \citep{aouad2025representing}. Tree ensemble methods have also been proposed \citep{chen2019use, chen2022decision}. A comprehensive account of nonparametric consumer surplus estimation can be found in \cite{hausman2017nonparametric} and \cite{bhattacharya2024nonparametric}.

A critical trade-off in these nonparametric demand estimation methods is that convergence can be slower, leading to worse finite-sample performance than in well-specified parametric models. In either case, the consumer surplus estimates are only as accurate as the demand model. In contrast, we present approaches that do not require modeling demand at all, and show that we can achieve faster convergence when we have a slowly converging demand model.

Some recent applications of consumer surplus estimation can be found, for example, in \cite{shiller2013first}, who use an ordered probit to analyze Netflix data.  \cite{dube2023personalized} employ a Bayesian parametric framework to estimate surplus from a large-scale randomized price experiment at ZipRecruiter. Other research has leveraged quasi-experimental variation in prices. For instance, \cite{cohen2016using} uses a regression discontinuity design to estimate price elasticities and surplus from Uber's surge pricing data.  

Much of this literature is focused on incorporating income effects, which is not the focus of our paper. We focus on settings where the expenditure represents a small fraction of the consumer's total budget. In such regimes, the income effects are negligible, and the Marshallian consumer surplus provides a near-exact approximation of the Hicksian compensating variation \citep{willig1976consumer}. Alternatively one can interpret our analysis as applying when consumer utilities are quasilinear.

\subsection{Causal Inference and Off-Policy Evaluation}

Much of the recent work on auditing pricing algorithms can be viewed through the lens of off-policy evaluation (OPE). The OPE literature initially centered on the inverse propensity weighting (IPW) framework: each observed outcome is weighted by the reciprocal of its treatment (in our setting, price) assignment probability, or propensity score, yielding an unbiased estimate of the counterfactual reward when the propensity model is correct \citep{rosenbaum1983central,beygelzimer2009offset}.  A complementary line of research advocates the direct method (DM), which replaces missing counterfactuals with fitted values from an outcome model. When that model is correctly specified, the DM can be more efficient than IPW \citep{qian2011performance,shalit2017estimating}. Recognizing that either component may be misspecified in practice, the doubly‑robust estimator \citep{robins1994estimation,dudik2011doubly,zhou2023offline} blends the two ideas and remains consistent so long as either the propensity model or the outcome model is estimated without systematic error.  Empirical evidence suggests that, when at least one nuisance model is reasonably accurate, the doubly robust estimator achieves lower mean‑squared error than either IPW or the DM on their own \citep{dudik2014doubly}. The idea of the doubly robust method originates from the missing data literature \citep{ robins1994estimation, tsiatis2006semiparametric} and has been widely adopted in causal inference \citep{ bang2005doubly,hernan2010causal, chernozhukov2018double,kennedy2024semiparametric} and policy learning/evaluation \citep{robins2004optimal,wallace2015doubly,shi2018high,kallus2020double,liao2022batch,bian2023variable,zhou2023offline}. 

In the management science community, there is a growing body of literature that leverages off-policy learning techniques to estimate revenue in pricing settings. This stream of research is often grounded in the ``predictive to prescriptive'' analytics framework \citep{bertsimas2020predictive}, which formally integrates machine learning predictions with optimization models to derive decision policies from observational data. Complementing this methodological foundation, recent studies have developed rigorous statistical learning frameworks for personalized revenue management \citep{chen2022statistical,kallus2018policy} and demonstrated the practical efficacy of these data-driven pricing algorithms through large-scale field experiments \citep{ferreira2016analytics}. Specifically, researchers have used OPE techniques to address pricing settings characterized by binary demand \citep{biggs2021loss, biggs2022convex, elmachtoub2023balanced}, as well as censored demand resulting from inventory shortages \citep{ban2020confidence, bu2022offline, tang2025offline}, and  unobserved confounding \citep{kallus2021minimax,miao2023personalized}. In contrast to these works, our primary objective is the estimation of consumer surplus rather than revenue. This shift substantially alters the estimation problem: unlike the standard OPE setting, where the outcome is directly observed, surplus relies on consumer valuations that are never seen. In our context, we observe only a binary purchase decision, a coarse proxy indicating whether valuation exceeds price rather than the continuous valuation itself. Because the true outcome variable is latent, conventional causal inference techniques cannot be directly applied, necessitating the development of the novel methodological tools presented here.

\subsection{Organization}
The remainder of this paper is organized as follows. Section \ref{sec:problem} formalizes the consumer surplus estimation problem. Section \ref{sec:cpw} introduces the Cumulative Propensity Weighting (CPW) estimator, a novel approach that leverages the randomness in algorithmic pricing to estimate surplus without explicit demand integration. Section \ref{sec:acpw} develops the Augmented CPW (ACPW) estimator, establishing its double robustness.  Section \ref{sec:iaacpw} extends the framework to inequality-aware surplus measures, introducing a parameter to trade off aggregate surplus against equity. Section \ref{sec: theory} provides the theoretical analysis, proving the asymptotic normality and efficiency of the proposed estimators. Section \ref{sec:experiments} presents numerical experiments validating the method's robustness and demonstrates its application to a large-scale U.S. auto loan dataset. Finally, Section \ref{sec:conclusions} concludes with managerial implications. The appendices contain proofs of all theorems and auxiliary lemmas. Lastly, we provide an extension on partial identification bounds for settings where the overlap assumption is violated in Appendix \ref{app:partial_identification}.

\section{Problem Formulation}\label{sec:problem}
Consider a population of heterogeneous consumers with features $X \in \mathcal X$, and valuations (i.e., willingness to pay), $V\in \mathbb{R}^+$, interested in purchasing at most one unit of an item. For a fixed price $p$, the average consumer surplus can be defined as the average excess of each consumer's valuation over the price they pay,
\begin{align*}
    \calS(p) = \EE\left[\left(V - p\right)_+\right],
\end{align*}
where $x_+=\max(x,0)$. Alternatively, in a lending scenario, we can consider $p$ as a periodic interest payment and interpret $V$ as the maximum interest the customer is willing to pay, with the surplus being the positive difference between them. Note, we may be interested in assessing the conditional surplus $\calS(p|X) = \EE_V[\left(V - p\right)_+ \given X]$ associated with a particular group of interest for comparison purposes, or population surplus $\calS(p) = \EE_X\EE_V[\left(V - p\right)_+ \given X]$, which is the surplus over all customers. For notational simplicity, we focus on the latter, but our results also hold for the former, unless otherwise noted. We highlight that this definition is consistent with traditional ``area under the demand curve" calculation of consumer surplus \citep{bhattacharya2024nonparametric}, where the demand curve is defined as the probability of purchase, as highlighted in Section \ref{sec:baseline}. 
We do not incorporate income effects into the model and focus on goods where expenditure represents a small fraction of the consumer's total budget. In this case, the income effects are negligible \citep{willig1976consumer}.  Alternatively, but resulting in the same framework, we focus on customers with quasilinear utility.

In general, firms are interested in offering and evaluating pricing policies that can be both targeted and stochastic, where the price offered to the customer with feature $X$ is associated with a pricing policy $\pi: \calX \rightarrow \Delta(\calP)$, which is a conditional probability mass/density over the price space $\calP$ given the feature $X \in \calX$. %
Then the average consumer surplus under the pricing policy $\pi$ is defined as 
\begin{align}\label{def: consumer surplus for random p}
    \calS(\pi) = \EE\left[\int_\calP \pi(p \given X)\left(V - p\right)_+ \text{d}p\right],
\end{align}
where the underlying expectation is taken with respect to the joint distribution of $(X, V)$.  If $V$ is observed, then one can estimate $\calS(\pi)$ by directly using the sample average to approximate the expectation in Equation \eqref{def: consumer surplus for random p}. However, in practice, consumers' valuations $V$ are typically unobserved, which presents a challenge for the estimation task. While a consumer's valuation $V$ is often unobserved, typically their binary purchase decision $Y$ is often recorded at the price they were offered $P \in \calP$. This purchase decision is determined by whether their valuation exceeds the offered price $P$:
\begin{align}\label{model: true model}
Y &= \mathbb{I}(V > P).
\end{align}
where $\mathbb{I}(\cdot)$ denotes the indicator function. Here, $Y=1$ denotes a purchase (the condition is met), and $Y=0$ denotes no purchase.
In addition to $V$ being unobserved, we often face a distribution shift, where we may want to evaluate surplus under a pricing policy that is different from the historical policy that generated the data. In general, we consider three objectives: (i) evaluating consumer surplus of a new pricing strategy (also referred as the target policy) $\pi$; (ii) evaluating the consumer surplus of a current or previously used pricing strategy $\pi_D$ from the historical data (also referred as the behavioral policy); and (iii) evaluating the change in surplus between historical and new policies: 
\begin{align}\label{def: difference in consumer surplus}
    \Delta(\pi) = \EE\left[\int_\calP \left(\pi(p \given X)- \pi_D(p \given X)\right)\left(V - p\right)_+ \text{d}p\right],
\end{align}
We note that the historical pricing distribution $\pi_D$ may be known or unknown, while $\pi$ is always known. When the firm is engaged in algorithmic pricing, often the historical pricing policy is encoded digitally and is therefore known. When it is unknown, it can typically be estimated from the data. Generally, the difference between evaluating (i) and (ii) arises from unknown $\pi_D$, which can introduce additional challenges in the estimation. We will focus on this case when evaluating (ii) unless otherwise noted. The offline dataset can thus be represented as ${(X_i, P_i, Y_i)}_{i=1}^n$, consisting of i.i.d. samples of $(X, P, Y)$ generated under a  historical pricing policy $\pi_D$.

Next, we present some conditions necessary for the identification of the average consumer surplus $\calS(\pi)$ under the observational data-generating distribution. First, we provide the formal definition of identification.
\begin{definition}[Identifiability]  \label{def:partial}
    A parameter of interest $\theta$ in a probabilistic model $\{\mathcal{P}_\theta: \theta \in \Theta \}$ is said to be identifiable if the mapping $\theta \mapsto \mathcal{P}_\theta$ is injective, i.e.,  $\mathcal{P}_{\theta_1} = \mathcal{P}_{\theta_2} \implies \theta_1 = \theta_2$. 
\end{definition}

For identification in this setting, we require two conditions to hold: 
\begin{assumption}\label{assumption:ignorability}
   (Ignorability) $P \independent V |X $.
\end{assumption}
\begin{assumption}\label{assumption:overlap}
    (Overlap) The price data generating distribution satisfies $\pi_D(p \given x)>0$, for all $p \in \calP$, and every $x\in \calX$. In addition, the support $\calP$ contains the support of the valuation $V$. For identifying $\Delta(\pi)$, we only require the support of $\pi_D$ to contain the support of $\pi$.
\end{assumption}

Assumption \ref{assumption:ignorability} is similar to the classical causal assumptions of ignorability or exchangeability, see, e.g., \citet{rosenbaum1983central,hernan2010causal}. It states that, conditional on consumer characteristics $X$, the distribution of valuations is unaffected by the offered price, facilitating the identification of surplus. It is commonly satisfied as long as the factors that drove the historical pricing decisions are recorded and available in the observed data. It is worth noting that Assumption \ref{assumption:ignorability} does not impose any parametric structure on the consumer valuations model. %
Assumption \ref{assumption:overlap} requires that every possible price $p \in \calP$ has a positive probability of being observed for the observational data. This means the previous pricing policy must involve some degree of randomization, which, as previously discussed, is necessary for a pricing policy to obtain and maintain optimality. Such a condition is similar to the positivity condition in causal inference for identifying the average treatment effect. Without this coverage assumption, nonparametric identification of the demand function, and consequently, the absolute consumer surplus, is impossible without relying on strong extrapolation assumptions. However, we note that the requirements are significantly relaxed when evaluating the difference in surplus between two policies. In that context, we only require overlap over the prices proposed by the new policy $\pi$, rather than the entire price space. While identifying absolute surplus requires observing demand at extreme prices (to capture total willingness to pay), estimating policy differences is often sufficient for decision-making and aligns with standard practices in the literature \citep{bhattacharya2024nonparametric}. While we establish identification and estimation results for both quantities, we acknowledge the difference is easier to practically implement.

To address settings where Assumption \ref{assumption:overlap} does not hold, Appendix \ref{app:partial_identification} introduces an extension that exploits demand function properties (specifically monotonicity and log-concavity) to bound the surplus. Experimental results in Appendix \ref{app:exp_partial} confirm the validity of these partial identification bounds and illustrate their superior tightness compared to naive bounds derived from the natural [0, 1] support of purchase probability.

Next, we present a commonly used baseline approach for estimating $\calS(\pi)$ and discuss its limitations. Without loss of generality, we assume $\mathcal P = [0, \infty)$.

\subsection{Baseline Solution: Direct Method} \label{sec:baseline}

A classic approach to calculate the consumer surplus is to calculate the area under the demand curve above a particular price (for example, \cite{bhattacharya2015nonparametric}). Under Assumption \ref{assumption:ignorability} we show this form is equivalent to our consumer surplus definition (\ref{def: consumer surplus for random p}) for the stochastic pricing policy setting 
    \begin{align}\label{eqn: identification}
        &\calS(\pi) = \EE\left[\int_0^\infty\pi(p \given X) \int_{z=p}^\infty \mu(X,z)dzdp \right],
    \end{align}  where $\mu(x,z) \equiv \E[Y \given X=x, P = z] \equiv\mathbb{P}[V > z \given X=x]$ is the probability of purchase, which can be considered the demand function in this setting.
A brief proof showing the equivalence is provided in Proposition \ref{prop:DM_equiv_valuation_diff} in Appendix \ref{sec:DM_equiv_valuation_diff} and follows from carefully changing the order of integration. While Assumption \ref{assumption:overlap} is not strictly required for the derivation of this identity, it is necessary for the identification and estimation of $\mu(x,z)$ over the integration range. 
 This equation shows that even when the valuation $V$ is unobserved, the surplus can still be identified by first computing the integral of the demand function $\mu(x,z)$ over prices above $p$, and then taking a weighted average over price using the target policy $\pi$. When surplus under the behavior policy $\pi_D$, or the difference in surplus is of interest, it simplifies to \begin{gather} \label{eqn: identification behavioral}
    \calS(\pi_D) = \EE_{X, P\sim \pi_D}\left[\int_{P}^{\infty} 
     \mu(X,z)dz\right], ~~ \Delta(\pi) =  \calS(\pi) - \calS(\pi_D).
\end{gather}

In practice, the demand function $\mu(x, p)$ is generally unknown and must be estimated by regressing $Y$ on $X$ and $P$, yielding an estimator $\widehat \mu(x, p)$.
Then the direct method (DM) uses the sample average to approximate Equations \eqref{eqn: identification} and \eqref{eqn: identification behavioral}
and gives
\begin{gather}
\label{eqn:dm_estimator}
   \widehat \calS_{DM}(\pi)= \frac{1}{n}\sum_{i=1}^n \int_0^\infty\pi(p | X_i) \int_{p}^\infty \widehat \mu(X_i, z)dzdp, ~~ \widehat \calS_{DM}(\pi_D)= \frac{1}{n}\sum_{i=1}^n  \int_{P_i}^\infty \widehat \mu(X_i, z)dz, \\
   \widehat \Delta_{DM}(\pi) =  \widehat \calS_{DM}(\pi) - \widehat \calS_{DM}(\pi_D), \end{gather} where $\widehat \mu(\cdot, \cdot)$ is the estimator of the demand function. 
We note that the historic surplus estimator $\widehat{\mathcal{S}}_{DM}(\pi_D)$ has the advantage of not requiring an estimate of the historic pricing policy distribution $\widehat{\pi}_D$ (when it is unknown), compared to the naive approach of substituting $\widehat{\pi}_D$ into $\widehat{\mathcal{S}}_{DM}(\pi)$, highlighting the suitability of each estimator for its particular task.

As discussed previously, the performance of the DM estimator depends critically on the accuracy of the outcome model $\widehat \mu(x, p)$, which can be challenging to estimate in practice. Traditional demand estimators that impose fixed parametric forms can yield biased results whenever actual purchasing behavior strays from those assumptions. In contrast, modern fully non‑parametric techniques, such as neural‑network demand models \citep{farrell2020deep, farrell2021deep}, avoid functional‑form misspecification but generally require a very large sample before they deliver reliable accuracy, limiting their practicality in many observational settings. Furthermore, the DM relies on numerical integration of estimated functions. This step requires the underlying estimation to be uniformly accurate across $\mathcal{P}$. Lastly, the numerical integration procedure itself can introduce bias and raise computational burden since the integration must occur over all prices for every datapoint.

These limitations associated with the DM estimator motivate the exploration of alternative approaches. In the next section, we present our newly proposed solutions for identifying and estimating $\mathcal{S}(\pi)$.

\section{Cumulative Propensity Weights Representation and Estimation}
\label{sec:cpw}

We next present an alternative approach for consumer surplus estimation that avoids the challenges of estimating a demand function and numerical integration. This estimator leverages the price variation already present in modern algorithmic pricing due to the need to balance exploration and exploitation. Rather than explicitly integrating to get the area under an estimated demand curve, the estimator approximates this area by an aggregate of weighted purchase outcomes, each of which is an unbiased estimate of demand at a given price. By carefully weighing the observations, we can recover the consumer surplus in expectation.  Formally, our estimator is motivated by the following alternative identification result.

\begin{theorem}\label{thm: alternative identification result}
    Under Assumptions 1 and 2,  we have
    \begin{align}\label{eqn: identification CPW}
        \calS(\pi) =\EE\left[\frac{F^\pi(P \given X)}{\pi_D(P \given X)}Y\right],
    \end{align} where $F^\pi(p|x)$ denotes the cumulative distribution function under the target policy, i.e.,  
    $F^\pi(p|x) \equiv \int_0^p \pi(u|x) du$.
\end{theorem}

\begin{proof} The proof follows from the law of iterated expectations and a change of the order of integration:
\begin{align*}
& \underbrace{\EE\!\left[ \int_{p=0}^\infty \pi(p\mid X) \int_{z=p}^\infty \mu(X,z) \, dz \, dp \right]}_{\mbox{Equation \eqref{eqn: identification}} }
= \EE\!\left[ \int_{p=0}^\infty \int_{z=p}^\infty \pi(p\mid X)\,\mu(X,z) \, dz \, dp \right] \\
= & \; \EE\!\left[ \int_{z=0}^\infty \int_{p=0}^z \pi(p\mid X)\,\mu(X,z) \, dp \, dz \right] = \EE\!\left[ \int_{z=0}^\infty \Big(\int_{p=0}^z \pi(p\mid X)\,dp\Big)\,\mu(X,z) \, dz \right] \\
= & \;\EE\!\left[ \int_{z=0}^\infty \frac{F^\pi(z\mid X)}{\pi_D(z\mid X)}\,\mu(X,z) \,\pi_D(z\mid X)\,dz \right] = \EE\!\left[ \frac{F^\pi(P\mid X)}{\pi_D(P\mid X)}\,Y \right].
\end{align*}
\end{proof}

Analogously, the historical consumer surplus and change in surplus can be identified by replacing $\pi$ with $\pi_D$ in Equation \eqref{eqn: identification CPW} and taking the difference:

\begin{equation}
     \calS(\pi_D) =\EE\left[\frac{F^{\pi_D}(P \given X)}{\pi_D(P \given X)}Y\right], \mbox{ and  } \Delta(\pi_D) =\EE\left[\frac{\left(F^{\pi}(P \given X)-F^{\pi_D}(P \given X)\right)}{\pi_D(P \given X)}Y\right].
\end{equation}

Based on Theorem \ref{thm: alternative identification result}, our proposed CPW estimators can be derived by taking the sample average of Equation \eqref{eqn: identification CPW}: 
\begin{align}
\label{eqn:ips_estimator}
   \widehat \calS_{CPW}(\pi)=& \frac{1}{n}\sum_{i=1}^n \frac{F^\pi(P_i | X_i)}{\widehat \pi_D(P_i | X_i)}Y_i,~~     
   \widehat \calS_{CPW}(\pi_D)= \frac{1}{n}\sum_{i=1}^n \frac{\widehat  F^{\pi_D}(P_i | X_i)}{\widehat \pi_D(P_i | X_i)}Y_i,\\
    &\widehat \Delta_{CPW}(\pi)= \frac{1}{n}\sum_{i=1}^n \frac{(F^\pi(P_i | X_i)-\widehat  F^{\pi_D}(P_i | X_i))}{\widehat \pi_D(P_i | X_i)}Y_i,
\end{align} 
where $\widehat \pi_D$ and $\widehat  F^{\pi_D}$ are the estimators for the density and cumulative density for the historic policy, respectively.

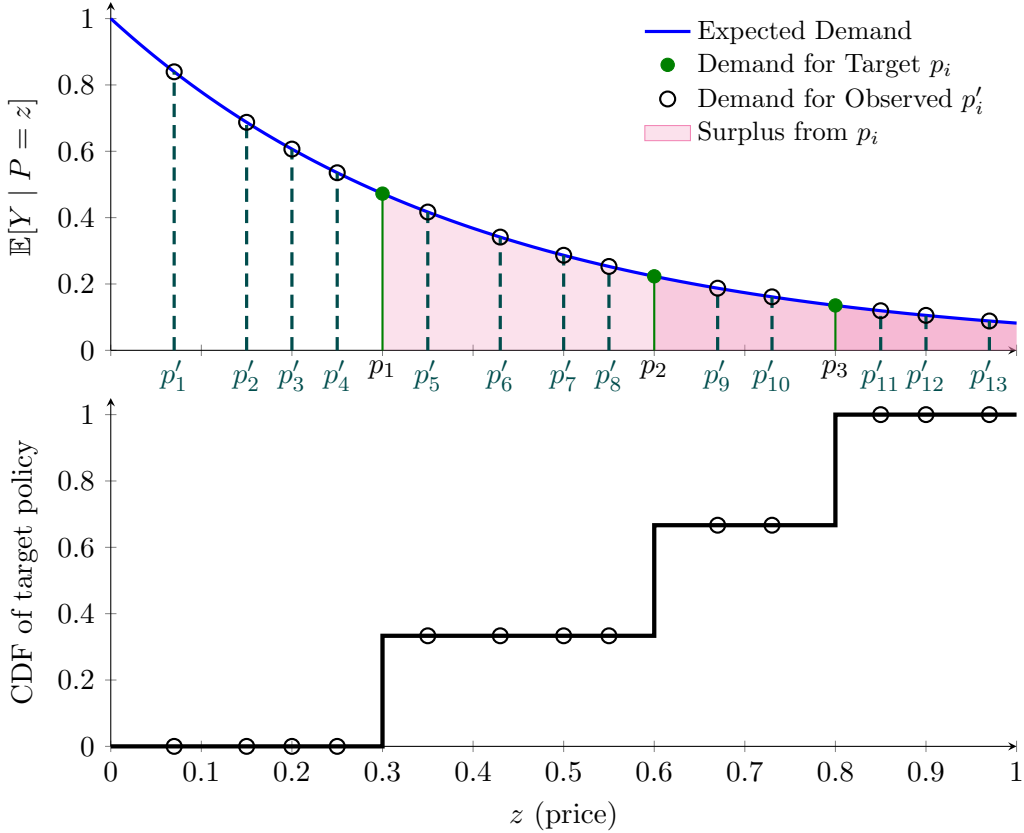
\begin{figure}
\centering
\begin{tikzpicture}

\def\pA{0.3}   %
\def\pB{0.6}   %
\def\pC{0.8}   %

\def\sA{0.07}  %
\def\sB{0.15}  %
\def\sC{0.20}  %
\def\sD{0.25}  %
\def\sE{0.35}  %
\def\sF{0.43}  %
\def\sG{0.50}  %
\def\sH{0.55}  %
\def\sI{0.67}  %
\def\sJ{0.73}  %
\def\sK{0.85}  %
\def\sL{0.90}  %
\def\sM{0.97}  %

\def\demand{exp(-2.5*x)}

\begin{groupplot}[
  group style={group size=1 by 2, vertical sep=18pt},
  width=13.5cm,
  height=6.2cm,
  xmin=0, xmax=1,
  domain=0:1,
  samples=400,
  clip=false,
  tick label style={/pgf/number format/fixed}
]

\nextgroupplot[
  axis lines=left,
  ymin=0, ymax=1.05,
  ylabel={$\mathbb{E}[Y\mid P=z]$},
  xticklabels=\empty,
  legend style={draw=none,at={(0.98,0.98)},anchor=north east,font=\small},
  legend cell align=left,
  ytick distance=0.2,
]

\addplot[name path=demand, blue, very thick] {\demand};
\addlegendentry{Expected Demand}

\path[name path=xaxis] (axis cs:0,0) -- (axis cs:1,0);

\addplot[green!50!black,thick,forget plot]
  coordinates {(\pA,0) (\pA,{exp(-2.5*\pA)})};
\addplot[green!50!black,thick,forget plot]
  coordinates {(\pB,0) (\pB,{exp(-2.5*\pB)})};
\addplot[green!50!black,thick,forget plot]
  coordinates {(\pC,0) (\pC,{exp(-2.5*\pC)})};

\addplot[only marks,mark=*,mark size=2.5pt,green!50!black]
  coordinates {(\pA,{exp(-2.5*\pA)})
              (\pB,{exp(-2.5*\pB)})
              (\pC,{exp(-2.5*\pC)})};
\addlegendentry{Demand for Target $p_i$}

\node[below] at (axis cs:\pA,0) {$p_1$};
\node[below] at (axis cs:\pB,0) {$p_2$};
\node[below] at (axis cs:\pC,0) {$p_3$};

\addplot[teal!60!black,very thick,dash pattern=on 5pt off 3pt,forget plot]
  coordinates {(\sA,0) (\sA,{exp(-2.5*\sA)})};
\addplot[teal!60!black,very thick,dash pattern=on 5pt off 3pt,forget plot]
  coordinates {(\sB,0) (\sB,{exp(-2.5*\sB)})};
\addplot[teal!60!black,very thick,dash pattern=on 5pt off 3pt,forget plot]
  coordinates {(\sC,0) (\sC,{exp(-2.5*\sC)})};
  \addplot[teal!60!black,very thick,dash pattern=on 5pt off 3pt,forget plot]
  coordinates {(\sD,0) (\sD,{exp(-2.5*\sD)})};
  \addplot[teal!60!black,very thick,dash pattern=on 5pt off 3pt,forget plot]
  coordinates {(\sE,0) (\sE,{exp(-2.5*\sE)})};
  \addplot[teal!60!black,very thick,dash pattern=on 5pt off 3pt,forget plot]
  coordinates {(\sF,0) (\sF,{exp(-2.5*\sF)})};
  \addplot[teal!60!black,very thick,dash pattern=on 5pt off 3pt,forget plot]
  coordinates {(\sG,0) (\sG,{exp(-2.5*\sG)})};
    \addplot[teal!60!black,very thick,dash pattern=on 5pt off 3pt,forget plot]
  coordinates {(\sH,0) (\sH,{exp(-2.5*\sH)})};
  \addplot[teal!60!black,very thick,dash pattern=on 5pt off 3pt,forget plot]
  coordinates {(\sI,0) (\sI,{exp(-2.5*\sI)})};
  \addplot[teal!60!black,very thick,dash pattern=on 5pt off 3pt,forget plot]
  coordinates {(\sJ,0) (\sJ,{exp(-2.5*\sJ)})};
  \addplot[teal!60!black,very thick,dash pattern=on 5pt off 3pt,forget plot]
  coordinates {(\sK,0) (\sK,{exp(-2.5*\sK)})};
  \addplot[teal!60!black,very thick,dash pattern=on 5pt off 3pt,forget plot]
  coordinates {(\sL,0) (\sL,{exp(-2.5*\sL)})};
  \addplot[teal!60!black,very thick,dash pattern=on 5pt off 3pt,forget plot]
  coordinates {(\sM,0) (\sM,{exp(-2.5*\sM)})};
  
\addplot[only marks,mark=o,mark size=2.8pt,thick,black]
  coordinates {(\sA,{exp(-2.5*\sA)})
              (\sB,{exp(-2.5*\sB)})
              (\sC,{exp(-2.5*\sC)})
              (\sD,{exp(-2.5*\sD)})
              (\sE,{exp(-2.5*\sE)})
              (\sF,{exp(-2.5*\sF)})
              (\sG,{exp(-2.5*\sG)})
              (\sH,{exp(-2.5*\sH)})
              (\sI,{exp(-2.5*\sI)})
              (\sJ,{exp(-2.5*\sJ)})
              (\sK,{exp(-2.5*\sK)})
              (\sL,{exp(-2.5*\sL)})
              (\sM,{exp(-2.5*\sM)})};
\addlegendentry{Demand for Observed $p_i'$}

\node[below, text=teal!60!black] at (axis cs:\sA,0) {$p_1'$};
\node[below, text=teal!60!black] at (axis cs:\sB,0) {$p_2'$};
\node[below, text=teal!60!black] at (axis cs:\sC,0) {$p_3'$};
\node[below, text=teal!60!black] at (axis cs:\sD,0) {$p_4'$};
\node[below, text=teal!60!black] at (axis cs:\sE,0) {$p_5'$};
\node[below, text=teal!60!black] at (axis cs:\sF,0) {$p_6'$};
\node[below, text=teal!60!black] at (axis cs:\sG,0) {$p_7'$};
\node[below, text=teal!60!black] at (axis cs:\sH,0) {$p_8'$};
\node[below, text=teal!60!black] at (axis cs:\sI,0) {$p_9'$};
\node[below, text=teal!60!black] at (axis cs:\sJ,0) {$p_{10}'$};
\node[below, text=teal!60!black] at (axis cs:\sK,0) {$p_{11}'$};
\node[below, text=teal!60!black] at (axis cs:\sL,0) {$p_{12}'$};
\node[below, text=teal!60!black] at (axis cs:\sM,0) {$p_{13}'$};

\addplot[magenta!60, fill opacity=0.25]
  fill between[of=demand and xaxis, soft clip={domain=\pA:1}];
\addplot[magenta!60, fill opacity=0.25]
  fill between[of=demand and xaxis, soft clip={domain=\pB:1}];
\addplot[magenta!60, fill opacity=0.25]
  fill between[of=demand and xaxis, soft clip={domain=\pC:1}];
\addlegendimage{area legend,fill=magenta!60,draw=magenta!60,fill opacity=0.25}
\addlegendentry{Surplus from $p_i$}

\nextgroupplot[
  axis lines=left,
  ymin=0, ymax=1.05,
  xlabel={$z$ (price)},
  ylabel={CDF of target policy},
  ytick distance=0.2,
  legend style={draw=none,at={(0.03,0.97)},anchor=north west,font=\small},
  legend cell align=left
]

\addplot[black,ultra thick,const plot]
  coordinates {(0,0)
               (\pA,0)   (\pA,1/3)
               (\pB,1/3) (\pB,2/3)
               (\pC,2/3) (\pC,1)
               (1,1)};

\addplot[only marks,mark=o,mark size=2.8pt,thick,black]
  coordinates {(\sA,0) (\sB,0) (\sC,0) (\sD,0) (\sE,1/3) (\sF,1/3)  (\sG,1/3) (\sH,1/3) (\sI,2/3) (\sJ,2/3) (\sK,1) (\sL,1) (\sM,1)};

\end{groupplot}
\end{tikzpicture}
\caption{Illustrative example highlighting the cumulative propensity weights and their relationship to the area under the demand curve. See Example \ref{cumulative_weights_example} for more details. }
\label{fig:cumulative_weights_figure}
\end{figure}
 We next present a simple example, illustrated in Figure \ref{fig:cumulative_weights_figure}, to give further intuition behind the CPW estimator.

\begin{example} \label{cumulative_weights_example}

Suppose that we are estimating the consumer surplus for a target-pricing policy that assigns equal probability to three discrete prices \(p_1<p_2<p_3\), with the corresponding expected demand shown in green circles. Assume the context is the same for all customers. The consumer surplus under this target policy can be represented as a scaling of the average of the three areas to the right of the policy prices under the demand curve: $\mathcal S(\pi)
=\EE\left[\int_0^\infty\pi(p \given X) \int_{z=p}^\infty \mu(X,z)dzdp \right] = \frac{1}{3}\sum_{i=1}^{3} \frac{1}{3}\int_{z=p_i}^{\infty} \E[Y \given P = z]dz  .$ In Figure~\ref{fig:cumulative_weights_figure} these are the three pink regions. Where they overlap the tint is darker, reflecting the fact that the area to the right of \(p_3\) is counted in \emph{all three} integrals, the region between \(p_2\) and \(p_3\) is counted in \emph{two}, and the region between \(p_1\) and \(p_2\) is counted in \emph{one}.

Now assume that the historical data were generated by a (known) historical pricing policy, uniformly distributed from 0 to 1. In the historical data, we happen to observe 13 random realized prices $\{p_i'\}_{i=1}^{13}$, which is a sparse representation of what would occur with more samples. Each realized pair \((p_i', Y_i)\) gives an unbiased snapshot of expected demand at that threshold, i.e., \(\E[Y\mid P=p_i']\), where the black circles indicate the conditional expectations to which empirical averages converge with sufficient data for each price point. If we were to aggregate the outcomes \(Y_i\) (or expected outcomes with enough data), we would approximate the total area under the demand curve. However, to approximate the consumer surplus for the target pricing policy by aggregation, we need different weights. In particular, to recreate the previous surplus calculation, we must weight observations at $p'_{11}, p'_{12}, p'_{13}$ by three times as much as at $p'_5, p'_6, p'_7,p'_8$, since the area under the curve in \([p_3, 1)\) is included in the surplus calculation for all three prices under the target policy (dark pink), whereas $[p_1,p_2)$ is only included once (light pink). $[0,p_1)$ is not included at all, while $[p_2,p_3)$ is included twice. %
This weight is the fraction of target policy prices whose surplus includes that historical price and is proportional to the cumulative target mass, which for the three-point target policy is $F^{\pi}(p_i') \;=\; \frac{1}{3}\sum_{j=1}^{3}\mathbf 1\{p_i'\ge p_j\}$ (i.e., $F^{\pi}(p_1')=0, F^{\pi}(p_5')=1/3, F^{\pi}(p_9')=2/3, F^{\pi}(p_{11}')=1$). This is shown in the lower panel of Figure~\ref{fig:cumulative_weights_figure}. As a result, with more sampled price points along the range of prices, the weighted average $\frac{1}{n}\sum_{i=1}^n \frac{F^{\pi}(p'_i )}{ \pi_D(p'_i )}Y_i$ will eventually approximate the area under the demand curve, weighted by the frequency it is included in the target surplus calculation, i.e., $\frac{1}{3}\sum_{i=1}^{3} \pi(p_i) \int_{z=p_i}^{\infty} \E[Y \given P = z]dz$, as the number of samples gets large. \footnote{In this case, the historic pricing policy is uniform, $\pi(p'_i )=1$, so the denominator does not impact the calculation, but would otherwise be the usual inverse propensity correction to make the historical policy as if it were uniform in expectation.} 
\end{example}

This approach contrasts with standard IPW in off-policy evaluation in two significant ways. First, the observable $Y$ is not the unobserved surplus, $(V-P)_+$, we are trying to estimate under the new pricing policy. Second, the numerator of the weighting term is the cumulative target policy density $F^\pi(P\mid X)$, not the target density $\pi(P|X)$ at $P$ that typically appears in IPW. This is due to the need to estimate the average of an integral (or area) rather than the usual average outcome. As such, standard IPW techniques cannot be applied.

Compared to the direct method (Equation \eqref{eqn: identification}), this result requires knowledge, or an estimate, of the pricing distribution $\pi_D$ instead of the demand function $\mu(X, P)$. When the historic pricing policy is known, this estimator is unbiased and completely model-free. This may occur if a company is investigating the consumer surplus implications of its own algorithmic pricing policy, or a policymaker mandates that the algorithm be made available for audit.  Alternatively, if the historical pricing policy is relatively simple, it may be much easier to estimate than a complex demand function. Furthermore, the DM requires numerical integration over the price space $P$ for every observation, which can be computationally expensive. In contrast, the CPW estimator is a simple weighted average, making it computationally efficient for large datasets.
This alternative estimator provides the regulator or firm with crucial flexibility in surplus estimation. 

Nevertheless, the performance of the CPW estimator remains sensitive to the accuracy of the estimated historical pricing policy distribution when it is not available and may be subject to misspecification. To further address this issue, we introduce the augmented CPW (ACPW) estimator, which combines elements of both the DM and CPW approaches, remaining consistent if either component is correctly specified, and is therefore more robust.

\section{Doubly Robust Representation and Estimation}\label{sec:acpw}

The construction of the ACPW estimator is grounded in the theory of the efficient influence function (EIF). The EIF is pivotal for two main reasons: it characterizes the semiparametric efficiency bound (the minimal asymptotic variance of any regular estimator), and it provides a constructive mechanism for achieving this bound. Intuitively, the EIF acts as a correction term that removes the first-order bias from a naive plug-in estimator (e.g., DM or CPW estimators in our context). This correction is essential for ensuring that the final estimator remains $\sqrt{n}$-consistent and asymptotically normal, even when the nuisance components (such as the demand function and cumulative weights) converge at slower rates. Formally, the EIF is defined as the canonical gradient of the target parameter, e.g., $\mathcal{S}(\pi)$, $\mathcal S(\pi_D)$ and $\Delta(\pi)$,  with respect to the underlying data distribution. By constructing our estimator based on this gradient, we ensure it is asymptotically efficient (i.e., minimax optimal). For a comprehensive treatment of this theory, we refer readers to \citet{tsiatis2006semiparametric}. We now present the derived EIF for $\calS(\pi)$ under our semi-parametric model \eqref{model: true model}. Let $\mathcal D = (X, P, Y)$.

\begin{theorem} \label{thm:eif}
Suppose Assumptions \ref{assumption:ignorability} and \ref{assumption:overlap} hold, the EIF for $\calS(\pi)$ is 
        \begin{gather} \label{eqn: eif}
        \psi^\pi(\mathcal{D}) = \int_0^\infty \pi(p| X) \int_{p}^\infty \mu(X, z)dzdp  + \frac{F^\pi(P | X)}{\pi_D(P | X)}(Y - \mu(X,P))-\calS(\pi).
    \end{gather}  
    If the behavior policy $\pi_D$ is evaluated, then the EIF takes the form \begin{gather} \label{eqn: eif behav}
        \psi^{\pi_D}(\mathcal {D})= \int_{P}^\infty \mu(X, z)dz  + \frac{F^{\pi_D}(P | X)}{\pi_D(P | X)}(Y - \mu(X,P))-\calS(\pi_D).
    \end{gather} 
    Finally, the EIF for the difference in surplus $\Delta(\pi)$ is given as $\psi^{\Delta}(\cal{D}) = \psi^\pi(\cal {D}) -\psi^{\pi_D}(\cal {D})$.
\end{theorem}

This is formally proved in Appendix \ref{sec:eif_proof}. A key property of the EIF is that it has mean zero. This motivates the following estimators, defined by setting the empirical mean of the EIFs in Equations \eqref{eqn: eif} and \eqref{eqn: eif behav} to zero: 
\begin{align} \label{eq:empirical dr}
  \widetilde \calS_{ACPW}(\pi) &= \frac{1}{n} \sum_{i=1}^n  \bigg[\int_0^\infty\pi(p | X_i) \int_{p}^\infty \widehat\mu(X_i, z)dzdp  + \frac{F^\pi(P_i | X_i)}{\widehat \pi_D(P_i | X_i)}(Y_i - \widehat \mu(X_i,P_i))\bigg], \\
  \label{empirical dr2}\widetilde \calS_{ACPW}(\widehat \pi_D) &= \frac{1}{n} \sum_{i=1}^n  \bigg[\int_{P_i}^\infty \widehat\mu(X_i, z)dz +  \frac{ \widehat F^{\pi_D}(P_i | X_i)}{\widehat\pi_D(P_i | X_i)}(Y_i - \widehat \mu(X_i,P_i))\bigg], \\
  \widetilde \Delta_{ACPW}(\pi) &=  \widetilde \calS_{ACPW}(\pi)- \widetilde \calS_{ACPW}(\widehat \pi_D).
\end{align}

One can observe that the ACPW estimator integrates elements of both the DM and CPW approaches, and it can be shown that the ACPW estimator remains consistent if either the demand function or the behavior policy is correctly specified, a property known as double robustness.  The double robustness property is formally stated in the following proposition for the case of the known target policy, but the other cases also hold with near identical proofs. 
\begin{proposition} \label{prop: DR}
Let $\bar{\mu}(x,p)$ and $\bar{\pi}_D(p|x)$ denote the population limits of the estimators $\widehat{\mu}(x,p)$ and $\widehat{\pi}_D(p|x)$, respectively, such that: $$\sup_{x,p} |\widehat{\mu}(x,p) - \bar{\mu}(x,p)| =o_p(1) \quad \text{and} \quad \sup_{x,p} |\widehat{\pi}_D(p|x) - \bar{\pi}_D(p|x)| =o_p(1).$$  If either of the following conditions holds: (i) $\bar\mu(X,P)=\mu(X,P)$, almost surely; (ii) $\bar \pi_D(P|X)= \pi_D(P|X)$, almost surely.
Then we have consistency such that: \begin{gather*}
    \left|\widetilde \calS_{ACPW}(\pi)-\calS(\pi)\right|=o_p(1),
\end{gather*}
where $o_p(1)$ denotes a quantity that converges to zero in probability as the sample size $n \to \infty$.
\end{proposition}

The proof of Proposition \ref{prop: DR} can be found in Appendix \ref{sec:proof_DR_prop}. This property is important because it provides the regulator with flexibility in surplus estimation, depending on whether consistent demand estimation or historical pricing policy density estimation is possible to achieve. Besides the desirable doubly robust property, another advantage of the ACPW estimator is that, under minimal rate conditions on the two nuisance estimators (the demand function and the behavior policy density), it can achieve the lowest possible variance bound when combined with data splitting or a cross-fitting procedure \citep{chernozhukov2018double}.  {We next outline the $K$-fold cross-fitting procedure, a minor algorithmic modification of the vanilla ACPW estimators in Equations \eqref{eq:empirical dr} and \eqref{empirical dr2}. 
Specifically, we partition the sample indices $\{1,\ldots,n\}$ into $K$ disjoint folds of approximately equal size, with any finite number $K$. For each observation $i$, let $k(i)$ denote the fold containing $i$. Denote by $\widehat \mu^{-k(i)}(x,p)$ and $\widehat \pi_D^{-k(i)}(p \mid x)$ the estimators of the demand function and the behavior policy, respectively, which are trained using only the data excluding the $k(i)$-th fold (hence the notation $-k(i)$). The resulting ACPW estimator with cross-fitting is denoted as
\begin{gather*}
    \widehat \calS_{ACPW}(\pi)=   \frac{1}{n} \sum_{i=1}^n  \left[\int_0^\infty\pi(p | X_i) \int_{p}^\infty \widehat \mu^{-k(i)}(X_i,z) dzdp  + \frac{F^\pi(P_i | X_i)}{\widehat\pi_D^{-k(i)} (P_i | X_i)}(Y_i - \widehat \mu^{-k(i)}(X_i,P_i))\right],\\
    \widehat \calS_{ACPW}(\pi)=   \frac{1}{n} \sum_{i=1}^n  \left[\int_0^\infty\pi(p | X_i) \int_{p}^\infty \widehat \mu^{-k(i)}(X_i,z) dzdp  + \frac{F^{\widehat \pi_D, -k(i)}(P_i | X_i)}{\widehat\pi_D^{-k(i)} (P_i | X_i)}(Y_i - \widehat \mu^{-k(i)}(X_i,P_i))\right].
\end{gather*}
This cross-fitting approach ensures that each observation is evaluated using nuisance estimates fitted on independent data, thereby reducing overfitting bias and enabling valid asymptotic inference even when the nuisance functions are estimated by flexible machine learning methods, as will be detailed in Section \ref{sec: theory}.}

\section{Inequality-Aware Surplus}
\label{sec:iaacpw}
A common concern among policymakers when evaluating a pricing policy is not only how the aggregate surplus changes, but also how it is distributed among consumers. In particular, there are concerns about equity and the impact on those who are worst off. One approach used in welfare economics to address these issues is to emphasize outcomes for customers with the lowest surplus when aggregating surpluses across the population. We follow this direction by extending the proposed off-policy consumer surplus estimation techniques to welfare measures that are sensitive to how surplus is distributed across customer types.

Let $\mathcal{S}(\pi\mid X)$ denote the surplus for customers with characteristics $X$, which represent customer types. Earlier sections implicitly aggregated welfare using the arithmetic mean, $\mathcal{S}(\pi)=\mathbb{E}_{X}[\mathcal{S}(\pi\mid X)]$, effectively averaging across all customer types. Following the Atkinson tradition and related work \citep{bergson1954concept, atkinson1970measurement,lewbel2017unobserved, dube2023personalized}, we instead consider the generalized-mean family
\begin{equation}\label{eq:atkinson-agg}
(\mathcal{S}^r(\pi))^{1/r},
\end{equation} where $\mathcal{S}^r(\pi):=\mathbb{E}_{X}\big[\mathcal{S}(\pi\mid X)^r\big]$. 
This coincides with the standard arithmetic average when $r=1$ and increasingly prioritizes lower-surplus groups as $r$ decreases, becoming more inequality averse. This supplies a transparent policy parameter $r$ that trades off aggregate surplus against its dispersion across customer segments. The continuous extension at $r=0$ yields the geometric mean, and $r=-1$ yields the harmonic mean. In the following, for brevity, we focus on $r\neq 0$ to avoid restating results that are functionally the same.
For a finite sample $\{X_i\}_{i=1}^{n}$, the standard DM estimator is:
\begin{equation}\label{eq:atkinson-agg-sample}
\left(\frac{1}{n}\sum_{i=1}^{n}\widehat{\mathcal{S}}(\pi\mid X_i)^{r}\right)^{1/r}=\left[\frac{1}{n}\sum_{i=1}^{n} \left(\int_0^\infty\pi(p | X_i) \int_{p}^\infty \widehat\mu(X_i, z)dzdp\right)^r\right]^{1/r},
\end{equation}

Without loss of generality, we will focus on estimating $\mathcal{S}^r(\pi)$, which can be transformed after estimation to recover $\mathcal{S}^r(\pi) ^{1/r}$. Next, we derive the efficient influence function for the target $\mathcal{S}^r(\pi)$ and behavioral $\mathcal{S}^r(\pi_D)$ policies and their corresponding efficient estimators. The EIF is given by the following theorem.

\begin{theorem} \label{thm:eif aware}
Under Assumptions \ref{assumption:ignorability} and \ref{assumption:overlap}, for $r\neq 0$,  the EIF for $\calS^r(\pi)$ is 
        \begin{gather} 
       \nonumber  r  \frac{(Y-\mu(X,P))F^\pi(P|X)}{\pi_D(P|X)} \left(\int_0^\infty\pi(p | X) \int_{p}^\infty \mu(X, z)dzdp \right)^{r-1}\\
        +\left( \int_0^\infty\pi(p | X) \int_{p}^\infty \mu(X, z)dzdp \right)^r-\calS^r(\pi), \label{eqn: eif aware}
    \end{gather}  and the EIF for $\calS^r(\pi_D)$ is 
    \begin{gather}
       \nonumber  r  \left[\frac{(Y-\mu(X,P))F^{\pi_D}(P|X)}{\pi_D(P|X)} + \int^\infty_P \mu(X,z)dz \right]\left(\int_0^\infty\pi_D(p | X) \int_{p}^\infty \mu(X, z)dzdp \right)^{r-1} 
        \\
        + (1-r)\left( \int_0^\infty\pi_D(p | X) \int_{p}^\infty \mu(X, z)dzdp \right)^r-\calS^r(\pi_D).
    \end{gather}
    By linearity, the EIF of $\Delta^r(\pi)$ follows by taking the difference between the above two EIFs.
\end{theorem}

{Theorem \ref{thm:eif aware} establishes the EIF for the inequality-aware surpluses, $\calS^r(\pi)$, $\calS^r(\pi_D)$ and their difference $\Delta_r(\pi)$ and is proved in Appendix \ref{sec:eif_proof_inequality}. In more detail, Equation \eqref{eqn: eif aware} shares a structure analogous to the EIF for the standard surplus in Equation \eqref{eqn: eif}: the $\int_0^\infty\pi(p | X) \int_{p}^\infty \mu(X, z)dzdp$ term corresponds to the DM component, while $ \frac{(Y-\mu(X,P))F^\pi(P|X)}{\pi_D(P|X)}$ is a mean-zero de-biasing term incorporating cumulative weights, which plays a crucial role in variance reduction, enabling the resulting estimator to attain the semiparametric efficiency bound. } Naturally, Theorem \ref{thm:eif aware} motivates the following estimators to estimate $\calS^r(\pi)$ and $\calS^r(\pi_D)$ respectively, obtained by setting the empirical mean of the EIF to zero together with cross-fitting:
\begin{align}\label{eq:aware estimator}
 \widehat \calS^r(\pi)= \frac{1}{n}\sum_{i=1}^n \Bigg[ &r \frac{(Y_i-\widehat \mu^{-k(i)}(X_i,P_i))F^\pi(P_i|X_i)}{\widehat{\pi}^{-k(i)}_D(P_i|X_i)} \left(\int_0^\infty \pi(p | X_i) \int_{p}^\infty \widehat \mu^{-k(i)}(X_i, z)dzdp\right)^{r-1} \nonumber \\
 &+ \left(\int_0^\infty\pi(p | X_i) \int_{p}^\infty \widehat \mu^{-k(i)}(X_i, z)dzdp \right)^r \Bigg],
\end{align}  

\begin{align*}
       \nonumber   \widehat \calS^r(\pi_D) =  \frac{1}{n}\sum_{i=1}^n &\Bigg[r  \left(\frac{(Y_i-\widehat \mu^{-k(i)}(X_i,P_i))\widehat F^{\pi_D,-k(i)}(P_i|X_i)}{\widehat\pi_D^{-k(i)}(P_i|X_i)} + \int^\infty_{P_i} \widehat \mu^{-k(i)}(X_i,z)dz \right) \\
& \quad \times \left(\int_0^\infty \widehat \pi_D^{-k(i)}(p | X_i) \int_{p}^\infty \widehat \mu^{-k(i)}(X_i, z)dzdp \right)^{r-1} \\
        & + (1-r)\left( \int_0^\infty \widehat\pi_D^{-k(i)}(p | X_i) \int_{p}^\infty \widehat \mu^{-k(i)}(X_i, z)dzdp \right)^r\Bigg]
    \end{align*}
Then $\widehat \Delta^r(\pi)=\widehat \calS^r(\pi) - \widehat \calS^r(\pi_D)$.

We refer to this class of estimators as the inequality-aware ACPW estimator (IA-ACPW). Although the estimator $\widehat{\mathcal{S}}^r(\pi)$ in Equation \eqref{eq:aware estimator} is derived from the  EIF in \eqref{eqn: eif aware} and incorporates both DM and CPW elements, it does not possess the double robustness property typically associated with the ACPW estimator. This stems from the fact that the functional $S^r(\pi)$ is nonlinear when $r \neq 1$. For example, the functional reweights observations such that small surpluses gain more leverage for $r < 1$. Consequently, a single data point perturbs the estimator differently than it would under a simple arithmetic mean. Due to this nonlinearity, $\widehat{\mathcal{S}}^r(\pi)$ is consistent only if the demand model is correctly specified, although the historic pricing policy $\widehat \pi_D$ can be misspecified. The necessity of consistency for the demand estimation can clearly be seen in Equation \eqref{eq:aware estimator}, where the first term disappears when $\frac{1}{n}\sum_{i=1}^n \left(Y_i-\widehat \mu^{-k(i)}(X_i,P_i)\right)$ converges to 0, leaving the second term which is clearly only consistent when demand is consistent. However, as we will show in the asymptotic analysis in Section \eqref{sec:normality_aicpw}, the cumulative weights play an important role in reducing variance, enabling the use of flexible machine learning policies with slower convergence rates. 

The estimation of the inequality-aware surplus of the behavioral policy $\mathcal{S}^r(\pi_D)$ presents an even greater challenge. Unlike $\widehat{\mathcal{S}}^r(\pi)$, the estimator $\widehat \calS^r(\pi_D)$ does not enjoy even single robustness; it requires the simultaneous consistency of both the demand model $\widehat \mu$ and the propensity model $\widehat \pi_D$. This added fragility arises because  $\pi_D$ is not known but estimated, and it serves a dual role: it acts as the propensity weight in the debiasing term and explicitly defines the integration measure in the demand term. Consequently, if $\widehat \pi_D$ is misspecified, the estimator converges to a functional of the wrong policy, preventing consistency even if the demand model $\widehat \mu$ is perfect.

An example showing inconsistency of $\widehat{\mathcal{S}}^r(\pi)$ with a misspecified demand model is given next.

\begin{example} (Inequality-aware ACPW estimator is not robust to demand misspecification)
Consider estimating the functional 
\[
\calS^r(\pi) = \biggl(\int_0^\infty \pi(p) \int_{z=p}^\infty \mu(z)\,dz\,dp\biggr)^r
\quad\text{with}\quad r=\tfrac12,
\]
in a setting without covariates, where
\(\mu(\cdot)\equiv \mathbb{E}[Y\mid P=\cdot]\) denotes the demand function.

Assume \(P\sim\mathrm{Uniform}[0,1]\), so the behavior policy satisfies
\(\pi_D(p)=1\) for all \(p\in[0,1]\).
Let the target policy also be $\pi(p)=1$, which implies \(F^\pi(z)=z\), and suppose the true demand function is $\mu(p)=1-p^2.$
Under this setup, define 
\[
\theta^*=\int_0^1 \pi(p) \int_p^1 \mu(z)\,dz\,dp
= \int_0^1 1 \cdot \left[ z - \frac{z^3}{3} \right]_p^1 \,dp
= \int_0^1 \left(\frac{2}{3} - p + \frac{p^3}{3}\right)\,dp
= \frac14,
\]
so that 
\(\mathcal{S}^r(\pi)=(\theta^*)^{1/2}=\frac12.\)

Now, suppose the analyst uses a misspecified linear demand model whose limiting
fit is
$\bar\mu(p)=1-\tfrac12 p .$ The corresponding first-stage limit is
\[
\bar\theta
=\int_0^1 \pi(p) \int_p^1 \bar\mu(z)\,dz\,dp
=\int_0^1 1 \cdot \left[ z - \frac{z^2}{4} \right]_p^1 \,dp
=\int_0^1 \left(\frac{3}{4} - p + \frac{p^2}{4}\right)\,dp
=\frac13.
\]
Hence, the population limit of the IA-ACPW estimator is 
\[
\bar\theta^{1/2}
\;+\;
\frac12\,\bar\theta^{-1/2}
\,
\underbrace{\mathbb{E}\!\left[
\frac{F^\pi(P)}{\pi_D(P)}\,
\bigl(Y-\bar\mu(P)\bigr)
\right]}_{\int_0^1 z(0.5 z-z^2)\,dz
=-1/12}=\sqrt{\frac13}
\;-\;
\frac{\sqrt{3}}{24},
\]
which differs from the true value \(1/2\). Therefore, even when the behavior policy is correctly specified, misspecification of
the demand model combined with the nonlinearity of the target functional yields a
non-vanishing second-order remainder term and leads to bias.
\end{example}

\section{Theoretical Analysis}\label{sec: theory}

In this section, we establish the asymptotic normality of our proposed estimators and highlight the conditions required to achieve this. 
{These results are particularly important, since they enable statistical inference and the construction of confidence intervals for surplus estimates. In practice, this allows firms to rigorously assess whether changes in their pricing strategy lead to statistically significant improvements in overall surplus or consumer welfare. For example, an e-commerce company may wish to evaluate whether a newly deployed dynamic pricing algorithm $\pi$ yields a higher expected consumer surplus than the existing pricing strategy $\pi_D$. By constructing confidence intervals for the respective estimators, the firm can formally test whether the observed improvement is statistically meaningful, rather than the result of random variation in sales data.}

We also show that these estimators attain the semiparametric efficiency bounds. This means that, among all regular and asymptotically linear estimators, our proposed estimators achieve the lowest possible asymptotic variance. Although all estimators for the same target achieve this bound, the assumptions required to achieve it can differ, and in particular we show that the conditions for ACPW are relatively mild, allowing fast rates of convergence even if the demand function or historical price density estimates converge at slower rates. Throughout, we assume that all derived EIFs have finite second moments, $\widehat \pi_D(p|x) > c$ for some constant $c$, and that $\widehat \mu(x,p)$ is bounded for all $p$ and $x$. 

In what follows, we start with the standard consumer surplus $(r=1)$, where the strongest results can be established, before progressing to the inequality-aware surplus, which presents additional challenges. We focus on analyzing the estimators for the surplus of a known policy $\pi$. Analogous results for the behavior policy surplus $\calS(\pi_D)$ and difference $\Delta(\pi)$ follow similar techniques and can be found in Appendix \ref{sec:behavioral_theory}. For completeness, we additionally present the results for the DM method, which forms part of our theoretical contribution. 

\subsection{Analysis of Standard Consumer Surplus $(r=1)$}
We impose three sets of technical conditions corresponding to three estimators: CPW, ACPW, and DM, although we cover assumptions for DM in Appendix \ref{sec:assumptions_DM} for brevity. We begin with the assumptions required for the CPW estimator. 

\subsubsection{Required Assumptions}
\begin{assumption} [Assumptions required for the CPW] \label{asmp: CPW}
(i) $\sqrt{\E\left[\widehat \omega(X,P)-\omega(X,P)\right]^2}=o_p(1)$ , where $\omega(x,p)\equiv\frac{F^\pi(p | x)}{\pi_D(p | x)}$, and $\widehat \omega(x,p)\equiv\frac{F^\pi(p | x)}{\widehat\pi_D(p | x)} $ is the estimator of  $\omega(x,p)$. \\
(ii) The behavior policy is estimated using a function class that satisfies the Donsker property.\\
   (iii) There exist basis functions $\phi(x,p) \in \mathbb{R}^L$  
and a vector $\beta \in \mathbb{R}^L$ 
such that \begin{gather} \label{eq:approximation}
    \sup_{x,p} |\mu(x,p)-\phi(x,p)^\top \beta | =O(L^{-s/d}),
\end{gather} where $s$ is a fixed positive constant and $O(\cdot)$ is the standard big-$O$ term. \\
(iv) The estimated CPW weights satisfy \begin{gather*}
       \left \lVert\frac{1}{n} \sum_{i=1}^n \phi^\pi(X_i)- \frac{1}{n} \sum_{i=1}^n \widehat \omega(X_i,P_i)\phi(X_i,P_i) \right \rVert_2  =o_p(n^{-1/2}),
    \end{gather*} where $\| \cdot \|_2$ is denoted as Euclidean norm, $\phi(\cdot, \cdot)$ is the basis function that satisfy Equation \eqref{eq:approximation}, and
    $\phi^\pi(x)=\int^\infty_0  \pi(p | x) \int_p \phi(x,z)dz dp $.
\end{assumption}

Assumption \ref{asmp: CPW} (i) is relatively mild, as it merely requires $\widehat \pi_D$ to be consistent, with no rate specified. Assumption \ref{asmp: CPW} (ii)
imposes a complexity (size) constraint on the function class used for estimating the behavior policy. Before further discussing this assumption, we formally define the Donsker class as follows.
{
\begin{definition}[$P$-Donsker Class]
Let $(\mathcal{X}, \mathcal{A}, P)$ be a probability space and $\mathcal{F}$ be a class of measurable functions. We denote by $L_2(P)$ the space of all measurable functions that are square-integrable with respect to $P$. The associated $L_2$-norm is defined as $\|\cdot \|_{P,2}$. We define the following components to characterize the complexity of $\mathcal{F}$:
\begin{itemize}
    \item Brackets: for any two functions $l, u \in L_2(P)$, the bracket $[l, u]$ is the set of functions $\{f : l(x) \leq f(x) \leq u(x) \text{ for all } x \in \mathcal{X}\}$. An $\epsilon$-bracket in $L_2(P)$ is a bracket $[l, u]$ such that $\|u - l\|_{P,2} < \epsilon$.
    
    \item Bracketing number: the bracketing number $N_{[]}(\epsilon, \mathcal{F}, L_2(P))$ is the minimum number of $\epsilon$-brackets in $L_2(P)$ needed to cover $\mathcal{F}$.
    
    \item Bracketing integral: the entropy integral is defined as
    \[ J_{[]}(\delta, \mathcal{F}, L_2(P)) = \int_{0}^{\delta} \sqrt{\log N_{[]}(\epsilon, \mathcal{F}, L_2(P))} \, d\epsilon \]
\end{itemize}

Assume that there exists a measurable function $F$ such that $|f(x)| \leq F(x)$ for all $f \in \mathcal{F}$ and $x \in \mathcal{X}$, with $\EE(F^2) < \infty$. Then the class $\mathcal{F}$ is called a \textbf{$P$-Donsker class} if    $J_{[]}(\delta, \mathcal{F}, L_2(P)) < \infty$, for some $\delta > 0$.
\end{definition}}

Intuitively speaking, a Donsker class is a collection of functions that is not too large or too complex. This helps ensure that the average behavior of these functions becomes stable as we collect more data.  \bla  Many commonly used machine learning models form Donsker classes. These include standard parametric models such as linear and generalized linear models, as well as nonparametric regression methods like wavelets and tensor product B-splines (see Section 6 of \citet{chen2015optimal} for a review).
However, it is important to note that many modern black-box machine learning algorithms do not inherently satisfy the Donsker property. High-capacity models, such as over-parameterized deep neural networks, unpruned random forests, or gradient boosting machines, often operate in function spaces with massive complexity. Without explicit structural constraints (e.g., sparsity, norm regularization, or bounded depth), these classes can be too rich to admit a uniform Central Limit Theorem. Nevertheless, under specific technical conditions, even these state-of-the-art models can be shown to satisfy the Donsker property. For instance, \citet{breiman2017classification} discusses conditions for decision trees, while \citet{zhou2023offline} (Lemma 4) and \citet{Schmidt-Hieber2020} (Lemma 5) provide the necessary sparsity and boundedness constraints for decision trees and neural networks, respectively, to remain within the Donsker regime.

Assumption \ref{asmp: CPW}(iii) is commonly adopted in the policy evaluation literature, see, for example, \citet{shi2022statistical, chen2022well, bian2025off}. When the demand function $\mu(x,p)$ lies within a Hölder or Sobolev smoothness class, Assumption \ref{asmp: CPW}(iii) is automatically satisfied, with $s$ as the Hölder smoothness parameter of the function $\mu(x,p)$. In such cases, one can approximate $\mu(x,p)$ using wavelet or tensor product B-spline basis functions.  
It can be observed from Equation  \eqref{eq:approximation} that the smoother the function and the greater the number of basis functions, the smaller the approximation error. %

Assumption \ref{asmp: CPW}(iv) requires the estimated CPW weight to have the approximately balancing property. It is relatively mild, since Lemma \ref{lemma: balancing} (see Appendix \ref{sec:lemma_banlancing}) establishes that the true weight satisfies \begin{align*}
    \E\left[\phi^\pi(X) - \omega(X,P)\phi(X,P)\right] = 0.
\end{align*} In fact, exact balance over empirical data, meaning equality with $0$ rather than convergence at rate $o_p(n^{-1/2})$, can be achieved when the number of basis functions $L$ is fixed \citep{graham2012inverse}. Note that the $o_p(n^{-1/2})$ balance can still be achieved in settings where the number of basis functions $L$ grows with the sample size \citep{wang2023projected}, and one can similarly follow the approach therein to construct a CPW estimator that satisfies Assumption \ref{asmp: CPW} (iv).

Next, we present assumptions to show the asymptotic properties of ACPW.

\begin{assumption}[Assumptions required for the ACPW] \label{asmp: ACPW}
Assume $\pi_D(p \given x)>c$, for all $p \in \calP$, and every $x$, for some constant $c$. In addition, suppose that the estimators for the demand function and the behavior policy are constructed using the cross-fitting procedure, and that they achieve the following convergence rate for $k = 1, \cdots, K$: 
    \begin{align} \label{eq: product rate}
        \sqrt{\E[(\widehat \mu^{-(k)}(X,P)-\mu(X,P))^2]} =O_p(n^{-\alpha_1}), \mbox{ and } \; \sqrt{\E[(\widehat \omega^{-(k)}(X,P)-\omega(X,P))^2]}=O_p(n^{-\alpha_2}),
    \end{align} with $\alpha_1, \alpha_2>0$, and $\alpha_1+\alpha_2>1/2$.
\end{assumption}

Assumption \ref{asmp: ACPW} allows the nuisances to be estimated at rates slower than the parametric $O_p(n^{-1/2})$, making it a mild condition. For example, it is satisfied when both estimators achieve $o_p(n^{-1/4})$, thereby accommodating flexible machine learning methods for consumer surplus estimation. Given these assumptions, we are able to show asymptotic normality of the proposed estimators.

\subsubsection{Asymptotic Normality}
\begin{theorem} \label{thm:dr rate}
Under Assumptions \ref{assumption:ignorability} and \ref{assumption:overlap}, the following results hold:\\
(i)  Suppose Assumption \ref{asmp: CPW} holds, and further assume that the number of basis functions $L$ satisfies $L \gg n^{d/2s}$, then 
    \begin{gather*}
     \sqrt{n} \left( \widehat \calS_{CPW}(\pi)-\calS(\pi)\right)\rightarrow\mathcal{N}\left(0,\Sigma(\pi) \right),
    \end{gather*} where $\Sigma(\pi)\equiv \mbox{Var}[\psi^\pi(\cal {D})]$, and $\psi^\pi(\cal {D})$ is the EIF for $\calS(\pi)$ given by Equation \eqref{eqn: eif}. 

  (ii)  Under Assumption \ref{asmp: ACPW},  
    \begin{gather*}
     \sqrt{n} \left( \widehat \calS_{ACPW}(\pi)-\calS(\pi)\right)\rightarrow \mathcal{N}\left(0,\Sigma(\pi) \right),
    \end{gather*}

 (iii)   Under Assumption \ref{asmp: dm} in Appendix \ref{sec:assumptions_DM}, 
    \begin{gather*}
     \sqrt{n} \left( \widehat \calS_{DM}(\pi)-\calS(\pi)\right)\rightarrow \mathcal{N}\left(0,\Sigma(\pi) \right).
    \end{gather*}
\end{theorem}

This is proved in Appendix \ref{sec:normality_target}. Theorem \ref{thm:dr rate} shows that all three proposed estimators achieve the semiparametric efficiency bound, i.e., among all regular asymptotically linear estimators, they attain the minimal asymptotic variance $\psi^\pi(\cal {D})$. They also converge to the same asymptotic distribution. However, the conditions required for attaining the efficiency bound differ across methods. In particular, the DM and CPW estimators rely on their respective nuisance functions being estimated within a Donsker class, together with additional requirements such as smoothness and balancing. In contrast, the ACPW estimator avoids such high-level conditions, requiring only a mild product-rate assumption and sample splitting during estimation. This is significant as it ensures that the ACPW estimator can have a fast convergence rate even when the DM and CPW methods have slower convergence rates (such as with more complicated machine learning estimators). As a result, ACPW has the flexibility to be used across a wider range of settings for surplus estimation.

It is also instructive to discuss why the plug-in estimators such as CPW estimator can achieve the asymptotic normality with the same rate and variance as ACPW. The intuition is that Assumption 3 elevates the CPW estimator from an inverse-probability weighting method to a calibrated estimator. The empirical average of the de-biasing term in $\psi^\pi(\mathcal D)$ (i.e., 
\\ 
$\int_0^\infty\pi(p | X) \int_{p}^\infty \mu(X, z)dzdp  - \frac{F^\pi(P | X)}{\pi_D(P | X)}\mu(X,P)$) is asymptotically negligible under Assumption 3. Specifically, Assumption 3(iv) enforces a constraint that forces the estimated weights to balance the empirical moments of the historical data against the target policy, using the basis functions $\phi$ as the balancing features. This connects directly to the demand model: since Assumption 3(iii) guarantees that the true demand function $\mu$ can be accurately approximated by a linear combination of these same basis functions, balancing $\phi$ effectively balances the demand function itself. Quantitatively, provided the number of basis functions $L$ is chosen sufficiently large (specifically $L \gg n^{d/2s}$), the approximation error becomes negligible at the root-$n$ scale ($o(n^{-1/2})$). By calibrating the weights to remove the variation explained by the covariates, the CPW estimator achieves the same error reduction as explicitly subtracting a control variate. Consequently, it attains the same semiparametric efficiency bound and asymptotic normality as the doubly robust ACPW estimator, even without explicitly estimating the demand function.

An analogous result to Theorem \ref{thm:dr rate} applies to the estimation of the behavior policy surplus, $\calS(\pi_D)$, and the difference in surplus $\Delta(\pi)$. For a complete statement, see Theorem \ref{thm:dr rate behavior} and Corollary \ref{thm:dr rate behavior difference} in Appendix \ref{sec:behavioral_theory}.

\subsubsection{Confidence Intervals}

Based on the asymptotic normality established in Theorem \ref{thm:dr rate}, we can construct valid confidence intervals for $\calS (\pi)$. A key theoretical insight from our analysis is that, although $\widehat \calS_{CPW}$, $\widehat \calS_{ACPW}$, and $\widehat \calS_{DM}$ rely on different modeling strategies, they are all asymptotically linear estimators governed by the same efficient influence function, $\psi^{\pi}(\mathcal{D})$. Consequently, they share the same asymptotic variance, $\Sigma(\pi) = \text{Var}(\psi^{\pi}(\mathcal{D}))$.

To perform inference, we estimate this variance using the empirical second moment of the estimated EIF. This provides a unified approach to variance estimation, since regardless of whether the point estimate is derived via direct modeling or propensity weighting, the uncertainty is quantified by the variability of the underlying influence function. The variance estimators for each method are given by:
\begin{align*}
   &\widehat \Sigma_{CPW}(\pi)=   \frac{1}{n} \sum_{i=1}^n  \left[\int_0^\infty\pi(p | X_i) \int_{p}^\infty \widehat \mu(X_i,z) dzdp  + \frac{F^\pi(P_i | X_i)}{\widehat\pi_D (P_i | X_i)}(Y_i - \widehat \mu(X_i,P_i))-\widehat \calS_{CPW}(\pi)\right]^2,\\
   &\widehat \Sigma_{ACPW}(\pi)=   \frac{1}{n} \sum_{i=1}^n  \bigg[\int_0^\infty\pi(p | X_i) \int_{p}^\infty \widehat \mu^{-k(i)}(X_i,z) dzdp \\
   &\qquad \qquad \qquad \qquad \qquad  + \frac{F^\pi(P_i | X_i)}{\widehat\pi_D^{-k(i)} (P_i | X_i)}(Y_i - \widehat \mu^{-k(i)}(X_i,P_i))-\widehat \calS_{ACPW}(\pi)\bigg]^2,\\
   & \widehat \Sigma_{DM}(\pi)=   \frac{1}{n} \sum_{i=1}^n  \left[\int_0^\infty\pi(p | X_i) \int_{p}^\infty \widehat \mu(X_i,z) dzdp  + \frac{F^\pi(P_i | X_i)}{\widehat\pi_D (P_i | X_i)}(Y_i - \widehat \mu(X_i,P_i))-\widehat \calS_{DM}(\pi)\right]^2.
\end{align*} 

Note that for $\widehat \Sigma_{CPW}(\pi)$, we utilize the estimated demand function $\widehat \mu$ to construct the variance estimator, even though it is not used for the point estimate $\widehat \calS_{CPW}(\pi)$ itself. Similarly, for $\widehat \Sigma_{CPW}(\pi)$, we utilize the estimated propensity weights $\widehat \pi_D$ to construct the variance estimator.
The following proposition establishes the consistency of these variance estimators.

\begin{proposition} \label{prop:variance}
    Under the assumptions of Theorem \ref{thm:dr rate}, the following consistency results hold:
    \begin{gather*}
        |\widehat \Sigma_{CPW}(\pi) - \Sigma(\pi)| = o_p(1), \quad 
        |\widehat \Sigma_{ACPW}(\pi) - \Sigma(\pi)| = o_p(1), \quad
        \text{and} \quad |\widehat \Sigma_{DM}(\pi) - \Sigma(\pi)| = o_p(1).
    \end{gather*}
\end{proposition}
This is proved in Appendix \ref{sec:variance_consistency}. Accordingly, a $(1-\alpha)$ confidence interval for $\mathcal S(\pi)$ can be constructed using any of the three estimators. For example, using the ACPW estimator, the interval is given by:
\begin{equation*}
\left[\widehat{\mathcal S}_{ACPW}(\pi) \; \pm\; z_{1-\alpha/2} \sqrt{\widehat \Sigma_{ACPW}(\pi)/n} \right],
\end{equation*}
where $z_{1-\alpha/2}$ denotes the upper $(1-\alpha/2)$-quantile of the standard normal distribution. In large samples, this interval contains the true value $\mathcal S(\pi)$ with probability $(1-\alpha)$. Intervals for the CPW and DM estimators are constructed analogously. This allows firms and regulators to establish whether surplus improvements or declines are statistically meaningful, and is important for rigorous evaluation.

\subsection{Asymptotic Normality of Inequality-Aware Surplus $(r\neq 1)$ }
\label{sec:normality_aicpw}
We next establish the asymptotic normality of $\widehat \calS^r(\pi)$ and show that it attains the semiparametric efficiency bound.

\begin{theorem} \label{thm: aware rate} 
  Suppose that Assumptions \ref{assumption:ignorability}, \ref{assumption:overlap}, and \ref{asmp: ACPW} hold. In addition, assume that  $\alpha_1>1/4$, then for $r\neq 0$, we have 
    \begin{align*}
       \sqrt{n} (\widehat \calS^r(\pi)- \calS^r(\pi))\rightarrow \mathcal{N}(0,\Sigma^r(\pi)),
    \end{align*} where $\Sigma^r(\pi)$ is the variance of the EIF in Equation \eqref{eqn: eif aware}. 
\end{theorem}

Theorem \ref{thm: aware rate} establishes that the proposed inequality-aware estimator achieves the semiparametric efficiency bound, and is proved in Appendix \ref{sec:normality_unequal}. A key distinction of this result, compared to the aggregate surplus case ($r=1$) presented in Theorem \ref{thm:dr rate}, is the requirement that the demand model converges at a rate faster than $n^{-1/4}$ (i.e., $\alpha_1 > 1/4$). This condition is stricter than the product-rate condition ($\alpha_1 + \alpha_2 > 1/2$), which is sufficient for the standard ACPW estimator.

This divergence arises from the nonlinearity of the target functional $\mathcal{S}^r(\pi) = \mathbb{E}[(\cdot)^r]$ when $r \neq 1$. In the analysis of the linear aggregate surplus ($r=1$), the remainder term of the estimator takes the form of a cross-product of errors between the demand and propensity models, $ (\widehat \omega - \omega)(\widehat{\mu} - \mu)$. This product structure allows for a trade-off in accuracy between the two nuisance functions, underpinning the double robustness property. In contrast, expanding the nonlinear functional $\mathcal{S}^r(\pi)$ via a second-order Von Mises expansion introduces a purely quadratic error term associated with the curvature of the functional. Consequently, the remainder term includes a component proportional to $\|\hat{\mu} - \mu\|^2$. For the estimator to be $\sqrt{n}$-consistent and asymptotically normal, this quadratic term must vanish faster than $n^{-1/2}$. This necessitates that $\|\hat{\mu} - \mu\| = o_p(n^{-1/4})$, forcing the demand estimator to satisfy a stricter individual convergence rate. This mathematical necessity aligns perfectly with our observation that the IA-ACPW estimator is only single-robust: because the propensity score cannot cancel out the quadratic error introduced by the nonlinearity, the consistency of the final estimator becomes dependent on the quality of the demand model.

Despite this stricter condition, the IA-ACPW estimator offers an important advantage over a direct method estimator that integrates a plug-in model of demand, with the inequality-aware aggregation applied. Such an estimator typically inherits the first-order bias of the demand model (i.e., error terms proportional to $|\hat{\mu} - \mu|$). When flexible machine learning methods are used to estimate demand, this bias often decays too slowly to permit valid statistical inference, rendering confidence intervals unreliable. In contrast, the IA-ACPW estimator leverages the cumulative propensity weights to perform a one-step bias correction. This correction effectively removes the first-order bias, leaving only the second-order quadratic remainder described above. Consequently, even if the demand model converges at a slower nonparametric rate (provided $\alpha_1 > 1/4$), the bias of the IA-ACPW estimator becomes negligible relative to its variance. This property is important for practitioners, as it enables the construction of valid confidence intervals while utilizing high-complexity machine learning models for demand estimation.

We prove similar results of inequality-aware for the behavioral policy in Appendix \ref{sec:normality_unequal_behavioral}. However, establishing the asymptotic normality of $\widehat{\mathcal{S}}^r(\pi_D)$ imposes strictly stronger conditions on the nuisance estimators than those required for the counterfactual policy $\pi$. Specifically, the theory requires that both the demand model $\widehat{\mu}$ and the propensity model $\widehat{\pi}_D$ converge at a rate faster than $n^{-1/4}$ in the $L_2$-norm (i.e., $\|\widehat{\mu}-\mu\| = o_p(n^{-1/4})$ and $\|\widehat{\pi}_D-\pi_D\| = o_p(n^{-1/4})$). This simultaneous requirement is a direct consequence of the estimator's lack of robustness. Unlike the standard setting, where the error term factorizes into a product of nuisance errors allowing for a rate trade-off (e.g., a slower propensity model can be compensated by a faster demand model), the nonlinearity of the functional $\mathcal{S}^r(\pi_D)$ with respect to the generated distribution introduces independent quadratic error terms for both models. Consequently, there is no safety net: if either model converges slower than $n^{-1/4}$, the second-order bias terms will not vanish at the $\sqrt{n}$-rate, preventing the estimator from achieving asymptotic normality. This highlights the inherent difficulty of evaluating inequality metrics on the behavioral policy itself when that policy must be estimated from the same data.

\section{Experiments}
\label{sec:experiments}

This section presents the experimental results. In Section \ref{sec:exp_dr}, we demonstrate that our proposed estimator has the double robustness property, unlike existing surplus estimators, and the validity of the confidence intervals. We then evaluate the performance of the proposed inequality-aware surplus estimators in Section \ref{sec:exp_ineq} and analyze their confidence intervals in Section \ref{sec:exp_conf}. Finally, to illustrate the practical application of our method,  Section \ref{sec:exp_auto} applies our framework to a dataset from an auto loan company to measure consumer surplus. 

\subsection{Double Robustness} \label{sec:exp_dr}
This section empirically validates the double robustness of the proposed augmented cumulative propensity weighting (ACPW) estimator. By simulating scenarios where either the demand model or the pricing (propensity) model is intentionally misspecified, we demonstrate that the ACPW estimator remains consistent as long as at least one of these components is correctly specified.

\noindent \textbf{Direct Model Misspecification:} We assume $X\sim U[-0.5,0.5], \beta \sim U[-0.1,0.1], V=90+3000\beta X^2 + \epsilon, \epsilon\sim U[0,10], P\sim U[\min(V)-0.05, \max(V)+0.05], Y=\mathbb{I}[V>P]$. We use a linear model to model the direct model $\mu(x,p)$, which is misspecified. The propensity model is correctly specified using a top-hat kernel density estimator. 

\noindent \textbf{Propensity Model Misspecification:} We assume $X\sim U[0,1], \beta \sim U[-1,1], V=100+300\beta X + \epsilon, \epsilon\sim U[0,10], P\sim U[\min(V)-0.05, \max(V)+0.05], Y=\mathbb{I}[V>P]$. We use a Gaussian density model to model the propensity model $\pi_D(p|x)$, which is misspecified. The direct model is correctly specified using a linear model.  

In both scenarios, we evaluate the Mean Squared Error (MSE) for surplus estimates under both the historical ``Behavior'' policy and a new ``Target'' personalized pricing policy (trained via gradient boosting trees). Specifically, we employ Gradient Boosted Trees (GBT) to estimate the demand function, $\mu(x, p)$. For the target policy, the final price is then determined by applying a softmax transformation over the expected revenue, $p \cdot \mu(x, p)$, evaluated across a discrete grid of five price points. 

\begin{figure}
\begin{subfigure}[b]{0.5\textwidth}
         \centering
         \includegraphics[width=\linewidth]{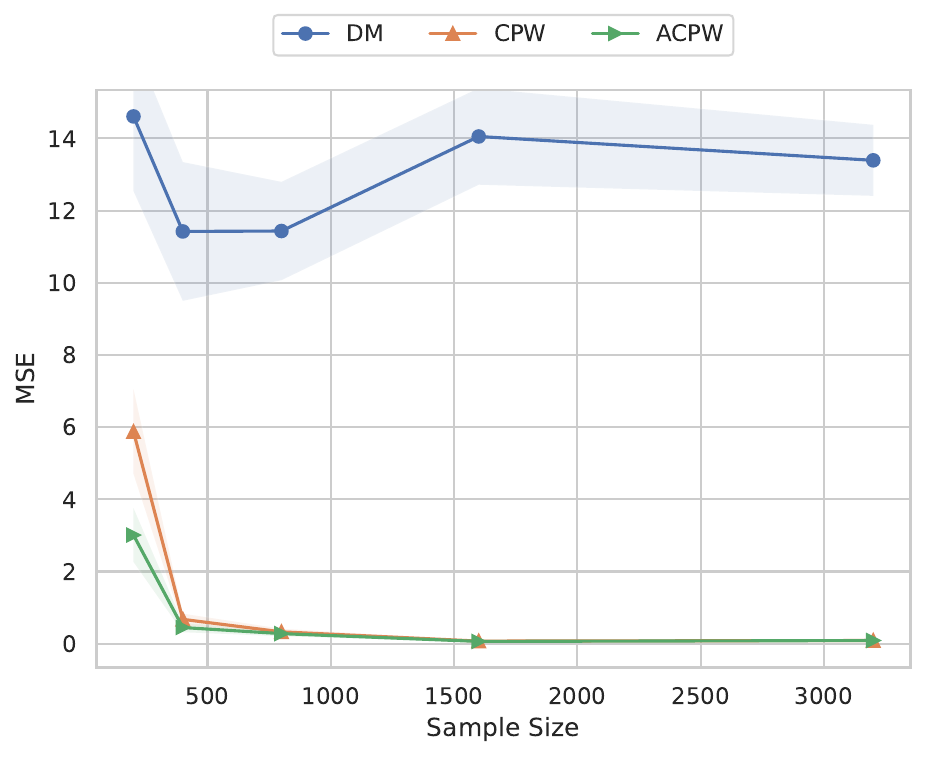}
         \caption{Behavior}
         \label{fig:dr_dmmis_beh}
     \end{subfigure}
\begin{subfigure}[b]{0.5\textwidth}
         \centering
         \includegraphics[width=\linewidth]{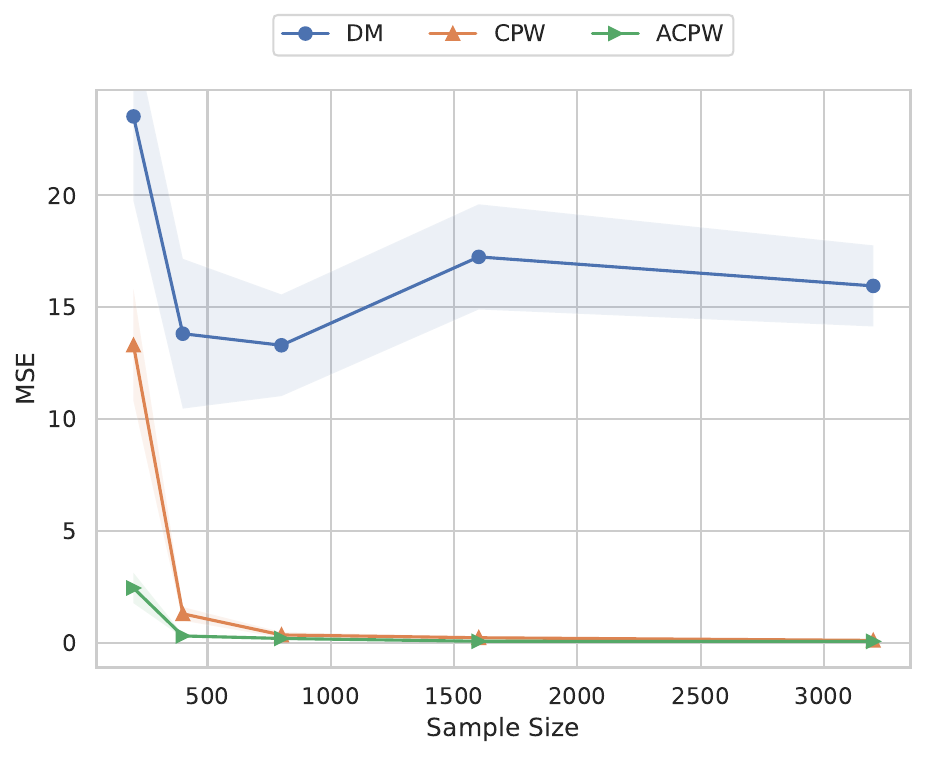}
         \caption{Target}
         \label{fig:dr_dmmis_tar}
     \end{subfigure}
     \caption{Direct Model Misspecification}
     \label{fig:dr_dm}
\end{figure}

\begin{figure}
\begin{subfigure}[b]{0.5\textwidth}
         \centering
         \includegraphics[width=\linewidth]{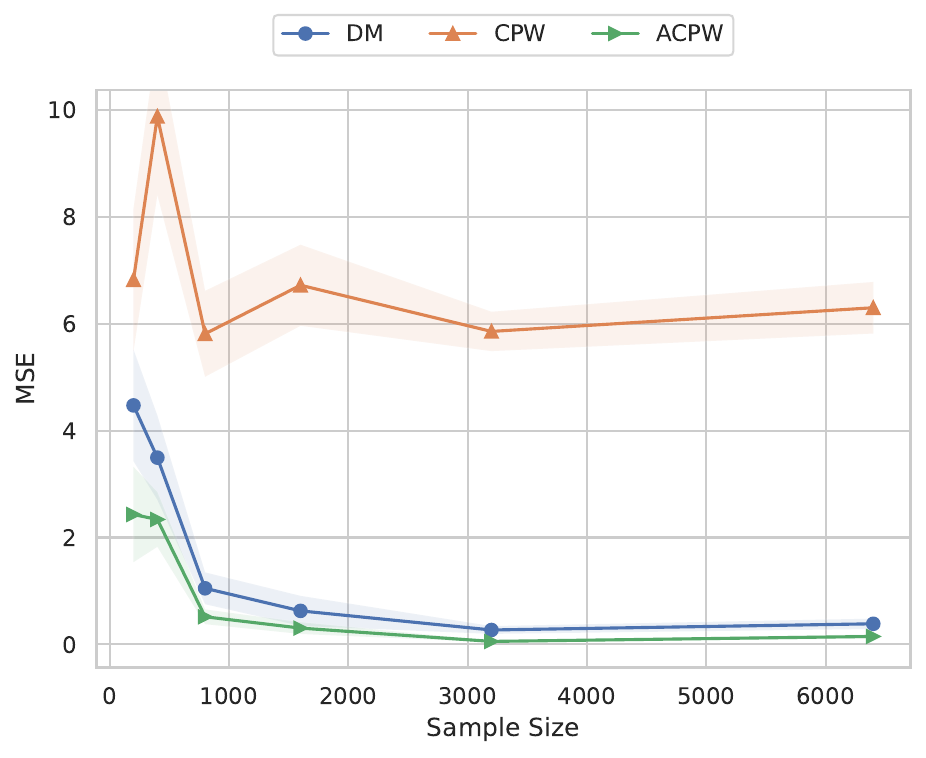}
         \caption{Behavior}
         \label{fig:dr_propmis_beh}
     \end{subfigure}
\begin{subfigure}[b]{0.5\textwidth}
         \centering
         \includegraphics[width=\linewidth]{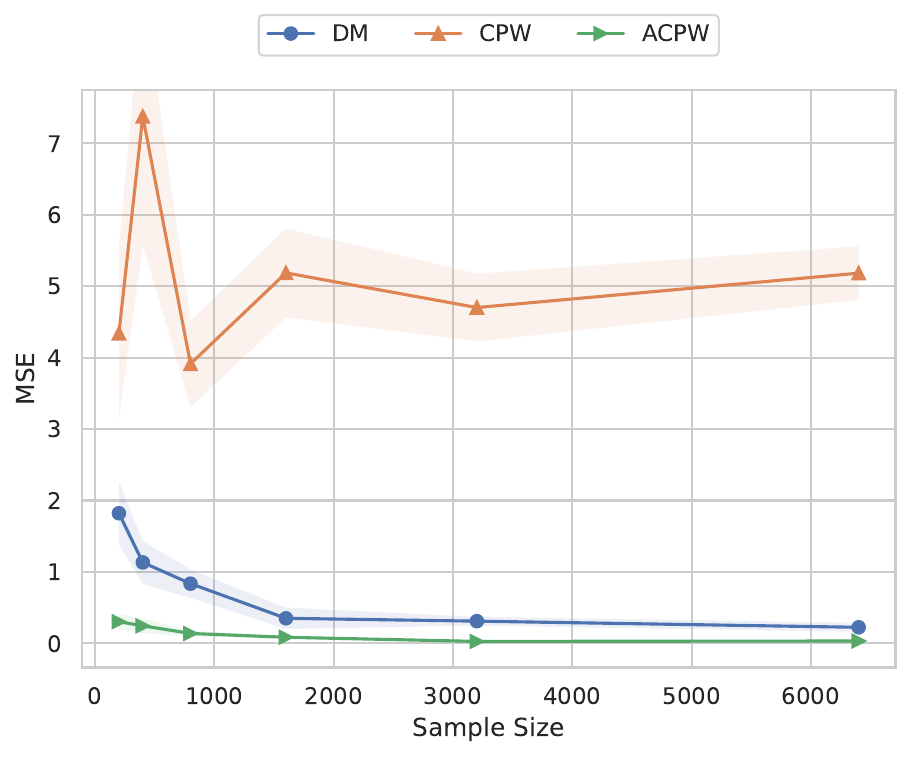}
         \caption{Target}
         \label{fig:dr_propmis_tar}
     \end{subfigure}
     \caption{Propensity Model Misspecification}
     \label{fig:dr_prop}
\end{figure}

The results are shown in Figures \ref{fig:dr_dm} and \ref{fig:dr_prop}, respectively. When the demand model is misspecified, the direct method (DM) exhibits significant, non-vanishing errors compared to the CPW and ACPW methods, whose estimation errors diminish to zero as the sample size increases. 
A similar trend is observed for DM and ACPW when the propensity scores are misspecified. CPW has a large MSE (approximately 5) that does not reduce with more samples. Compared to DM and CPW, the proposed ACPW method performs robustly across both conditions, which adds an additional layer of protection for accurate surplus identification in practice. By combining the strengths of both DM and CPW, it ensures reliable consumer surplus identification even when the underlying economic behavior or historical pricing data is complex or poorly understood.

\begin{figure}
\begin{subfigure}[b]{0.5\textwidth}
         \centering
         \includegraphics[width=\linewidth]{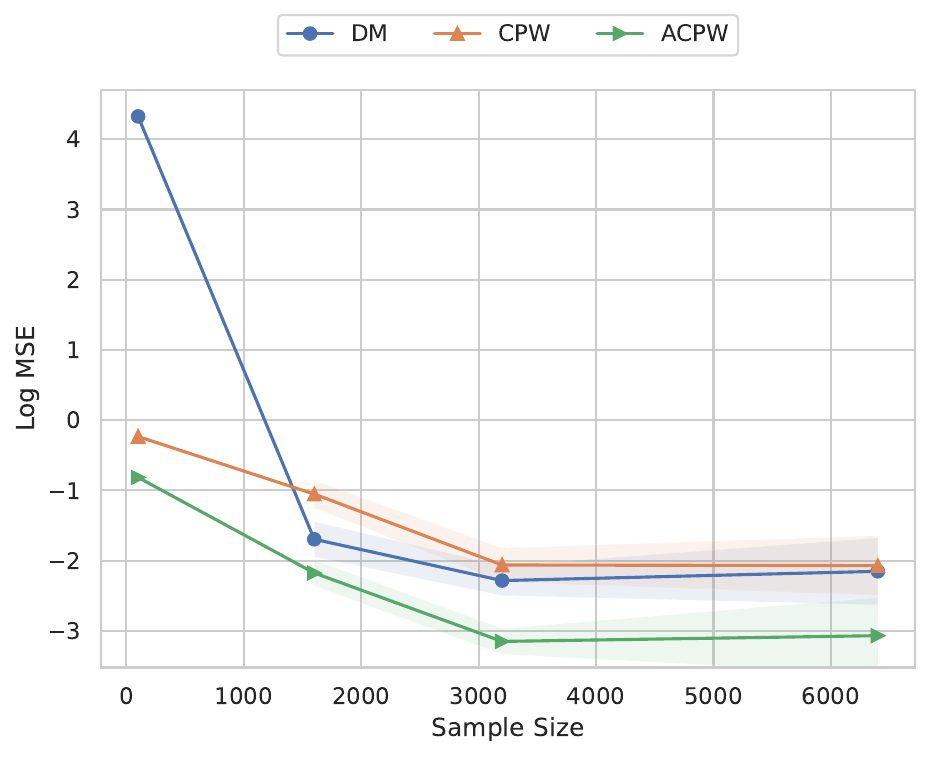}
         \caption{Behavior}
         \label{fig:dr_beh}
     \end{subfigure}
\begin{subfigure}[b]{0.5\textwidth}
         \centering
         \includegraphics[width=\linewidth]{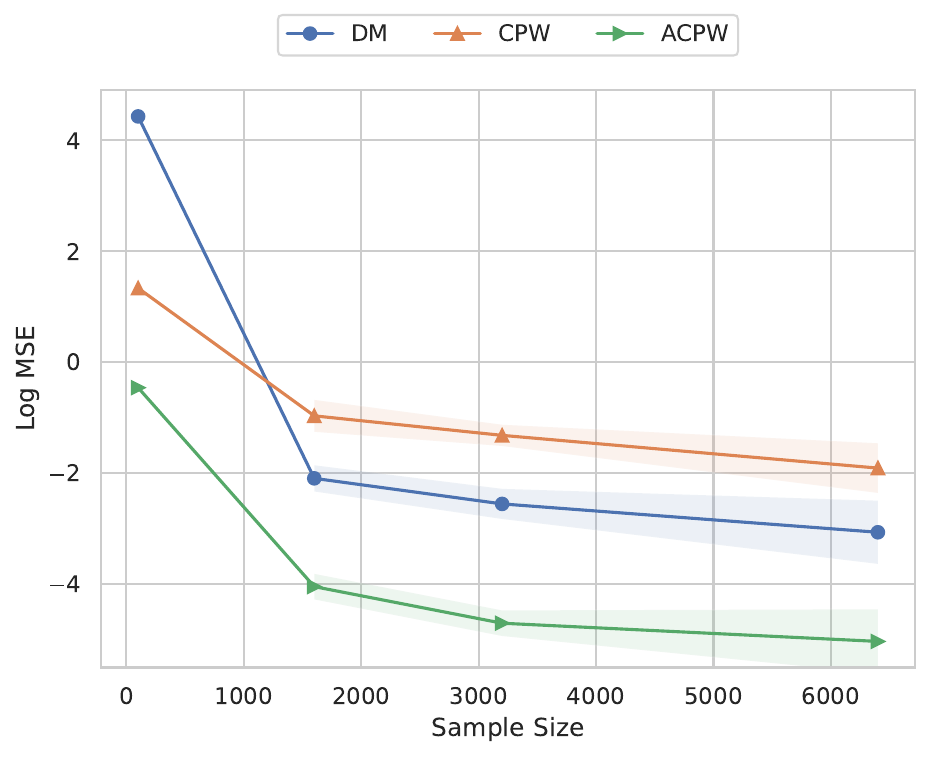}
         \caption{Target}
         \label{fig:dr_tar}
     \end{subfigure}
     \caption{Direct and propensity models are both well-specified.}
     \label{fig:convergence}
\end{figure}

\noindent \textbf{Convergence Rate:}
We assume $X\sim U[0,1], \beta \sim U[-1,1], V=100+300\beta X + \epsilon, \epsilon\sim U[0,10], P\sim U[\min(V)-0.05, \max(V)+0.05], Y=\mathbb{I}[V>P]$. We use the kernel density model with the tophat kernel. The direct model is correctly specified using a linear model. 
If both models are well-specified, ACPW has a faster convergence rate, as shown in Figure \ref{fig:convergence}. Here we plot on a log-scale for better visualization. 
ACPW provides a ``best of both worlds'' solution: it is robust to the failure of one model and achieves the fastest possible statistical convergence when models are accurate, making it a highly reliable tool for regulatory auditing.

\subsection{Inequality-Aware Surplus} \label{sec:exp_ineq}

To evaluate the performance of our inequality-aware surplus estimators, we conduct simulations under a correctly specified model setting. The data is generated as follows: the feature vector $X$ is drawn from $U[0,1]$, and a consumer's valuation $V$ is a linear function $V=100+300\beta^T X + \epsilon$, where coefficients $\beta \sim U[-1,1]$ and the error $\epsilon \sim U[0,1]$. The price $P$ is drawn from $U[9,12]$, and the purchase outcome is $Y=\mathbb{I}[V>P]$. For our estimators, the direct model is correctly specified as linear, and the propensity score model is correctly specified using a kernel density estimator with a tophat kernel. We report the results with 200 replications. 

We evaluate the estimators' Mean Squared Error (MSE) when $r=0.5$. 
Intuitively, the parameter $r$ controls the sensitivity of the surplus measure to the welfare of the worst-off consumers. As $r$ decreases, the estimator places increasingly higher weight on low-surplus outcomes. By choosing $r=0.5$, we adopt a moderate aversion to inequality. This criterion can help prioritize policies that are more beneficial to disadvantaged customer groups. 

\begin{figure}
     \begin{subfigure}[b]{0.45\textwidth}
         \centering
         \includegraphics[width=\linewidth]{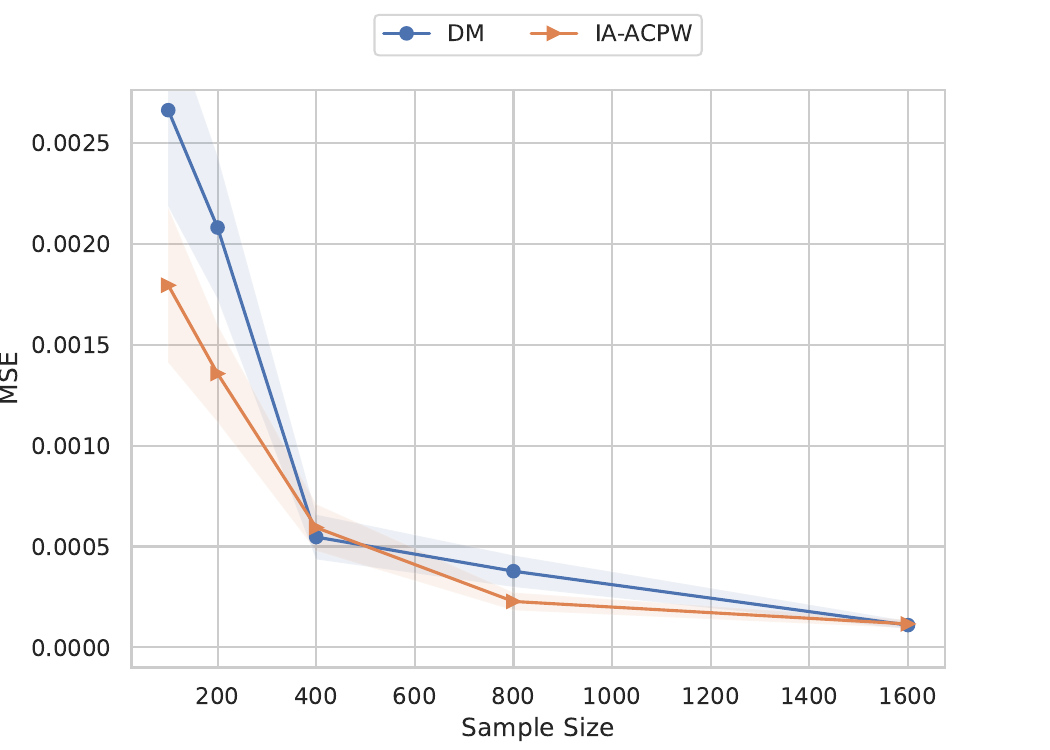}
         \caption{$r=0.5$ Behavior}
     \end{subfigure}
     \hfill
     \begin{subfigure}[b]{0.45\textwidth}
         \centering
         \includegraphics[width=\linewidth]{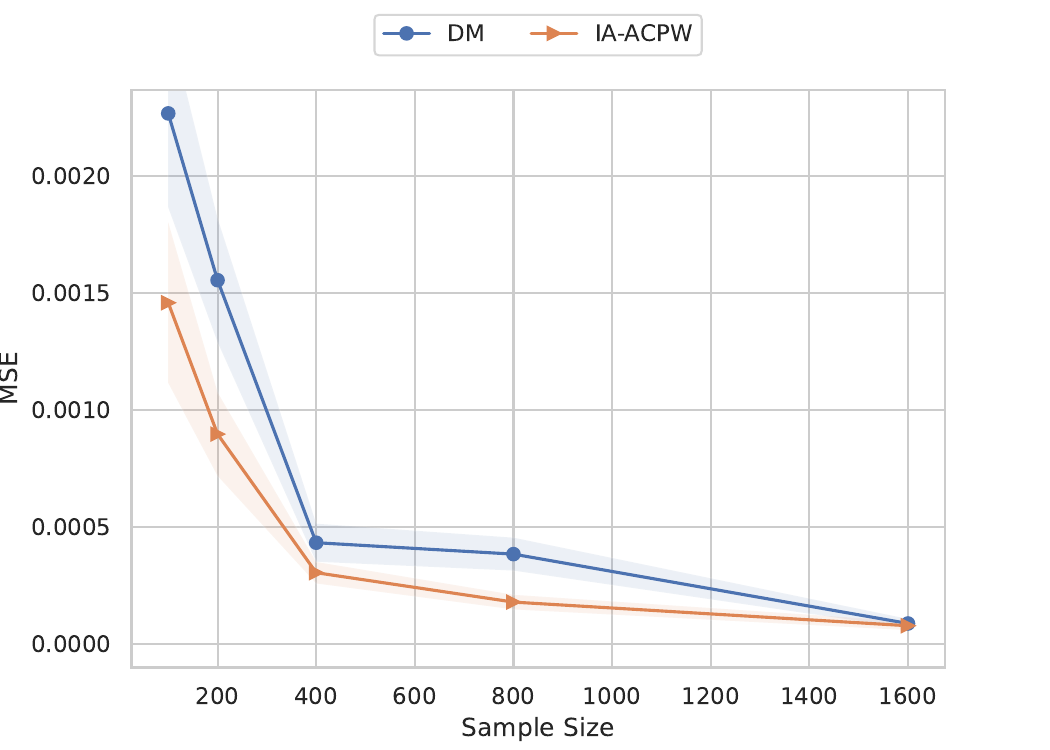}
         \caption{$r=0.5$ Target}
     \end{subfigure}
    \caption{MSE for Inequality-Aware Surplus Estimation} 
    \label{fig:ineq_0.5}
\end{figure}

We compare the IA-ACPW estimator against the standard direct method (DM) baseline. The results are presented in Figure \ref{fig:ineq_0.5}.
Both the DM and IA-ACPW estimators exhibit decreasing MSE as the sample size increases. The IA-ACPW estimator consistently achieves lower estimation error than the DM baseline. This variance reduction is driven by the influence-function-based construction of the estimator, which leverages the cumulative propensity weights. 

\subsection{Confidence Intervals} \label{sec:exp_conf}
This section empirically validates the reliability and coverage of the proposed confidence intervals for surplus estimation. A robust auditing tool must not only provide a point estimate but also a mathematically sound measure of uncertainty, allowing regulators and firms to determine if changes in welfare are statistically significant.

We generate data with a feature vector $X \sim U[0,1]$. The valuation $V$ follows a linear model, $V = 10 + 300\beta^T X + \epsilon$, where $\beta \sim U[-1,1]$ and $\epsilon \sim U[0,1]$. The price is drawn from $P \sim U[9,11]$, and the outcome is $Y = \mathbb{I}[V>P]$.

To assess the robustness of our method, we pair a propensity score model (a kernel density estimator with a tophat kernel) with a direct model (a gradient boosting tree). We test the confidence interval coverage over 200 simulation runs for sample sizes of 2000, 4000, and 8000, with a nominal coverage rate of 90\%.

\begin{figure}
     \begin{subfigure}[b]{0.45\textwidth}
         \centering
         \includegraphics[width=\linewidth]{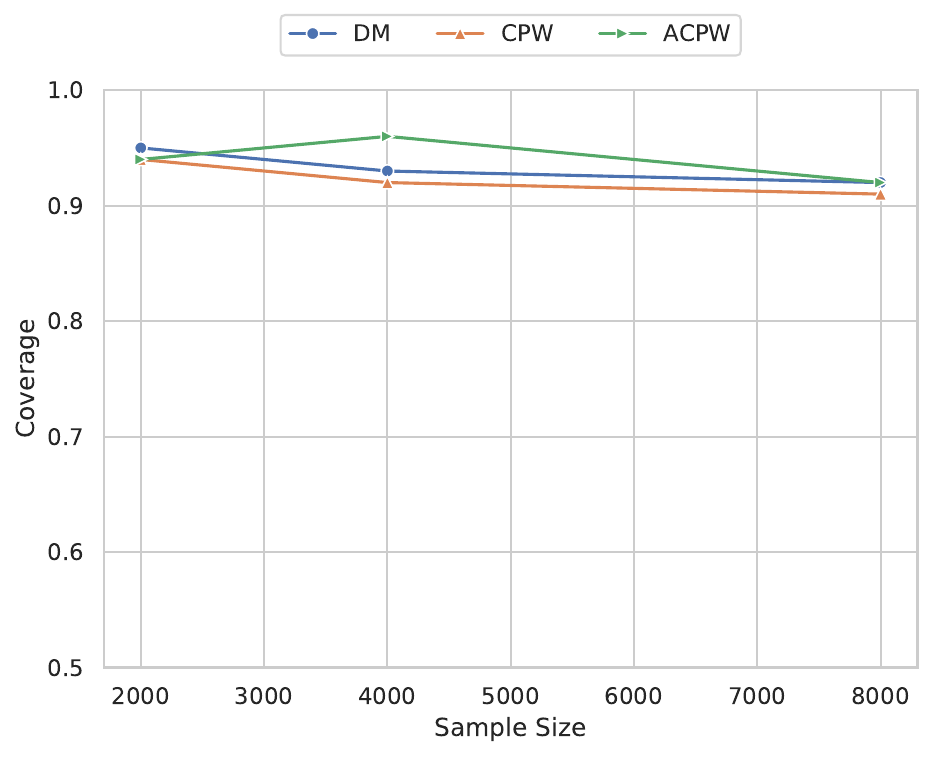}
         \caption{Behavior}
         \label{fig:beh_conf}
     \end{subfigure}
     \hfill
     \begin{subfigure}[b]{0.45\textwidth}
         \centering
         \includegraphics[width=\linewidth]{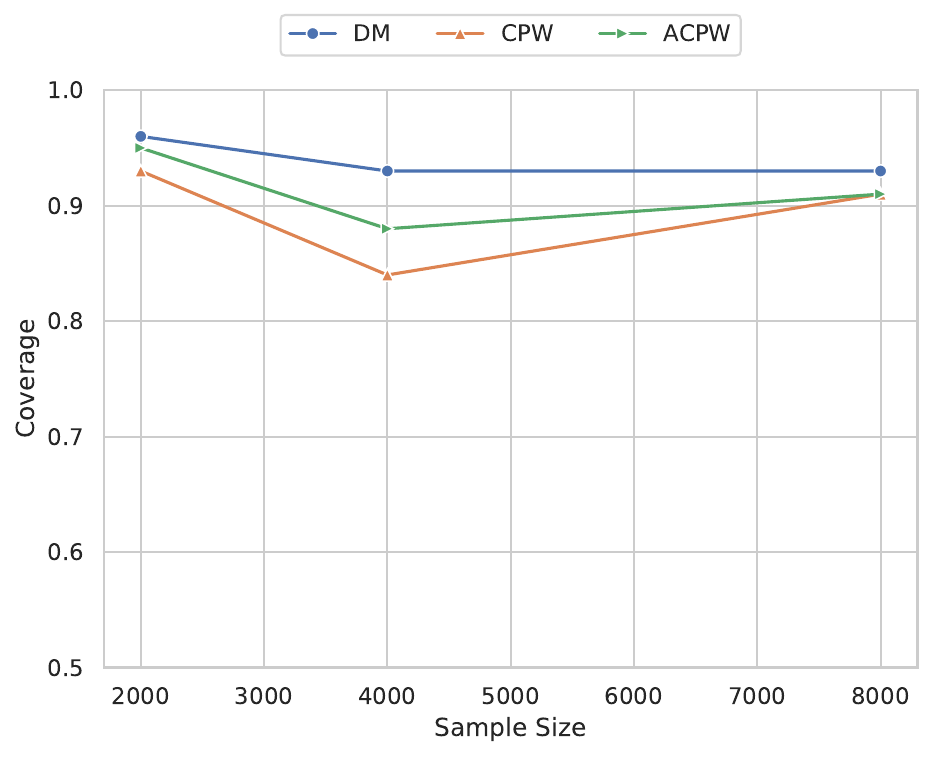}
         \caption{Target}
         \label{fig:tar_conf}
     \end{subfigure}
     \caption{Confidence Interval when $r=1$.}
     \label{fig:confr1}
\end{figure}

\begin{figure}
     \begin{subfigure}[b]{0.45\textwidth}
         \centering
         \includegraphics[width=\linewidth]{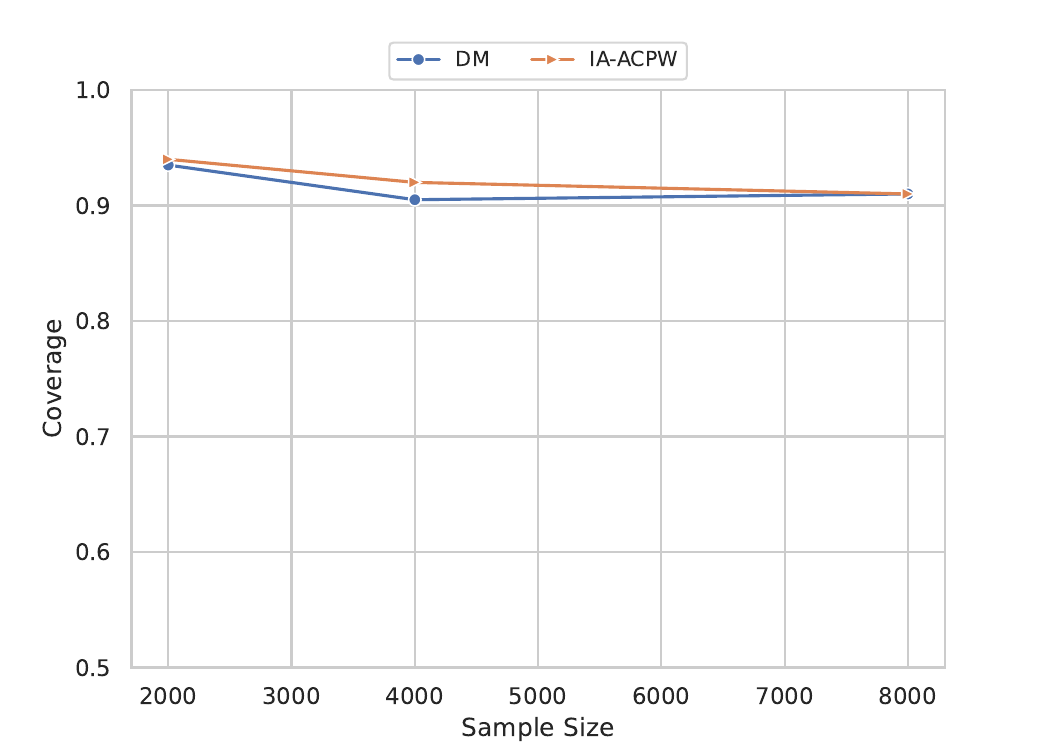}
         \caption{$r=0.5$ Behavior}
         \label{fig:halfbeh_conf}
     \end{subfigure}
     \hfill
     \begin{subfigure}[b] {0.45\textwidth}  \label{fig:conf}
         \centering
         \includegraphics[width=\linewidth]{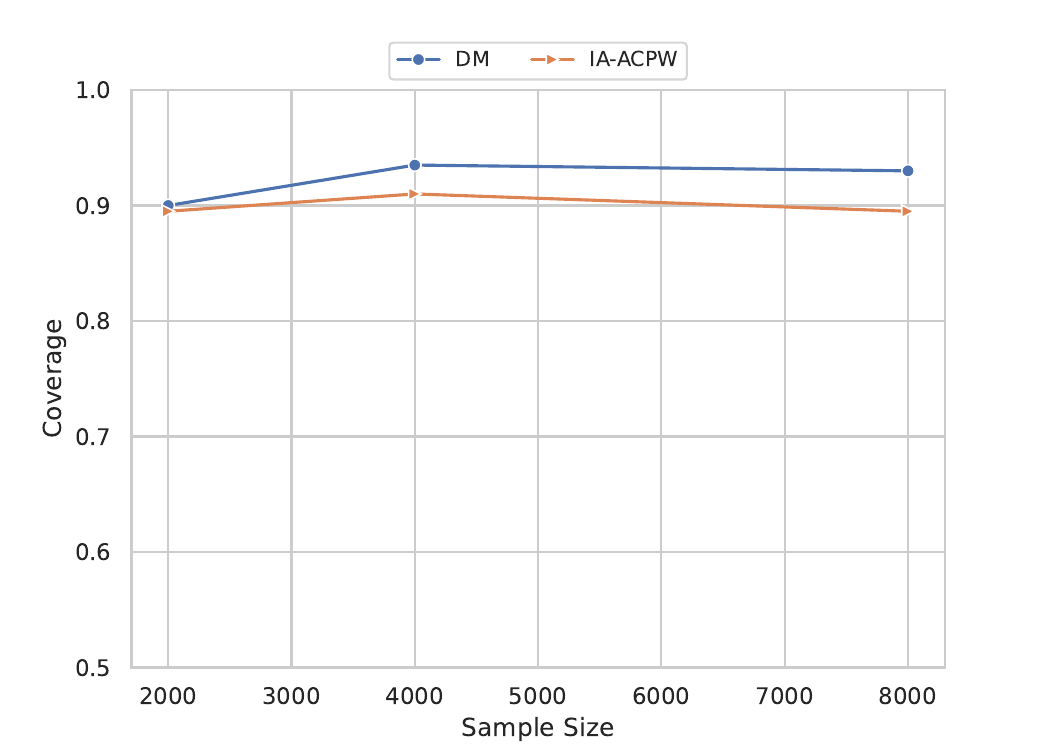}
         \caption{$r=0.5$ Target}
         \label{fig:halftar_conf}
     \end{subfigure}
     \caption{Confidence Intervals for $r=0.5$.}
     \label{fig:half_conf}
\end{figure}

The results with $r=1$ are shown in Figure \ref{fig:confr1} for the behavior and target policy. We test the confidence intervals for DM, CPW, and ACPW. The results are presented in Figure \ref{fig:half_conf} for $r=0.5$ under both a behavior and a target policy. We test the confidence intervals for DM and IA-ACPW for the inequality-aware surplus. 

Across both the Behavior and Target policies, the empirical coverage rates for the IA-ACPW estimator consistently hover around the 90\% target.
The intervals remain accurate not just for the standard average surplus ($r=1$), but also for the inequality-averse variant with $r=0.5$. 
The ability of these intervals to maintain proper coverage, even when utilizing flexible models like gradient boosting, is critical for many applications. It ensures that a firm or regulator can construct a $(1-\alpha)$ confidence interval $[ \hat{\mathcal{S}} \pm z_{1-\alpha/2}\sqrt{\hat{\Sigma}/n} ]$ and have high confidence that the true consumer welfare impact lies within that range.

\subsection{Automobile Loans}  \label{sec:exp_auto}

To demonstrate the practical utility of our framework, we apply it to a real-world dataset on personalized pricing. We utilized the CPRM-12-001: On-Line Auto Lending dataset from Columbia University, a comprehensive collection of U.S. auto loan applications from July 2002 to November 2004. 
It includes details such as application date, requested loan terms (amount and duration), applicant personal information (e.g., state, car type, FICO score), approval status, the Annual Percentage Rate (APR) offered for approved loans, and whether a contract was ultimately executed.  We remove the last 45 days before `11/16/2004' since these records may not have an accurate outcome \citep{ban2021personalized}. 

The price is computed as the net present value of future payments minus the
loan amount: 
\begin{align*}
    p = Monthly\ Payment \times \sum_{\tau=1}^{Term} (1 + Rate)^{-\tau} - Loan\ Amount, 
\end{align*}

where $Rate$ is the monthly London interbank offered rate. 
Following \cite{ban2021personalized}, we set the price range to $[0, 7500]$.

We define a feature set for each loan application. 
The set of continuous predictors includes the applicant's FICO score, the loan term in months, the approved loan amount, the one-month interest rate, and a relative competitor interest rate. These features are standardized to have a mean of zero and a standard deviation of one.
The set of categorical predictors includes loan term class, partner binary, car type, application type, customer tier, and state. These are converted into numerical representations via label encoding.
The final feature vector, denoted by $\boldsymbol{x}$, is the concatenation of the standardized continuous features and the encoded categorical features.

We segment applicants into four groups based on their credit history (Good vs. Bad Credit, proxied by FICO score) and the political affiliation of their state (Red State vs. Blue State).  
The red and blue state groups are based on the state's political affiliation in the 2000 U.S. presidential election. Blue States are: CA, CT, DC, DE, HI, IA, IL, MA, MD, ME, MI, MN, NJ, NM, NY, OR, PA, RI, VT, WA, and WI. Red States are then defined as all other states present in our final dataset. This classification allows for a comparative analysis of policy effects across distinct economic and political environments.
The price distributions for these distinct groups are visualized in Figure \ref{fig:price_dist}. While the historical behavior policy clearly assigns higher prices to the Low FICO groups, there is no discernible difference in the prices assigned to Red versus Blue states within the same credit tier.

We evaluate the surplus under a behavior policy (historical pricing from the lender) against a personalized pricing policy trained on the data to maximize the lender's revenue. 
First, we estimate the baseline inequality-aware surplus for each group using the ACPW estimator under the behavior policy, which is the historical price prescribed by the insurance company in the original dataset. Next, we show the results for the inequality aversion parameter $r=0.5$. 
Then we evaluate the impact of a personalized pricing policy trained on the historical data. The demand function is trained using a gradient boosting tree. 
The final price is then determined by applying a softmax transformation over the expected revenue, $p \cdot \mu(x, p)$, sampled from a discrete grid of five price points.
The results are shown in Table \ref{tab:autoloan_dr}.

\subsubsection{Empirical Observations}

\begin{figure}
    \centering
    \includegraphics[width=0.5\linewidth]{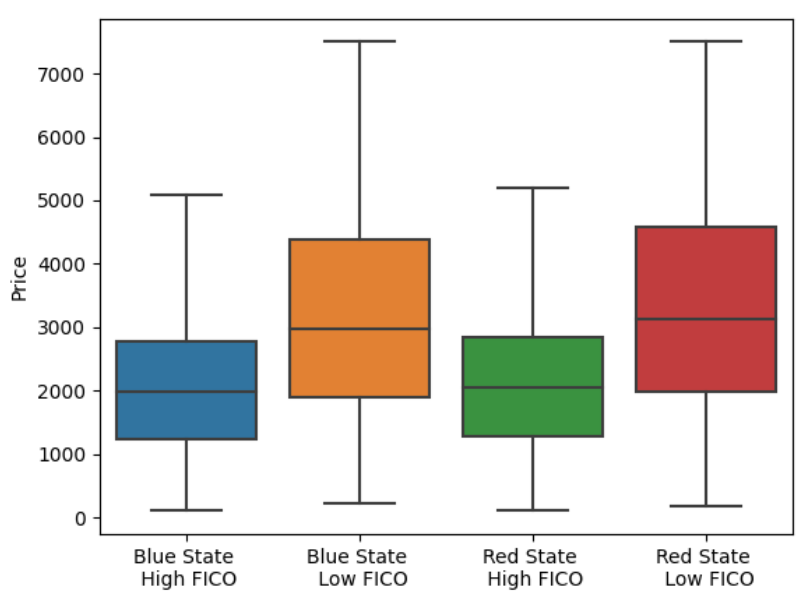}
    \caption{Price Distribution}
    \label{fig:price_dist}
\end{figure}

We apply the ACPW and IA-ACPW estimators to the auto loan dataset to quantify the welfare implications of algorithmic personalization and the trade-off between aggregate surplus and equity. Our analysis begins by examining the aggregate consumer surplus ($r=1$) and highlights disparities in consumer welfare across geographic and credit segments under the historical behavior policy.
For the good-credit segment, the estimated consumer surplus in red states is $\$1,313.00 \pm 47.40$, which is approximately 9\% higher than the $\$1,204.61 \pm 43.79$ observed in blue states. This geographic premium persists within the bad-credit segment, where surplus in red states exceeds that of blue states by more than 7\% ($\$952.59 \pm 36.12$ vs $\$888.20 \pm 36.03$). As the price distributions across these states are similar, as shown in Figure \ref{fig:price_dist}, these consistent welfare gaps are likely driven by underlying differences in the distributions of consumers' willingness to pay between regions.

We next analyze the impact of introducing a personalized pricing algoritm. Consumer surplus decreases for every demographic group, with the Red State, Good Credit group experiencing the sharpest decline of roughly 26\% (\$1,313.00 $\to$ \$965.44), representing a significant loss in aggregate surplus. However, despite this aggregate loss, the distribution of welfare becomes less unequal. Under the historical policy, the most advantaged group (Red State, Good Credit) had a surplus 1.48 times that of the least advantaged group (Blue State, Bad Credit). Under personalization, this ratio compresses to 1.30 times. This dynamic, where aggregate surplus falls but inequality reduces, mirrors the findings of \citet{dube2023personalized}, suggesting the personalized pricing can act as a progressive transfer mechanism.

To capture nuances regarding the most vulnerable consumers, we utilize the inequality-aware surplus metric ($r=0.5$). While the hierarchy between groups remains consistent with the standard surplus for the historical policy, the relative difference between the best (Red State, Good Credit) and worst (Blue State, Bad Credit) outcomes is notably smaller, with a ratio of 1.22 compared to 1.48. This indicates that the worst-off individuals within these demographic groups already experience more similar surplus levels, regardless of their group label. Furthermore, the decline in inequality-aware surplus under personalization is notably reduced (ranging from 7\% for Red State, Bad Credit to approximately 17\% for Red State, Good Credit and Blue State, Good Credit) compared to the standard surplus declines (12\% for Red State, Bad Credit 
to 26\% for Red State, Good Credit). This suggests that the surplus lost to personalization is primarily extracted from high-valuation customers, who are weighted less heavily in this metric. Consequently, with personalized pricing, the relative inequality-aware surplus difference between the best and worst surplus becomes even smaller, dropping to a ratio of just 1.11. This provides further evidence that personalized pricing functions as a leveler, disproportionately extracting value from the top to flatten the welfare distribution, thereby protecting the relative standing of those who are worst off. However, this reduction in inequality must be traded-off
against the fact that all segments have declining absolute surplus even for the inequality-aware surplus.

\begin{table}[!ht]
    \centering
    \resizebox{\textwidth}{!}{
    \begin{tabular}{ccccc}
    \toprule 
        ~ & Blue State, Good Credit & Blue State, Bad Credit & Red State, Good Credit & Red State, Bad Credit \\ \midrule
        Historical Surplus& $1204.61 \pm 43.79$ & $888.20 \pm 36.03$ & $1313.00 \pm 47.40$ & $952.59 \pm 36.12$  \\
        Surplus w/ Personalization & $892.37 \pm 35.79$ & $740.71 \pm 30.88$ & $965.44 \pm 38.99$ & $839.68 \pm 32.16$  \\ \midrule
        Historical Surplus ($r=0.5$)& $29.92 \pm 0.57$ & $25.56 \pm 0.54$ & $31.26 \pm 0.60$ & $26.66 \pm 0.56$ \\
        Surplus w/ Personalization ($r=0.5$) &  $24.57 \pm 0.41$ & $23.33 \pm 0.37$ & $25.82 \pm 0.44$ & $24.67 \pm 0.36$  \\
        \bottomrule
    \end{tabular}}
    \caption{Surplus Estimation with 90\% confidence interval by ACPW and the Surplus under a Personalization Algorithm.}
    \label{tab:autoloan_dr}
\end{table}

\section{Conclusions and Managerial Implications}
\label{sec:conclusions}

This paper proposes a data-driven way to measure consumer welfare using transactional data. Our cumulative-weights framework provides a robust mechanism to audit the economic and distributional consequences of targeted pricing strategies, even when demand estimation is prohibitively difficult.

We establish the theoretical foundations for three primary estimators: cumulative propensity weighting, a direct method plug-in, and an augmented cumulative propensity weighting variant, demonstrating that all three achieve the semi-parametric efficiency bound under different assumptions. The ACPW estimator, in particular, offers a ``best of both worlds'' solution for practitioners: it remains consistent if either the demand model or the historical pricing policy is consistent. In addition, it attains fast convergence even when flexible machine learning methods are used to estimate the nuisance functions at slower nonparametric rates. Furthermore, we extend this framework to accommodate inequality-aware surplus measures, enabling welfare assessments that prioritize disadvantaged customer segments through a tunable inequality-aversion parameter.

For pricing managers, the ACPW estimator serves as a critical internal auditing tool to de-risk algorithmic deployments. Managers can now quantify how different customer segments are affected by specific pricing policies, enabling the proactive identification of potential fairness issues or disparate impacts before they escalate into legal or reputational crises. Furthermore, integrating these methods into the policy design phase allows firms to simulate the impact of new pricing strategies on both the average surplus and its distribution. This facilitates a more balanced long-term strategy that optimizes for both profitability and the maintenance of a healthy, sustainable customer ecosystem. 

From a regulatory standpoint, this research addresses the critical gap between the growing sophistication of personalized pricing and the limited tools available for public oversight. The ACPW estimator offers a statistically defensible methodology for bodies such as the Federal Trade Commission to assess whether current pricing practices harm consumers or specific protected groups. 
By establishing credible upper and lower bounds on surplus, regulators can make evidence-based judgments even when price variation is sparse, thereby avoiding the need for firms to undertake commercially risky or unprofitable experiments. Ultimately, these cumulative-weight-based methods deliver transparent, regulator-ready tools for diagnosing the social and economic consequences of the modern algorithmic pricing landscape.

\bibliographystyle{plainnat}
\bibliography{reference}

\pagebreak

\setcounter{section}{0}
\section{Auxiliary Results}
\label{app:proof}

Throughout, we use $C$ to denote a generic constant that can vary from line to line.

\subsection{Equivalence of the Area Under the Demand Curve Surplus Representation}
\label{sec:DM_equiv_valuation_diff}

\begin{proposition}\label{prop:DM_equiv_valuation_diff} Under Assumption \ref{assumption:ignorability},
\[
\E\big[(V-P)_+\big]
=\E_X\left[\int_{p=0}^{\infty} \pi_D(p\mid X)
\int_{z=p}^{\infty} \mu(X,z)\, dz\, dp \right].
\]
\end{proposition}

\begin{proof}
This result follows from carefully changing the order of integration:
\begin{align*}
\E\big[(V-P)_+\big]
&= \E_X\left[\int_{p=0}^{\infty}
\left( \int_{v=p}^{\infty} (v-p)\, f_{V\mid X}(v\mid X)\, dv \right) \pi_D(p\mid X)\, dp \right] \\
&\overset{s=v-p}{=}\; \E_X\left[\int_{p=0}^{\infty}
\left( \int_{s=0}^{\infty} s\, f_{V\mid X}(s+p\mid X)\, ds \right) \pi_D(p\mid X)\, dp \right] \\
&=\; \E_X\left[\int_{p=0}^{\infty}
\left( \int_{s=0}^{\infty} \int_{t=0}^{s} f_{V\mid X}(s+p\mid X)\, dt\, ds \right) \pi_D(p\mid X)\, dp \right] \\
&=\; \E_X\left[\int_{p=0}^{\infty}
\left( \int_{t=0}^{\infty} \int_{s=t}^{\infty} f_{V\mid X}(s+p\mid X)\, ds\, dt \right) \pi_D(p\mid X)\, dp \right] \\
&=\; \E_X\left[\int_{p=0}^{\infty}
\left( \int_{t=0}^{\infty} \bar F_{V\mid X}(t+p\mid X)\, dt \right) \pi_D(p\mid X)\, dp \right] \\
&\overset{p+t=z}{=}\; \E_X\left[\int_{p=0}^{\infty}
\int_{z=p}^{\infty} \bar F_{V\mid X}(z\mid X)\, \pi_D(p\mid X)\, dz\, dp \right] \\
&=\; \E_X\left[\int_{p=0}^{\infty} \pi_D(p\mid X)
\int_{z=p}^{\infty} \mu(X,z)\, dz\, dp \right].
\end{align*}

The last equality uses the identity
\[
\mu(X,z)=\E[Y\mid X,P=z]=\mathbb{P}[Y=1\mid X,P=z]
=\mathbb{P}[V>z\mid X,P=z].
\]

By Assumption 1 ($V \perp P \mid X$), we have $\mathbb{P}[V > z \mid X, P=z] = \mathbb{P}[V > z \mid X] = \bar{F}_{V \mid X}(z \mid X)$, which completes the proof.
\end{proof}

\subsection{Auxiliary Lemmas}
\label{sec:lemma_banlancing}

\begin{lemma} [Balancing Property] \label{lemma: balancing}
 For any measurable function $\phi(X,P)$, the following holds 
    \begin{gather} \label{eq:balance}
       \E\left[ \int_0^\infty\pi(p | X) \int_{p}^\infty \phi(X,a)dadp  \right] = \E\left[\frac{F^\pi(P | X)}{\pi_D(P | X)}\phi(X,P) \right]
    \end{gather}
\end{lemma}

\begin{proof}
    \begin{gather*}
         \E\left[ \int_0^\infty\pi(p | X) \int_{p}^\infty \phi(X,a)dadp  \right]= \E\left[ \int_0^\infty  \int_0^a \pi(p|X) dp \; \phi(X,a)da \right] \mbox{ (exchanging the integral order) } \\
         = \E\left[ \int_0^\infty  F^\pi(p | X) \; \phi(X,p)dp \right]=\E\left[ \int_0^\infty  \frac{F^\pi(p | X) }{\pi_D(p|X)}  \pi_D(p|X) \phi(X,p)dp \right]= \E\left[\frac{F^\pi(P | X)}{\pi_D(P| X)}\phi(X,P) \right].
    \end{gather*}
This completes the proof.

\end{proof}

\begin{lemma}\label{lem:error transform} Assume there exists a constant $c$ such that $\pi_D(p|x)>c$, for all $p$ and $x$, then 
\[
\E\!\left[(\widehat h(X)-h(X))^2\right]
\;\lesssim\;
\E\!\left[(\widehat \mu(X,P)-\mu(X,P))^2\right],
\]
where
\[
h(x)= \int F^\pi(p \mid x)\,\mu(x, p)\,dp,
\qquad
\widehat h(x)= \int F^\pi(p \mid x)\,\widehat \mu(x, p)\,dp .
\]
\end{lemma}

\begin{proof}
We have
\begin{align*}
\E[(\widehat h(X)-h(X))^2]
&= \E\left[\left(\int F^\pi(p \mid X)\bigl(\widehat \mu(X,p)-\mu(X,p)\bigr)\,dp\right)^2\right].
\end{align*}
Using the identity
\(
F^\pi(p\mid X)
= \frac{F^\pi(p\mid X)}{\pi_D(p\mid X)}\pi_D(p\mid X),
\)
we obtain
\begin{align*}
\E[(\widehat h(X)-h(X))^2]&= \E\left[\left(\int \frac{F^\pi(p\mid X)}{\pi_D(p\mid X)}\,
\pi_D(p\mid X)\bigl(\widehat \mu(X,p)-\mu(X,p)\bigr)\,dp\right)^2\right] \\
&= \E\left[\left(\E_{P\mid X}\!\left[\frac{F^\pi(P\mid X)}{\pi_D(P\mid X)}\,
(\widehat \mu(X,P)-\mu(X,P))\right]\right)^2\right].
\end{align*}
Applying Jensen’s inequality to the conditional expectation yields
\begin{align*}
\E[(\widehat h(X)-h(X))^2] &\le \E\left[\E_{P\mid X}\!\left[\left(\frac{F^\pi(P\mid X)}{\pi_D(P\mid X)}\right)^2
(\widehat \mu(X,P)-\mu(X,P))^2\right]\right].
\end{align*}
By the strong overlap condition,
\(
\frac{F^\pi(P\mid X)}{\pi_D(P\mid X)} \le 1/c,
\)
and hence
\begin{align*}
\E[(\widehat h(X)-h(X))^2]&\le 1/c^2\,\E\left[(\widehat \mu(X,P)-\mu(X,P))^2\right].
\end{align*}
Absorbing the constant into the \(\lesssim\) notation completes the proof.
\end{proof}

\section{Proof of Theorem \ref{thm:eif} : Deriving EIFs for Consumer Surplus (r=1)}
\label{sec:eif_proof}

Following \citet{tsiatis2006semiparametric}, to derive the EIF, we first find the tangent space and its orthogonal complement. We begin with the following auxiliary Lemmas for the corresponding results.

\begin{lemma}
\label{thm:orthogonality_proof}
    The orthogonal complement of the observed tangent space is \begin{align*}
      \Lambda^\perp  = \{ M(X,P,Y):   E(M|P,X)=\E(M|V,X)=0 \},
    \end{align*} any measurable function $M(X,P,Y).$
\end{lemma}

\begin{proof}
We start by characterizing the full data tangent space. It can be shown that the full data tangent space $\Lambda^F$ is
$\Lambda_X^F\bigoplus\Lambda_V^F\bigoplus\Lambda_P^F$, where $\bigoplus$ is the direct sum,
\begin{gather*}
    \Lambda_X^F=\{g_X(X): \E(g_X(X))=0 \},\\
    \Lambda_P^F=\{g_P(P,X): \E(g_P(P,X)|X)=0 \},  \\
   \mbox{and } \Lambda_V^F=\{g_V(V,X): \E(g_V(V,X)|X)=0 \}.
\end{gather*} 

By Theorem 7.1 in \citet{tsiatis2006semiparametric}, the observed tangent space $\Lambda$ is $\E(\Lambda^F| P,X,Y)$. Hence, an element in $\Lambda$ can be written as \begin{align*}
    g_X(X)+g_P(P,X)+\E(g_V(V,X) | Y,P,X),
\end{align*} for $ g_X(X) \in \Lambda_X^F, \; g_P(P,X) \in \Lambda_P^F, \mbox{ and } g_V(V,X) \in \Lambda_V^F,$

Denote $\Delta = \{ M(X,P,Y):   E(M|P,X)=\E(M|V,X)=0 \}$, the aim is to show that $\Delta=\Lambda^\perp$.
We first show that $\Delta \subseteq \Lambda^\perp$. For any $M(X,P,Y) \in \Delta$, $ g_X(X) \in \Lambda_X^F, \; g_P(P,X) \in \Lambda_P^F, \mbox{ and } g_V(V,X) \in \Lambda_V^F,$ \begin{align*}
    & \E\left[  M(X,P,Y) \left( g_X(X)+g_P(P,X)+\E(g_V(V,X) | Y,P,X) \right )         \right] \\
    = & \E\left(  M(X,P,Y)  g_X(X)  \right) +\E\left(  M(X,P,Y)  g_P(P,X)  \right)+
    \E\left[  M(X,P,Y) \E(g_V(V,X) | Y,P,X)         \right]\\
    = & \E(g_X(X) \underbrace{\E (M|X)}_0)+ \E\left(  \underbrace{\E(M|P,X)}_0  g_P(P,X)  \right) +\E\left(  \underbrace{\E(M|V,X)}_0  g_V(V,X)  \right)\\
    = & 0. 
\end{align*} Thus, $\Delta \subseteq \Lambda^\perp$.

We next show $\Lambda^\perp \subseteq \Delta.  $ For any function $h(X,P,Y) \in \Lambda^\perp$, let $g_X(X)=\E(h|X)$, $g_P(P,X)=0$, and $g_V(V,X)=0$. It can be verified that $g_X \in \Lambda_X^F$, $g_P \in \Lambda_P^F$, $g_V \in \Lambda_V^F$. Since $h(X,P,Y) \in \Lambda^\perp$, we have $\E(\E^2(h|X))=0$ if and only if $\E(h|X)=0$.

Now let $g_X(X)=\E(h|X)$, $g_P(P,X)=\E(h|P,X)$, and $g_V(V,X)=0$. It can be verified that such a construction also meets the condition that $g_X \in \Lambda_X^F$, $g_P \in \Lambda_P^F$, $g_V \in \Lambda_V^F$. Since $h(X,P,Y) \in \Lambda^\perp$. Since $h(X,P,Y) \in \Lambda^\perp$, we have $\E(\E^2(h|P,X))=0$ if and only if $\E(h|P, X)=0$.

Finally, let $g_X(X)=\E(h|X)$, $g_P(P,X)=\E(h|P,X)$, and $g_V(V,X)=\E(h|V,X)$. Using a similarly argument, we have $\E(h|V, X)=0$.

Thus,  $\Lambda^\perp  = \{ M(X,P,Y):   \E(M|P,X)=\E(M|V,X)=0 \}.$ This completes the proof.

\end{proof}

\begin{lemma}
\label{thm:diff_spaces}
    Any element in the space $\{M(P,Y,X): \E(M|V,X)=0 \}$ can be written as $h(P,X)-E(h|X)$, meaning that the space consists of functions that depend only on $P$ and $X$.
\end{lemma}

\begin{proof}
Since $Y$ is binary, any function $M(P,Y,X)$ can be written as $YM_1(P,X)+(1-Y)M_0(P,X)=Y(M_1-M_0)+M_0$.

For any $M \in \{M(P,Y,X): \E(M|V,X)=0 \}$, we have \begin{align*}
    & \E(Y(M_1-M_0)+M_0 | V,X)=0, \\
    \implies & \E(Y(M_1-M_0) |V,X)=\E(-M_0 | V,X).
\end{align*}

Since $A \independent V |X $, we have $\E(M_0 (P,X)|V,X)=\E(M_0|X)$. Hence, we have 
\begin{align*}
    \E(Y(M_1-M_0) |V,X)=\E(-M_0 | X).
\end{align*} In the above equation, the LHS is a function of $V$ and $X$, while the RHS is only a function of $X$. Therefore, the above equation holds if and only if $M_1=M_0$, and $\E(M_0(P,X)|X)=0$. This completes the proof.

\end{proof}

\begin{lemma} \label{lemma: tangent}
     The observed tangent space is the entire Hilbert space.
\end{lemma}
   
\begin{proof}
    Since all the elements in the space $\{M(P,Y,X): \E(M|V,X)=0 \}$ are only a function of $P,X$.  We have \begin{gather*}
        \{M(P,Y,X): \E(M|V,X)=0 \}  \bigcap \{M(P,Y,X): \E(M|P,X)=0 \}  =\{0\}.
    \end{gather*}

Thus the orthogonal complement of the observed tangent space is $\{0\}$, which means that the 
observed tangent space is the entire Hilbert space.
    
\end{proof}

\begin{proof} Using Lemma \ref{thm:orthogonality_proof}, \ref{thm:diff_spaces}, and \ref{lemma: tangent}, the tangent space is the entire Hilbert space. Correspondigly, there exists a unique influence function. Moreover, it is the efficient influence function.

Next, we show an equivalent form of consumer surplus that is more convenient to use, starting with the  definition from Equation \eqref{eqn: identification} :
\begin{align*}
& \EE\!\left[ \int_{p=0}^\infty \pi(p\mid X) \int_{z=p}^\infty \EE[Y \mid X,P=z] \, dz \, dp \right]
= \EE\!\left[ \int_{p=0}^\infty \int_{z=p}^\infty \pi(p\mid X)\,\mu(X,z) \, dz \, dp \right] \\
= & \; \EE\!\left[ \int_{z=0}^\infty \int_{p=0}^z \pi(p\mid X)\,\mu(X,z) \, dp \, dz \right] = \EE\!\left[ \int_{z=0}^\infty \Big(\int_{p=0}^z \pi(p\mid X)\,dp\Big)\,\mu(X,z) \, dz \right] \\
= & \;\EE\!\left[ \int_{z=0}^\infty F^\pi(z\mid X) \,\mu(X,z) dz \right],  
\end{align*}

where $\mu(X, z) = \mathbb{E}[Y|X, P=z]$ and $F^\pi(z|X) = \int_0^z \pi(s|X) ds$. We study two cases: (i) where we are evaluating a target pricing policy $\pi$, in which case $F^\pi(z|X)$ is a fixed and known weight function, and (ii) we are evaluating an unknown behavior pricing policy $\pi_D$, in which case $F^{\pi_D}(z|X)$ is also a function of the data.

\textit{Case 1: known target pricing policy $\pi$}

We define the estimand of interest we will work with:
$$\Psi(\mathcal{P}) \equiv \mathcal{S}(\pi) = \mathbb{E}_X \left[ \int_{z=0}^{\infty} \mu(X, z) F^\pi(z|X) dz \right]$$
 This is a function of the true observed distribution $\mathcal{P}$. We follow the approach to finding the efficient influence function (EIF) and notation from \cite{kennedy2024semiparametric}.  In particular, we follow the pathwise differentiability approach, defining a parametric submodel $\mathcal{P}_t = (1-t)\mathcal{P} + t\delta_{\tilde{x}, \tilde{y}, \tilde{p}}$, where $\delta$ is a point mass at the observed data $\{\tilde{x}, \tilde{y}, \tilde{p}\}$. Under this parametric model, a generic joint distribution is $f_{X,Y,P,t}(x,y,p)=(1-t)f_{X,Y,P}(x,y,p)+ t\delta_{\tilde{x}, \tilde{y}, \tilde{p}} $

The pathwise derivative is given by:
$$ \frac{d\Psi(\mathcal{P}_t)}{dt}\Big|_{t=0} = \frac{d}{dt} \int f_{X,t}(x) \int_{z=0}^{\infty} \mu_t(x, z) F^\pi(z|x) dz dx \Big|_{t=0} $$

First, we show an identity for the derivative of a generic conditional density, $\frac{\partial f_t(a|x)}{\partial t}\big|_{t=0} = \frac{\delta_{\tilde{x}}(x)}{f_X(x)} (\delta_{\tilde{a}}(a) - f(a|x))$.
Recall that the conditional density is defined as $f_t(a|x) = \frac{f_t(x,a)}{f_t(x)}$. The pathwise derivative of the marginal density is $\frac{\partial f_t(x)}{\partial t}\big|_{t=0} = \delta_{\tilde{x}}(x) - f_X(x)$. The pathwise derivative of the joint density is $\frac{\partial f_t(x,a)}{\partial t}\big|_{t=0} = \delta_{\tilde{x}}(x)\delta_{\tilde{a}}(a) - f(x,a)$.
Applying the quotient rule at $t=0$:
\begin{align*}
\frac{\partial f_t(a|x)}{\partial t}\Big|_{t=0} &= \frac{f_X(x) \frac{\partial f_t(x,a)}{\partial t} - f(x,a) \frac{\partial f_t(x)}{\partial t}}{f_X(x)^2} \\
&= \frac{f_X(x)[\delta_{\tilde{x}}(x)\delta_{\tilde{a}}(a) - f(x,a)] - f(x,a)[\delta_{\tilde{x}}(x) - f_X(x)]}{f_X(x)^2} \\
&= \frac{\delta_{\tilde{x}}(x) f_X(x) \delta_{\tilde{a}}(a) - f(x,a)\delta_{\tilde{x}}(x)}{f_X(x)^2} \\
&= \frac{\delta_{\tilde{x}}(x)}{f_X(x)} \left( \delta_{\tilde{a}}(a) - \frac{f(x,a)}{f_X(x)} \right) \\
&= \frac{\delta_{\tilde{x}}(x)}{f_X(x)} (\delta_{\tilde{a}}(a) - f(a|x))
\end{align*}

Using the product rule and this identity the derivative decomposes into two primary components:
\begin{align*}
\frac{d\Psi(\mathcal{P}_t)}{dt}\Big|_{t=0} &= \int \left[ \int_{0}^{\infty} \mu(x, z) F^\pi(z|x) dz \right] (\delta_{\tilde{x}}(x) - f_X(x)) dx \\
&\quad + \mathbb{E}_X \left[ \int_{0}^{\infty} \frac{\partial \mu_t(X, z)}{\partial t}\Big|_{t=0} F^\pi(z|X) dz \right]
\end{align*}

Let $h(X) = \int_{0}^{\infty} \mu(X, z) F^\pi(z|X) dz$. Evaluating the first integral yields $h(\tilde{x}) - \Psi(\mathcal{P})$. For the second term, we apply the quotient rule to $\mu(X, z) = \frac{f_{Y,P}(1,z|X)}{\pi_D(z|X)}$, where this equality follows from Bayes (and the definition $f_{Y,P}(1,z|X)=\mathbb{P}(Y=1, P=z|X))$:
\begin{align*}
\frac{d\Psi(\mathcal{P}_t)}{dt}\Big|_{t=0} &= h(\tilde{x}) - \Psi(\mathcal{P}) \\
&\quad + \mathbb{E}_X \Bigg[ \frac{\delta_{\tilde{x}}(X)}{f_X(X)} \int_{0}^{\infty} \frac{F^\pi(z|X)}{\pi_D(z|X)} (\delta_{\tilde{y}, \tilde{p}}(1, z|X) - f_{Y,P}(1, z|X)) dz \\
&\quad - \frac{\delta_{\tilde{x}}(X)}{f_X(X)} \int_{0}^{\infty} \frac{f_{Y,P}(1, z|X) F^\pi(z|X)}{\pi_D(z|X)^2} (\delta_{\tilde{p}}(z|X) - \pi_D(z|X)) dz \Bigg]
\end{align*}

Evaluating the inner integrals at the point mass $(\tilde{y}, \tilde{p})$ yields:
\begin{align*}
\frac{d\Psi(\mathcal{P}_t)}{dt}\Big|_{t=0} &= h(\tilde{x}) - \Psi(\mathcal{P}) + \frac{F^\pi(\tilde{p}|\tilde{x})}{\pi_D(\tilde{p}|\tilde{x})} \tilde{y} - \int_0^\infty \frac{f_{Y,P}(1, z|\tilde{x})}{\pi_D(z|\tilde{x})} F^\pi(z|\tilde{x}) dz \\
&\quad - \frac{\mu(\tilde{x}, \tilde{p}) F^\pi(\tilde{p}|\tilde{x})}{\pi_D(\tilde{p}|\tilde{x})} + \int_0^\infty \mu(\tilde{x}, z) F^\pi(z|\tilde{x}) dz \\
&= h(\tilde{x}) - \Psi(\mathcal{P}) + \frac{F^\pi(\tilde{p}|\tilde{x})}{\pi_D(\tilde{p}|\tilde{x})} (\tilde{y} - \mu(\tilde{x}, \tilde{p})) - h(\tilde{x}) + h(\tilde{x})
\end{align*}
 For the specific functional provided, the EIF is:
\begin{align*}
\frac{F^\pi(P|X)}{\pi_D(P|X)} (Y - \mu(X, P)) + h(X) - \Psi(\mathcal{P})
\end{align*}

\textit{Case 2: unknown behavior pricing policy $\pi_D$}

This case is similar to above, but with an extra term corresponding to the unknown $F^{\pi_D}(z|X) = \int_0^z \pi_D(s|X) ds$ which is now a functional of $\mathcal{P}$. 

The estimand of interest is defined as:
$$ \Psi(\mathcal{P}) \equiv \mathcal{S}(\pi_D) = \mathbb{E}_X \left[ \int_{z=0}^{\infty} \mu(X, z) F^{\pi_D}(z|X) dz \right] $$

The pathwise derivative is given by:
$$ \frac{d\Psi(\mathcal{P}_t)}{dt}\Big|_{t=0} = \frac{d}{dt} \int f_{X,t}(x) \int_{z=0}^{\infty} \mu_t(x, z) F_{t}^{\pi_D}(z|x) dz dx \Big|_{t=0} $$

Using the product rule and the identities for the derivative of a conditional density and a cumulative distribution, the derivative decomposes into three primary components:
\begin{align*}
\frac{d\Psi(\mathcal{P}_t)}{dt}\Big|_{t=0} &= \int \left[ \int_{0}^{\infty} \mu(x, z) F^{\pi_D}(z|x) dz \right] (\delta_{\tilde{x}}(x) - f_X(x)) dx \\
&\quad + \mathbb{E}_X \left[ \int_{0}^{\infty} \frac{\partial \mu_t(X, z)}{\partial t}\Big|_{t=0} F^{\pi_D}(z|X) dz \right] \\
&\quad + \mathbb{E}_X \left[ \int_{0}^{\infty} \mu(X, z) \frac{\partial F_{t}^{\pi_D}(z|X)}{\partial t}\Big|_{t=0} dz \right]
\end{align*}

Where the third term is in addition to the terms in case (i), and which we will focus on. Recall that the conditional cumulative distribution is defined as $F_t^{\pi_D}(z|x) = \frac{H_t(x,z)}{f_t(x)}$, where $H_t(x,z) = \int_{-\infty}^z f_t(x, p) dp$ is the joint cumulative mass.
The pathwise derivative of the denominator is $\frac{\partial f_t(x)}{\partial t} = \delta_{\tilde{x}}(x) - f_X(x)$.
The pathwise derivative of the numerator is $\frac{\partial H_t(x,z)}{\partial t} = \delta_{\tilde{x}}(x)\mathbb{I}(\tilde{p} \le z) - H(x,z)$.
Applying the quotient rule at $t=0$:
\begin{align*}
\frac{\partial F_t^{\pi_D}(z|x)}{\partial t} &= \frac{f_X(x) \frac{\partial H_t}{\partial t} - H(x,z) \frac{\partial f_t}{\partial t}}{f_X(x)^2} \\
&= \frac{f_X(x)[\delta_{\tilde{x}}(x)\mathbb{I}(\tilde{p} \le z) - H] - H[\delta_{\tilde{x}}(x) - f_X(x)]}{f_X(x)^2} \\
&= \frac{\delta_{\tilde{x}}(x) f_X(x) \mathbb{I}(\tilde{p} \le z) - H \delta_{\tilde{x}}(x)}{f_X(x)^2} \\
&= \frac{\delta_{\tilde{x}}(x)}{f_X(x)} \left( \mathbb{I}(\tilde{p} \le z) - \frac{H(x,z)}{f_X(x)} \right) \\
&= \frac{\delta_{\tilde{x}}(x)}{f_X(x)} (\mathbb{I}(\tilde{p} \le z) - F^{\pi_D}(z|x))
\end{align*}

Let $h_D(X) = \int_{0}^{\infty} \mu(X, z) F^{\pi_D}(z|X) dz$. Evaluating the first integral yields $h_D(\tilde{x}) - \Psi(\mathcal{P})$, while for the second term, we apply the quotient rule to $\mu(X, z)$, as before. For the third term, we use the identity $\frac{\partial F_{t}^{\pi_D}(z|X)}{\partial t}\big|_{t=0} = \frac{\delta_{\tilde{x}}(X)}{f_X(X)}(\mathbb{I}(\tilde{p} \le z) - F^{\pi_D}(z|X))$:
\begin{align*}
\frac{d\Psi(\mathcal{P}_t)}{dt}\Big|_{t=0} &= h_D(\tilde{x}) - \Psi(\mathcal{P}) \\
&\quad + \mathbb{E}_X \Bigg[ \frac{\delta_{\tilde{x}}(X)}{f_X(X)} \int_{0}^{\infty} \frac{F^{\pi_D}(z|X)}{\pi_D(z|X)} (\delta_{\tilde{y}, \tilde{p}}(1, z|X) - f_{Y,P}(1, z|X)) dz \\
&\quad - \frac{\delta_{\tilde{x}}(X)}{f_X(X)} \int_{0}^{\infty} \frac{f_{Y,P}(1, z|X) F^{\pi_D}(z|X)}{\pi_D(z|X)^2} (\delta_{\tilde{p}}(z|X) - \pi_D(z|X)) dz \Bigg] \\
&\quad + \mathbb{E}_X \Bigg[ \frac{\delta_{\tilde{x}}(X)}{f_X(X)} \int_{0}^{\infty} \mu(X, z) (\mathbb{I}(\tilde{p} \le z) - F^{\pi_D}(z|X)) dz \Bigg]
\end{align*}

Evaluating the inner integrals at the point mass $(\tilde{y}, \tilde{p})$ yields:
\begin{align*}
\frac{d\Psi(\mathcal{P}_t)}{dt}\Big|_{t=0} &= h_D(\tilde{x}) - \Psi(\mathcal{P}) + \frac{F^{\pi_D}(\tilde{p}|\tilde{x})}{\pi_D(\tilde{p}|\tilde{x})} \tilde{y} - \int_0^\infty \frac{f_{Y,P}(1, z|\tilde{x})}{\pi_D(z|\tilde{x})} F^{\pi_D}(z|\tilde{x})- \frac{\mu(\tilde{x}, \tilde{p}) F^{\pi_D}(\tilde{p}|\tilde{x})}{\pi_D(\tilde{p}|\tilde{x})}  dz \\
&\quad  - \int_0^\infty \mu(\tilde{x}, z) \mathbb{I}(\tilde{p} \le z) dz+ \int_0^\infty \mu(\tilde{x}, z) F^{\pi_D}(z|\tilde{x}) dz -\int_0^\infty \mu(\tilde{x}, z) F^{\pi_D}(z|\tilde{x}) dz\\
&= h_D(\tilde{x}) - \Psi(\mathcal{P}) + \frac{F^{\pi_D}(\tilde{p}|\tilde{x})}{\pi_D(\tilde{p}|\tilde{x})} (\tilde{y} - \mu(\tilde{x}, \tilde{p})) + \int_{\tilde{p}}^\infty \mu(\tilde{x}, z) dz - h_D(\tilde{x})
\end{align*}

For the specific functional provided, the EIF is:
\begin{align*}
\frac{F^{\pi_D}(P|X)}{\pi_D(P|X)} (Y - \mu(X, P)) + \int_{P}^{\infty} \mu(X, z) dz - \Psi(\mathcal{P})
\end{align*} 
\end{proof}

\section{Proof of Proposition \ref{prop: DR} : Double Robustness}
\label{sec:proof_DR_prop}

\begin{proof}
    We establish that $\left|\widetilde \calS_{ACPW}(\pi)-\calS(\pi)\right|=o_p(1)$ by considering the two conditions of model specification.

    \textit{Case 1:}
    If the demand model is correctly specified such that  $\sup_{x,p} |\widehat{\mu}(x,p) - {\mu}(x,p)| =o_p(1) $. We decompose the estimator as:
    \begin{align*}\widetilde \calS_{ACPW}(\pi) &= \frac{1}{n} \sum_{i=1}^n \int_0^\infty \pi(p | X_i) \int_{p}^\infty \widehat\mu(X_i, z) dz dp + \frac{1}{n} \sum_{i=1}^n \frac{F^\pi(P_i | X_i)}{\widehat \pi_D(P_i | X_i)} (Y_i - \widehat \mu(X_i,P_i))\\
    &= \frac{1}{n} \sum_{i=1}^n \int_0^\infty \pi(p | X_i) \int_{p}^\infty \mu(X_i, z) dz dp + \frac{1}{n} \sum_{i=1}^n \frac{F^\pi(P_i | X_i)}{\bar \pi_D(P_i | X_i)} (Y_i - \mu(X_i,P_i)) + o_p(1)\end{align*}
    Since $\E[Y_i | X_i, P_i] = \mu(X_i, P_i)$, the conditional expectation of the second term is: 
    $$\E\left[ \frac{F^\pi(P_i | X_i)}{\bar \pi_D(P_i | X_i)} (Y_i - \mu(X_i,P_i)) \bigg| X_i, P_i \right] = \frac{F^\pi(P_i | X_i)}{\bar \pi_D(P_i | X_i)} \big( \mu(X_i, P_i) - \mu(X_i, P_i) \big) = 0$$Thus, by the Law of Large Numbers, the second term converges to $0$. The first term converges to $\calS(\pi)$, ensuring consistency even if $\widehat \pi_D$ converges to an incorrect limit $\bar \pi_D \neq \pi_D$.
    
    \textit{Case 2:} 
    If the behavior policy is correctly specified in the sense that $\sup_{x,p} |\widehat{\pi}_D(p|x) - \pi_D(p|x)| =o_p(1)$. We rewrite the estimator to isolate the effect of the nuisance demand model $\bar \mu$:
    \begin{align*}\widetilde \calS_{ACPW}(\pi) &= \frac{1}{n} \sum_{i=1}^n \frac{F^\pi(P_i | X_i)}{\widehat \pi_D(P_i | X_i)} Y_i + \frac{1}{n} \sum_{i=1}^n \left[ \int_0^\infty \pi(p | X_i) \int_{p}^\infty \widehat\mu(X_i, z) dz dp - \frac{F^\pi(P_i | X_i)}{\widehat \pi_D(P_i | X_i)} \widehat \mu(X_i,P_i) \right] \\
    &= \frac{1}{n} \sum_{i=1}^n \frac{F^\pi(P_i | X_i)}{\pi_D(P_i | X_i)} Y_i + \underbrace{\mathbb{E} \left[ \int_0^\infty \pi(p | X) \int_{p}^\infty \bar\mu(X, z) dz dp - \frac{F^\pi(P | X)}{\pi_D(P | X)} \bar \mu(X,P) \right]}_{\text{0 by Lemma \ref{lemma: balancing}}} + o_p(1)\end{align*}The remaining term converges to $\calS(\pi)$ by Theorem \ref{thm: alternative identification result}.
\end{proof}

\section{Proof of Theorem \ref{thm:eif aware} : EIF Derivation for Inequality-Aware Surplus}
\label{sec:eif_proof_inequality}

\begin{proof}

Lemma \ref{lemma: tangent} establishes that the observed tangent space constitutes the entire Hilbert space. Since the tangent space is the entire Hilbert space. There exists a unique influence function. Moreover, it is the efficient influence function.

It remains to calculate the canonical gradient of the functional $\mathcal{S}^r(\pi)$. Following the procedure outlined in the proof of Theorem \ref{thm:eif}, the gradient can be derived in a similar fashion.

\textit{Case 1: Target policy evaluation.}

Let the inner functional be defined as:
$$ h(X) = \int_0^\infty \pi(p|X) \int_p^\infty \mu(X, z) dz dp = \int_0^\infty \mu(X, z) F^\pi(z|X) dz, $$
where $\mathcal{S}^r(\pi) = \Psi(\mathcal{P})$.
The full estimand of interest is $\Psi(\mathcal{P}) = \mathbb{E}_X [h(X)^r]$.  We define the parametric submodel $\mathcal{P}_t = (1-t)\mathcal{P} + t\delta_{\tilde{x}, \tilde{y}, \tilde{p}}$.
Then the pathwise derivative of the full functional is:
$$ \frac{d\Psi(\mathcal{P}_t)}{dt} \bigg|_{t=0} = \left( h(\tilde{x})^r - \Psi(\mathcal{P}) \right) + \mathbb{E}_X \left[ r h(X)^{r-1} \frac{d h_t(X)}{dt} \bigg|_{t=0} \right] $$

We focus on the conditional derivative $\frac{d h_t(X)}{dt}$. Since $F^\pi$ is fixed, only $\mu(X,z)$ is perturbed. Note that $\mu(X,z) = \frac{f_{Y,P}(1,z|X)}{\pi_D(z|X)}$. Applying the chain rule to the perturbation: 
\begin{align*}
\frac{d \mu_t(X,z)}{dt} \bigg|_{t=0} &= \frac{d}{dt} \left( \frac{f_{Y,P,t}(1,z|X)}{\pi_{D,t}(z|X)} \right) \bigg|_{t=0} \\
&= \frac{\pi_D(z|X) \frac{d}{dt}f_{Y,P,t}(1,z|X) - f_{Y,P}(1,z|X) \frac{d}{dt}\pi_{D,t}(z|X)}{\pi_D(z|X)^2} \\
&= \frac{1}{\pi_D(z|X)} \left( \delta_{\tilde{y}, \tilde{p}}(1, z|X) - f_{Y,P}(1, z|X) \right) \\
&\quad - \frac{\mu(X,z)}{\pi_D(z|X)} \left( \delta_{\tilde{p}}(z|X) - \pi_D(z|X) \right)
\end{align*}

Now, substitute this into the integral for $h(X)$:
\begin{align*}
\frac{d h_t(X)}{dt} \bigg|_{t=0} &= \int_0^\infty \frac{d \mu_t(X,z)}{dt} F^\pi(z|X) dz \\
&= \mathbb{I}(X=\tilde{x}) \left[ \frac{F^\pi(\tilde{p}|X)}{\pi_D(\tilde{p}|X)} \tilde{y} - \int \mu(X,z)F^\pi(z|X) dz \right] \\
&\quad - \mathbb{I}(X=\tilde{x}) \left[ \frac{\mu(X,\tilde{p})F^\pi(\tilde{p}|X)}{\pi_D(\tilde{p}|X)} - \int \mu(X,z)F^\pi(z|X) dz \right]
\end{align*}

Notice that the integral terms $\int \mu(X,z)F^\pi(z|X)dz$ (which equal $h(X)$) cancel out:
\begin{align*}
\frac{d h_t(X)}{dt} \bigg|_{t=0} &= \mathbb{I}(X=\tilde{x}) \left[ \frac{F^\pi(\tilde{p}|X)}{\pi_D(\tilde{p}|X)} \tilde{y} - \frac{\mu(X,\tilde{p})F^\pi(\tilde{p}|X)}{\pi_D(\tilde{p}|X)} \right] \\
&= \mathbb{I}(X=\tilde{x}) \left[ \frac{F^\pi(\tilde{p}|X)}{\pi_D(\tilde{p}|X)} (\tilde{y} - \mu(X, \tilde{p})) \right]
\end{align*}

Finally, substitute this back into the derivative of $\Psi(\mathcal{P})$ and combine with the marginal $X$ variation, the resulting EIF is:
\begin{align*}
h(X)^r - \mathcal{S}^r(\pi) + r h(X)^{r-1} \left[ \frac{F^\pi(P|X)}{\pi_D(P|X)} (Y - \mu(X, P)) \right]
\end{align*}

\textit{Case 2: Behavior policy evaluation.}

The estimand is $\Psi(\mathcal{P}) = \mathbb{E}_X [h(X)^r]$, where $h(X) = \int_0^\infty \mu(X, z) F_P(z|X) dz$.
The pathwise derivative is:
\begin{align*}
\frac{d\Psi(\mathcal{P}_t)}{dt} \bigg|_{t=0} &= (h(\tilde{x})^r - \Psi(\mathcal{P})) + \mathbb{E}_X \left[ r h(X)^{r-1} \frac{d h_t(X)}{dt} \bigg|_{t=0} \right]
\end{align*}

We now derive the conditional derivative $\frac{d h}{dt}$ using the product rule on the two unknown components, $\mu$ and $F_P$:
\begin{align*}
\frac{d h_t(X)}{dt} \bigg|_{t=0} &= \underbrace{\int_0^\infty \frac{d \mu_t(X,z)}{dt} F_P(z|X) dz}_{\text{Part A: Outcome variation}} + \underbrace{\int_0^\infty \mu(X,z) \frac{d F_{P,t}(z|X)}{dt} dz}_{\text{Part B: CDF variation}}
\end{align*}

Evaluating Part A (as derived previously):
$$ \text{Part A} = \frac{\delta_{\tilde{x}}(X)}{f_X(X)} \left[ \frac{F_P(\tilde{p}|X)}{\pi_D(\tilde{p}|X)}(\tilde{y} - \mu(X, \tilde{p})) \right] $$

Evaluating Part B (using the derivative of the CDF $\frac{d F_{P,t}(z|X)}{dt} = \frac{\delta_{\tilde{x}}(X)}{f_X(X)} (\mathbb{I}(\tilde{p} \leq z) - F_P(z|X))$):
\begin{align*}
\text{Part B} &= \frac{\delta_{\tilde{x}}(X)}{f_X(X)} \int_0^\infty \mu(X,z) (\mathbb{I}(\tilde{p} \leq z) - F_P(z|X)) dz \\
&= \frac{\delta_{\tilde{x}}(X)}{f_X(X)} \left[ \int_{\tilde{p}}^\infty \mu(X,z) dz - h(X) \right]
\end{align*}

Substituting these back into the expectation over $X$, the $\frac{\delta_{\tilde{x}}(X)}{f_X(X)}$ terms sift the expression to the observed point $\tilde{x}$:
\begin{align*}
\text{EIF}(\tilde{x}, \tilde{y}, \tilde{p}) &= h(\tilde{x})^r - \Psi(\mathcal{P}) + r h(\tilde{x})^{r-1} \left[ \frac{F_P(\tilde{p}|\tilde{x})}{\pi_D(\tilde{p}|\tilde{x})}(\tilde{y} - \mu(\tilde{x}, \tilde{p})) + \int_{\tilde{p}}^\infty \mu(\tilde{x}, z) dz - h(\tilde{x}) \right]
\end{align*}

Dropping the tildes for the general form $(X, Y, P)$:
$$ \text{EIF} = h(X)^r + r h(X)^{r-1} \left[ \frac{(Y - \mu(X,P))F_P(P|X)}{\pi_D(P|X)} + \int_P^\infty \mu(X,z) dz - h(X) \right] - \mathcal{S}^r(\pi_D) $$
\end{proof}

\section{Proof of Theorem \ref{thm:dr rate} : Asymptotic Normality for Target Policy} 
\label{sec:normality_target}

We prove results for CPW, ACPW, and DM in the following sections. First we state the necessary assumptions to prove this result for the DM,

\subsection{Assumptions for the Direct Method}
\label{sec:assumptions_DM}
    \begin{assumption}[Assumptions required for the DM] \label{asmp: dm}
    (i) $\sqrt{\E\left[\widehat \mu(X,P)-\mu(X,P)\right]^2}=o_p(1).$\\
(ii)  The demand function is estimated using a function class $\mathcal{R}$ that satisfies the Donsker property.\\
(iii) The estimated demand function satisfies \begin{align*}
        \frac{1}{n} \sum_{i=1}^n \omega(X_i,P_i) \left(\widehat \mu(X_i,P_i)- Y_i \right)=o_p(n^{-1/2}), \mbox{ where } \omega(x,p)=\frac{F^\pi(p | x)}{\pi_D(p | x)}.
    \end{align*}
    (iv) Assume $\pi_D(p \given x)>c$, for all $p \in \calP$, and every $x$, for some constant $c$.
\end{assumption}

Assumption \ref{asmp: dm} (i) is relatively mild, as it merely requires $\widehat \mu$ to be consistent, with no rate specified. Assumption \ref{asmp: dm} (ii)
imposes a complexity (size) constraint on the function class used for estimating the demand function. Intuitively speaking, a Donsker class is a collection of functions that is not too large or too complex. This helps ensure that the average behavior of these functions becomes stable as we collect more data. 
Assumption \ref{asmp: dm} (iii) is mild and holds when the ratio $\omega(x,p)$  belongs to the function class $\mathcal{R}$. For instance, when $\mathcal{R}$ contains Hölder smooth functions, a series logistic regression estimator based on the sieve approach \citep{geman1982nonparametric} satisfies Assumption \ref{asmp: dm} (iii). 

\subsection{Proof for Part (iii): CPW for Target Policy Evaluation}

\begin{proof}

Denote $\epsilon_i \equiv Y_i-\mu(X_i,P_i)$. Recall that the EIF is \begin{align*}
    h(X_i) + \omega(X_i,P_i)(Y_i-\mu(X_i,P_i))-\mathcal S(\pi)=h(X_i) + \omega(X_i,P_i)\epsilon_i-\mathcal S(\pi),
\end{align*} where $h(X)\equiv \int_0^\infty\pi(p | X) \int_{p}^\infty \mu(X, z)dzdp .$

Then the CPW estimator $\widehat S_{CPW}(\pi)$ is 
\begin{align*}
    & \widehat S_{CPW}(\pi)= \frac{1}{n} \sum_{i=1}^n \widehat \omega(X_i,P_i) Y_i=  \frac{1}{n} \sum_{i=1}^n \widehat \omega(X_i,P_i) \mu(X_i,P_i)+\frac{1}{n} \sum_{i=1}^n \widehat \omega(X_i,P_i)\epsilon_i\\
    = &\; \frac{1}{n} \sum_{i=1}^n \widehat \omega(X_i,P_i) \phi(X_i,P_i)^\top \beta +\frac{1}{n} \sum_{i=1}^n \widehat \omega(X_i,P_i)\epsilon_i + \underbrace{O(L^{-s/d}) }_{o_p(n^{-1/2}) \mbox{ by Assumption \ref{asmp: CPW}}} \\
    = &\;   \frac{1}{n} \sum_{i=1}^n  \phi^\pi(X_i)^\top \beta +\frac{1}{n} \sum_{i=1}^n \widehat \omega(X_i,P_i)\epsilon_i +o_p(n^{-1/2}) \\
    = &\;   \frac{1}{n} \sum_{i=1}^n  \int_0^\infty\pi(p | X_i) \int_{p}^\infty \mu(X_i, z) dzdp +\frac{1}{n} \sum_{i=1}^n \widehat \omega(X_i,P_i)\epsilon_i +o_p(n^{-1/2})\\
    = &\;  \underbrace{ \frac{1}{n} \sum_{i=1}^n  h(X_i) +\frac{1}{n} \sum_{i=1}^n  \omega(X_i,P_i)\epsilon_i}_{\frac{1}{n} \sum_{i=1}^n \psi^\pi({\cal D}_i)+S(\pi)}+\frac{1}{n} \sum_{i=1}^n \widehat \omega(X_i,P_i)\epsilon_i -\frac{1}{n} \sum_{i=1}^n  \omega(X_i,P_i)\epsilon_i +o_p(n^{-1/2}).
\end{align*}

Thus, it remains to show that \begin{align*}
    \frac{1}{n} \sum_{i=1}^n \widehat \omega(X_i,P_i)\epsilon_i -\frac{1}{n} \sum_{i=1}^n  \omega(X_i,P_i)\epsilon_i =o_p(n^{-1/2}).
\end{align*} According to Lemma 19.24 in \citet{van2000asymptotic},
the above holds by noticing that $\widehat w$ belongs to the  Donsker class and satisfies $$\sqrt{\E\left[\widehat \omega(X,P)-\omega(X,P)\right]^2}=o_p(1),$$ and $\E\left[(\widehat \omega(X,P)-\omega(X,P))\epsilon \right]=0$.

Rearranging the terms establishes that $\sqrt{n}(\widehat S_{CPW}(\pi) - S(\pi)) = \frac{1}{\sqrt{n}} \sum_{i=1}^n \psi^\pi(\mathcal{D}_i) + o_p(1)$. Since the Efficient Influence Functions $\psi^\pi(\mathcal{D}_i)$ are i.i.d. with mean zero and finite variance, the Central Limit Theorem implies that this leading term converges in distribution to a normal random variable with variance $\text{Var}(\psi^{\pi}(\mathcal{D}))$. This completes the proof.

\end{proof}

\subsection{Proof for Part (ii): ACPW for Target Policy}

The proof is based on the cross-fitting technique \citep{chernozhukov2018double} , which we demonstrated below.
We divide the data into \( K \) approximately equal-sized folds. For each observation \( i \), we train the nuisance models, i.e., the behavior policy \( \widehat{\pi}_D(P \mid X) \) and the reward model \( \widehat \mu(X, P) \), using only the data that does not include observation \( i \)'s fold, denoted by \( -k(i) \). This yields estimators \( \widehat{\pi}_D^{-k(i)}(P_i \mid X_i) \) and \( \widehat \mu^{-k(i)}(X_i, P_i) \) evaluated on the held-out observation \( i \). This procedure ensures that the nuisance estimates are independent of the data used for evaluation, mitigating overfitting and allowing for valid statistical inference.

Given the propensity score model $\widehat{\pi}_D^{-k(i)}(P_i|X_i)$ and reward model $\widehat \mu^{-k(i)}(X_i,P_i)$ fitted by cross fitting, the corresponding empirical estimator is 
\begin{align}
    \widehat{\cal S}_\text{ACPW}(\pi) =  \frac{1}{n} \sum_{i=1}^n  \bigg[\int_0^\infty\pi(p | X_i) \int_{p}^\infty \widehat\mu^{-k(i)}(X_i, z)dzdp  + \frac{F^\pi(P_i | X_i)}{\widehat \pi_D^{-k(i)}(P_i | X_i)}(Y_i - \widehat \mu^{-k(i)}(X_i,P_i))\bigg]
\end{align}

Let the oracle estimator be
\begin{align}
\bar{\cal S}_\text{ACPW}(\pi) = \frac{1}{n} \sum_{i=1}^n  \bigg[\int_0^\infty\pi(p | X_i) \int_{p}^\infty \mu(X_i, z)dzdp  + \frac{F^\pi(P_i | X_i)}{\pi_D(P_i | X_i)}(Y_i -  \mu(X_i,P_i))\bigg]
\end{align}
We have the following lemma:

\begin{lemma}
Under assumptions in Theorem \ref{thm:dr rate} (ii), we have
\[
\bigl|\widehat{\mathcal S}_{\mathrm{ACPW}}(\pi)
-
\bar{\mathcal S}_{\mathrm{ACPW}}(\pi)\bigr|
=o_p(n^{-1/2}),
\] where $\bar{\mathcal S}_{\mathrm{ACPW}}(\pi)$ denotes the oracle ACPW estimator.
\end{lemma}

\begin{proof}
     Let $\{I_k\}_{k=1}^K$ be a partition of the indices $\{1, \dots, n\}$ such that $|I_k| = n/K$. We denote $n_k = |I_k|$. We decompose the difference as:$$ \widehat{\mathcal S}_{\mathrm{ACPW}}(\pi) - \bar{\mathcal S}_{\mathrm{ACPW}}(\pi) = D_1(\pi) + D_2(\pi) + D_3(\pi) $$where each $D_j(\pi) = \frac{1}{n} \sum_{k=1}^K \sum_{i \in I_k} \psi_{j,i}$ for appropriate summands $\psi_{j,i}$.

$$ D_1(\pi) = \frac{1}{n} \sum_{k=1}^K \sum_{i \in I_k} \underbrace{ \left( \frac{F^\pi(P_i \mid X_i)}{\widehat{\pi}_D^{-k(i)}(P_i \mid X_i)} - \frac{F^\pi(P_i \mid X_i)}{\pi_D(P_i \mid X_i)} \right) (Y_i - \mu(X_i, P_i)) }_{\psi_{1,i}}. $$

$$ D_2(\pi) = \frac{1}{n} \sum_{k=1}^K \sum_{i \in I_k} \underbrace{ \left[ \frac{F^\pi(P_i \mid X_i)}{\pi_D(P_i \mid X_i)} (\mu(X_i, P_i) - \widehat{\mu}^{-k(i)}(X_i, P_i)) - \int F^\pi(z \mid X_i) (\mu(X_i, z) - \widehat{\mu}^{-k(i)}(X_i, z)) dz \right] }_{\psi_{2,i}} $$

$$ D_3(\pi) = \frac{1}{n} \sum_{k=1}^K \sum_{i \in I_k} \underbrace{ \left( \frac{F^\pi(P_i \mid X_i)}{\widehat{\pi}_D^{-k(i)}(P_i \mid X_i)} - \frac{F^\pi(P_i \mid X_i)}{\pi_D(P_i \mid X_i)} \right) (\mu(X_i, P_i) - \widehat{\mu}^{-k(i)}(X_i, P_i)) }_{\psi_{3,i}}. $$

Term $D_1(\pi)$. For a fixed fold $k$, let $\mathcal{D}_k^c$ denote the data used to estimate $\widehat{\pi}_D^{-k(i)}$. For $i \in I_k$, the summands are:$$ \psi_{1,i} = \left( \frac{F^\pi(P_i \mid X_i)}{\widehat{\pi}_D^{-k(i)}(P_i \mid X_i)} - \frac{F^\pi(P_i \mid X_i)}{\pi_D(P_i \mid X_i)} \right) (Y_i - \mu(X_i, P_i)) $$Conditioning on $\mathcal{D}_k^c$, the terms $\{\psi_{1,i}\}_{i \in I_k}$ are i.i.d. and:$$ \mathbb{E}[\psi_{1,i} \mid \mathcal{D}_k^c] = \mathbb{E}\left[ \mathbb{E}[\psi_{1,i} \mid X_i, P_i, \mathcal{D}_k^c] \mid \mathcal{D}_k^c \right] = 0 $$since $\mathbb{E}[Y_i - \mu(X_i, P_i) \mid X_i, P_i] = 0$. The variance of $D_1(\pi)$ satisfies:$$ \text{Var}(D_1(\pi)) = \frac{1}{n^2} \sum_{k=1}^K \mathbb{E}\left[ \left( \sum_{i \in I_k} \psi_{1,i} \right)^2 \right] = \frac{1}{n^2} \sum_{k=1}^K n_k \mathbb{E}[\psi_{1,i}^2]. $$Under the $L_2$-consistency of $\widehat{\pi}_D^{-k(i)}$ and bounded overlap, $\mathbb{E}[\psi_{1,i}^2] = o(1)$. Thus, $\text{Var}(D_1(\pi)) = o(n^{-1})$, which implies $D_1(\pi) = o_p(n^{-1/2})$ by Chebyshev's inequality.

Term $D_2(\pi)$. For $i \in I_k$, define:$$ \psi_{2,i} = \frac{F^\pi(P_i \mid X_i)}{\pi_D(P_i \mid X_i)} \Delta\mu_i - \int F^\pi(z \mid X_i) \Delta\mu(X_i, z) dz $$where $\Delta\mu(X_i, \cdot) = \mu(X_i, \cdot) - \widehat\mu^{-k(i)}(X_i, \cdot)$. By the balancing property (Lemma \ref{lemma: balancing}):$$ \mathbb{E}\left[ \frac{F^\pi(P_i \mid X_i)}{\pi_D(P_i \mid X_i)} \Delta\mu_i \;\middle|\; X_i, \mathcal{D}_k^c \right] = \int F^\pi(z \mid X_i) \Delta\mu(X_i, z) dz $$Thus $\mathbb{E}[\psi_{2,i} \mid \mathcal{D}_k^c] = 0$. Similar to Step 1, the cross-fitting ensures these are uncorrelated across $i \in I_k$. Given the $L_2$-consistency of $\widehat{\mu}$, we also have $L_2$-consistency of $\int F^\pi(z \mid X_i) \widehat \mu(X_i, z) dz$ (Lemma \ref{lem:error transform}). Similarly by bounding the first term, we have $\mathbb{E}[\psi_{2,i}^2] = o(1)$, leading to $D_2(\pi) = o_p(n^{-1/2})$.

Term $D_3(\pi)$. This is the ``product'' error term. For each fold $k$:$$ |D_{3,k}(\pi)| \leq \left\| \frac{F^\pi}{\widehat{\pi}_D^{-k(i)}} - \frac{F^\pi}{\pi_D} \right\|_{L_{2,P_k}} \left\| \mu - \widehat\mu^{-k(i)} \right\|_{L_{2,P_k}} $$where $L_{2,P_k}$ denotes the empirical $L_2$ norm over fold $k$. Under the assumption that $\|\widehat{\mu}-\mu\|_{L_2} = O_p(n^{-\alpha_1})$ and $\|\widehat{\omega}-\omega\|_{L_2} = O_p(n^{-\alpha_2})$, we have:$$ |D_3(\pi)| \leq \sum_{k=1}^K \frac{n_k}{n} O_p(n^{-(\alpha_1+\alpha_2)}) = O_p(n^{-(\alpha_1+\alpha_2)}) $$Since $\alpha_1 + \alpha_2 > 1/2$, it follows that $D_3(\pi) = o_p(n^{-1/2})$.

The result follows from $D_1(\pi) + D_2(\pi) + D_3(\pi) = o_p(n^{-1/2})$.

Observing the definition of the Oracle estimator $\bar{\mathcal S}_{\mathrm{ACPW}}(\pi)$, we see that it is exactly the sample average of the Efficient Influence Functions plus the true parameter:
$$ \bar{\mathcal S}_{\mathrm{ACPW}}(\pi) = \mathcal{S}(\pi) + \frac{1}{n} \sum_{i=1}^n \psi^\pi(\mathcal{D}_i) $$
where $\psi^\pi(\mathcal{D}_i)$ are i.i.d. with mean zero and variance $\Sigma(\pi) = \text{Var}(\psi^\pi(\mathcal{D}))$. By the Central Limit Theorem, $\sqrt{n}(\bar{\mathcal S}_{\mathrm{ACPW}}(\pi) - \mathcal{S}(\pi)) \xrightarrow{d} \mathcal{N}(0, \Sigma(\pi))$. 
By Slutsky's theorem, since the difference between the empirical and oracle estimator is $o_p(n^{-1/2})$, the empirical estimator $\widehat{\mathcal S}_{\mathrm{ACPW}}(\pi)$ shares the same asymptotic distribution:
$$ \sqrt{n}(\widehat{\mathcal S}_{\mathrm{ACPW}}(\pi) - \mathcal{S}(\pi)) \xrightarrow{d} \mathcal{N}(0, \Sigma(\pi)). $$
\end{proof}

\subsection{Proof for Part (iii): DM for Target Policy}
\begin{proof}
    The direct method estimator $\widehat S_{DM}(\pi)$ is obtained by \begin{gather*}
        \widehat S_{DM}(\pi)=\frac{1}{n} \sum_{i=1}^n \widehat h(X_i),
    \end{gather*} where 
\begin{gather*}
        \widehat h(x)= \int_0^\infty\pi(p | x) \int_{p}^\infty \widehat \mu(x, a) dadp.
    \end{gather*}
 It follows that
\begin{align*}
       & \widehat S_{DM}(\pi)= \frac{1}{n} \sum_{i=1}^n \widehat h(X_i)-\frac{1}{n} \sum_{i=1}^n h(X_i)+\frac{1}{n} \sum_{i=1}^n  h(X_i)\\
     = & \, \underbrace{\frac{1}{n} \sum_{i=1}^n \widehat h(X_i)-\frac{1}{n} \sum_{i=1}^n h(X_i)-\E\left[ \widehat h(X)- h(X)\right]}_{o_p(n^{-1/2}) \mbox{ by Lemma 19.24 in \citet{van2000asymptotic}}}+\E\left[ \widehat h(X)- h(X)\right]\\
      & \,+\frac{1}{n} \sum_{i=1}^n  h(X_i)\\
     = & \, \E\left[ \widehat h(X)- h(X)\right]+\frac{1}{n} \sum_{i=1}^n  h(X_i)+o_p(n^{-1/2})\\
     = & \, \E\left[ \omega(X,P) \left(\widehat \mu(X,P)- \mu(X,P)\right)\right]+\frac{1}{n} \sum_{i=1}^n  h(X_i)+o_p(n^{-1/2}), 
    \end{align*} where the last equation simply follows  Equation \eqref{eq:balance} in Lemma \ref{lemma: balancing}. Denote $\epsilon_i \equiv Y_i-\mu(X_i,P_i)$. It then follows that 
    \begin{align*}
       & \E\left[ \omega(X,P) \left(\widehat \mu(X,P)- \mu(X,P)\right)\right]+\frac{1}{n} \sum_{i=1}^n  h(X_i)+o_p(n^{-1/2})\\
     = & \, \underbrace{\E\left[ \omega(X,P) \left(\widehat \mu(X,P)- \mu(X,P)\right)\right]-\frac{1}{n} \sum_{i=1}^n \omega(X_i,P_i) \left(\widehat \mu(X_i,P_i)- \mu(X_i,P_i)\right)}_{o_p(n^{-1/2}) \mbox{ by Lemma 19.24 in \citet{van2000asymptotic}}}\\
     & + \frac{1}{n} \sum_{i=1}^n \omega(X_i,P_i) \left(\widehat \mu(X_i,P_i)- \mu(X_i,P_i)\right) +\frac{1}{n} \sum_{i=1}^n  h(X_i)+o_p(n^{-1/2})\\
     = & \, \frac{1}{n} \sum_{i=1}^n \omega(X_i,P_i) \left(\widehat \mu(X_i,P_i)- \mu(X_i,P_i)\right) +\frac{1}{n} \sum_{i=1}^n  h(X_i)+o_p(n^{-1/2})\\
     = & \, \frac{1}{n} \sum_{i=1}^n \omega(X_i,P_i) \left(\widehat \mu(X_i,P_i)- Y_i +\epsilon_i \right) +\frac{1}{n} \sum_{i=1}^n  h(X_i)+o_p(n^{-1/2})\\
     = & \, \underbrace{\frac{1}{n} \sum_{i=1}^n \omega(X_i,P_i) \left(\widehat \mu(X_i,P_i)- Y_i \right)}_{o_p(n^{-1/2}) \mbox{ by Assumption \ref{asmp: dm}}} + \underbrace{\frac{1}{n} \sum_{i=1}^n  \omega(X_i,P_i) \epsilon_i + \frac{1}{n} \sum_{i=1}^n  h(X_i)}_{\frac{1}{n} \sum_{i=1}^n \psi^\pi({\cal D}_i)+S(\pi)}+o_p(n^{-1/2}).
    \end{align*}

Thus we have \begin{gather*}
    \widehat S_{DM}(\pi)= \frac{1}{n} \sum_{i=1}^n \psi^\pi({\cal D}_i)+o_p(n^{-1/2}).
\end{gather*}

Since the EIF $\psi^\pi(\mathcal{D}_i)$ are i.i.d. with mean zero and finite variance, the Central Limit Theorem implies that this leading term converges in distribution to a normal random variable with variance $\text{Var}(\psi^{\pi}(\mathcal{D}))$. This completes the proof. 

\end{proof}

\section{Asymptotic Normality of Behavioral Policy}
\label{sec:behavioral_theory}

This section presents the results for the behavioral surplus estimation. To maintain conciseness, proofs that follow standard procedures or duplicate previous logic have been omitted.

Since the behavior policy $\pi_D$ is unknown, the cumulative distribution function $F^{\pi_D}(p|x)$ must also be estimated from the data. Consequently, the estimated weight function is given by $\widehat{\omega}(p|x) = \frac{\widehat{F}^{\pi_D}(p|x)}{\widehat{\pi}_D(p|x)}$. With a slight abuse of notation, we continue to denote this ratio as $\widehat{\omega}$ to reflect its role as the plug-in estimator for the true ratio. We now introduce the following assumptions for the CPW estimator.

\begin{assumption} [Assumptions required for the CPW] \label{asmp: CPW behavior}
(i) $\sqrt{\E\left[\widehat \omega(X,P)-\omega(X,P)\right]^2}=o_p(1)$ , where $\omega(x,p)\equiv\frac{F^{\pi_D}(p | x)}{\pi_D(p | x)}$, and $\widehat \omega(x,p)\equiv\frac{\widehat F^{\pi_D}(p | x)}{\widehat\pi_D(p | x)} $ is the estimator of  $\omega(x,p)$. \\
(ii) The ratio $\widehat \omega$ is estimated using a function class that satisfies the Donsker property.\\
   (iii) There exist basis functions $\phi(x,p) \in \mathbb{R}^L$  
and a vector $\beta \in \mathbb{R}^L$ 
such that \begin{gather} \label{eq:approximation_behavior}
    \sup_{x,p} |\mu(x,p)-\phi(x,p)^\top \beta | =O(L^{-s/d}),
\end{gather} where $s$ is a fixed positive constant. \\
(iv) The estimated CPW weights satisfy \begin{gather*}
       \left \lVert\frac{1}{n} \sum_{i=1}^n \int_{P_i} \phi(X_i,z)dz- \frac{1}{n} \sum_{i=1}^n \widehat \omega(X_i,P_i)\phi(X_i,P_i) \right \rVert_2  =o_p(n^{-1/2}),
    \end{gather*} where $\phi(\cdot, \cdot)$ is the basis function that satisfy Equation \eqref{eq:approximation_behavior}.
\end{assumption}

Next, we state the asymptotic normality result under the behavioral policy.

\begin{theorem} \label{thm:dr rate behavior} Suppose that the assumptions in Theorem \ref{thm:eif aware} hold, we have the following results: \\
  (i)  Suppose Assumption \ref{asmp: CPW behavior} holds, and further assume that the number of basis functions $L$ satisfies $L \gg n^{d/2s}$, then the CPW estimator $\widehat \calS_{CPW}(\pi_D)$ attains the semiparametric efficiency bound, 
    \begin{gather*}
     \sqrt{n} \left( \widehat \calS_{CPW}(\pi_D)-\calS(\pi_D)\right)\rightarrow\mathcal{N}\left(0,\mbox{Var}[\psi^{\pi_D}(\cal {D})] \right).
    \end{gather*}

 (ii)   Under Assumption \ref{asmp: ACPW behavior}, the ACPW estimator $\widehat \calS_{ACPW}(\pi_D)$ attains the semiparametric efficiency bound: 
    \begin{gather*}
     \sqrt{n} \left( \widehat \calS_{ACPW}(\pi_D)-\calS(\pi_D)\right)\rightarrow \mathcal{N}\left(0,\mbox{Var}[\psi^{\pi_D}(\cal {D})] \right).
    \end{gather*}
   (iii) Under Assumption \ref{asmp: dm}, the DM estimator $\widehat \calS_{DM}(\pi_D)$ attains the semiparametric efficiency bound, 
    \begin{gather*}
     \sqrt{n} \left( \widehat \calS_{DM}(\pi_D)-\calS(\pi_D)\right)\rightarrow \mathcal{N}\left(0,\mbox{Var}[\psi^{\pi_D}(\cal {D})] \right),
    \end{gather*} where $\psi^{\pi_D}(\cal {D})$ is the EIF for $\calS(\pi_D)$.
\end{theorem}

\subsection{Proof for CPW for Behavioral Policy}
\begin{proof}

Denote $\epsilon_i \equiv Y_i-\mu(X_i,P_i)$. Recall that the EIF is \begin{align*}
    g(X_i,P_i) + \omega(X_i,P_i)(Y_i-\mu(X_i,P_i))-S(\pi_D)=g(X_i,P_i) + \omega(X_i,P_i)\epsilon_i-S(\pi_D),
\end{align*} where $g(X,P)\equiv \int_{P}^\infty \mu(X, z)dz.$

Then the CPW estimator $\widehat S_{CPW}(\pi_D)$ is 
\begin{align*}
    & \widehat S_{CPW}(\pi_D)= \frac{1}{n} \sum_{i=1}^n \widehat \omega(X_i,P_i) Y_i=  \frac{1}{n} \sum_{i=1}^n \widehat \omega(X_i,P_i) \mu(X_i,P_i)+\frac{1}{n} \sum_{i=1}^n \widehat \omega(X_i,P_i)\epsilon_i\\
    = &\; \frac{1}{n} \sum_{i=1}^n \widehat \omega(X_i,P_i) \phi(X_i,P_i)^\top \beta +\frac{1}{n} \sum_{i=1}^n \widehat \omega(X_i,P_i)\epsilon_i + \underbrace{O(L^{-s/d}) }_{o_p(n^{-1/2}) \mbox{ by Assumption \ref{asmp: CPW behavior}}} \\
    = &\;   \frac{1}{n} \sum_{i=1}^n  \int_{P_i} \phi(X_i,z)^{\top} \beta dz  +\frac{1}{n} \sum_{i=1}^n \widehat \omega(X_i,P_i)\epsilon_i +o_p(n^{-1/2}) \\
    = &\;   \frac{1}{n} \sum_{i=1}^n   \int_{P_i}^\infty \mu(X_i, z) dz +\frac{1}{n} \sum_{i=1}^n \widehat \omega(X_i,P_i)\epsilon_i +o_p(n^{-1/2})\\
    = &\;  \underbrace{ \frac{1}{n} \sum_{i=1}^n  g(X_i,P_i) +\frac{1}{n} \sum_{i=1}^n  \omega(X_i,P_i)\epsilon_i}_{\frac{1}{n} \sum_{i=1}^n \psi^{\pi_D}({\cal D}_i)+S(\pi)}+\frac{1}{n} \sum_{i=1}^n \widehat \omega(X_i,P_i)\epsilon_i -\frac{1}{n} \sum_{i=1}^n  \omega(X_i,P_i)\epsilon_i +o_p(n^{-1/2}).
\end{align*}

Thus, it remains to show that \begin{align*}
    \frac{1}{n} \sum_{i=1}^n \widehat \omega(X_i,P_i)\epsilon_i -\frac{1}{n} \sum_{i=1}^n  \omega(X_i,P_i)\epsilon_i =o_p(n^{-1/2}).
\end{align*} According to Lemma 19.24 in \citet{van2000asymptotic},
the above holds by noticing that $\widehat w$ belongs to the  Donsker class and satisfies $$\sqrt{\E\left[\widehat \omega(X,P)-\omega(X,P)\right]^2}=o_p(1),$$ and $\E\left[(\widehat \omega(X,P)-\omega(X,P))\epsilon \right]=0$.

Rearranging the terms establishes that $\sqrt{n}(\widehat S_{CPW}(\pi_D) - S(\pi_D)) = \frac{1}{\sqrt{n}} \sum_{i=1}^n \psi^\pi(\mathcal{D}_i) + o_p(1)$. Since the Efficient Influence Functions $\psi^{\pi_D}(\mathcal{D}_i)$ are i.i.d. with mean zero and finite variance, the Central Limit Theorem implies that this leading term converges in distribution to a normal random variable with variance $\text{Var}(\psi^{\pi_D}(\mathcal{D}))$. This completes the proof.
\end{proof}

\subsection{Proof for ACPW for Behavioral Policy}

\begin{assumption}[Assumptions required for the ACPW] \label{asmp: ACPW behavior}
Assume $\pi_D(p \given x)>c$, for all $p \in \calP$, and every $x$, for some constant $c$. Suppose that the estimators for the demand function and the behavior policy are constructed using the cross-fitting procedure, and that they achieve the following convergence rate: 
    \begin{align} \label{eq: product rate}
        \sqrt{\E[(\widehat \mu(X,P)-\mu(X,P))^2]} =O_p(n^{-\alpha_1}), \mbox{ and } \; \sqrt{\E[(\widehat \omega(X,P)-\omega(X,P))^2]}=O_p(n^{-\alpha_2}),
    \end{align} with $\alpha_1, \alpha_2>0$, and $\alpha_1+\alpha_2>1/2$. Note here that $\widehat \omega =\widehat F^{\pi_D}/\widehat \pi_D.$
\end{assumption}

\begin{proof}
We rely on the cross-fitting technique. Let $\{I_k\}_{k=1}^K$ be a partition of the indices $\{1, \dots, n\}$ such that $|I_k| = n/K$. We denote $n_k = |I_k|$.

The behavioral ACPW estimator is:
$$ \widehat{\mathcal S}_{\mathrm{ACPW}}(\pi_D) = \frac{1}{n} \sum_{k=1}^K \sum_{i \in I_k} \left[ \frac{\widehat{F}^{\pi_D,-k(i)}(P_i|X_i)}{\widehat{\pi}_D^{-k(i)}(P_i|X_i)}(Y_i - \widehat{\mu}^{-k(i)}(X_i, P_i)) + \int_{P_i}^\infty \widehat{\mu}^{-k(i)}(X_i, z) dz \right] $$

We compare this to the Oracle estimator (which uses true nuisance parameters but the same observed integral structure):
$$ \bar{\mathcal S}_{\mathrm{ACPW}}(\pi_D) = \frac{1}{n} \sum_{i=1}^n \left[ \frac{F^{\pi_D}(P_i|X_i)}{\pi_D(P_i|X_i)}(Y_i - \mu(X_i, P_i)) + \int_{P_i}^\infty \mu(X_i, z) dz \right] $$

We decompose the difference as:
$$ \widehat{\mathcal S}_{\mathrm{ACPW}}(\pi_D) - \bar{\mathcal S}_{\mathrm{ACPW}}(\pi_D) = D_1(\pi_D) + D_2(\pi_D) + D_3(\pi_D) $$

\textit{Term $D_1(\pi_D)$:}
$$ D_1(\pi_D) = \frac{1}{n} \sum_{k=1}^K \sum_{i \in I_k} \underbrace{ \left( \frac{\widehat{F}^{\pi_D,-k(i)}(P_i \mid X_i)}{\widehat{\pi}_D^{-k(i)}(P_i \mid X_i)} - \frac{F^{\pi_D}(P_i \mid X_i)}{\pi_D(P_i \mid X_i)} \right) (Y_i - \mu(X_i, P_i)) }_{\psi_{1,i}}. $$
Conditioning on the nuisance training data $\mathcal{D}_k^c$, the terms $\{\psi_{1,i}\}_{i \in I_k}$ are i.i.d. with mean zero, since $\mathbb{E}[Y_i - \mu(X_i, P_i) \mid X_i, P_i] = 0$.
The variance of $D_1$ is determined by the expected squared error of the weights: $\mathbb{E}\left[ \left( \widehat \omega-\omega\right)^2 \right]$.
Then, under the $L_2$-consistency of $\widehat \omega $ (Assumption \ref{asmp: ACPW behavior}), we have $\mathbb{E}[\psi_{1,i}^2] = o(1)$, and by Chebyshev's inequality, $D_1(\pi_D) = o_p(n^{-1/2})$.

\textit{Term $D_2(\pi_D)$:}
$$ D_2(\pi_D) = \frac{1}{n} \sum_{k=1}^K \sum_{i \in I_k} \underbrace{ \left[ \frac{F^{\pi_D}(P_i \mid X_i)}{\pi_D(P_i \mid X_i)} (\mu(X_i, P_i) - \widehat{\mu}^{-k(i)}(X_i, P_i)) - \int_{P_i}^\infty (\mu(X_i, z) - \widehat{\mu}^{-k(i)}(X_i, z)) dz \right] }_{\psi_{2,i}}. $$
Let $\Delta\mu(X, z) = \mu(X, z) - \widehat{\mu}^{-k(i)}(X, z)$. To show $\mathbb{E}[\psi_{2,i} \mid \mathcal{D}_k^c] = 0$, we check the expectations.
The expectation of the integral term (using Fubini's theorem to switch integration order) is:
$$ \mathbb{E}\left[ \int_{P_i}^\infty \Delta\mu(X_i, z) dz \;\middle|\; X_i \right] = \int_0^\infty \pi_D(p|X_i) \int_p^\infty \Delta\mu(X_i, z) dz dp = \int_0^\infty \Delta\mu(X_i, z) F^{\pi_D}(z|X_i) dz. $$
The expectation of the weighted term is:
$$ \mathbb{E}\left[ \frac{F^{\pi_D}(P_i|X_i)}{\pi_D(P_i|X_i)} \Delta\mu(X_i, P_i) \;\middle|\; X_i \right] = \int_0^\infty \frac{F^{\pi_D}(z|X_i)}{\pi_D(z|X_i)} \Delta\mu(X_i, z) \pi_D(z|X_i) dz = \int_0^\infty \Delta\mu(X_i, z) F^{\pi_D}(z|X_i) dz. $$
Since the expectations match, $\mathbb{E}[\psi_{2,i} \mid \mathcal{D}_k^c] = 0$. Given the $L_2$-consistency of $\widehat{\mu}$, $\mathbb{E}[\psi_{2,i}^2] = o(1)$, leading to $D_2(\pi_D) = o_p(n^{-1/2})$.

\textit{Term $D_3(\pi_D)$:}

\begin{align*}
    & D_3(\pi_D) =\frac{1}{n} \sum_{k=1}^K \sum_{i \in I_k} \left( \frac{\widehat{F}^{\pi_D,-k(i)}}{\widehat{\pi}_D^{-k(i)}} - \frac{F^{\pi_D}}{\pi_D} \right) (\mu - \widehat{\mu}^{-k(i)}) \\
    =& \frac{1}{n} \sum_{k=1}^K \sum_{i \in I_k} \underbrace{ \left( \widehat \omega^{-k(i)} - \omega_i \right) (\mu_i - \widehat{\mu}^{-k(i)}) }_{\psi_{3,i}}. 
\end{align*}

By the Cauchy-Schwarz inequality, $|D_3|$ is bounded by the product of the $L_2$ norm of this weight difference and the $L_2$ norm of the demand error $\|\mu - \widehat{\mu}\|_{L_2}$.
Substituting the rates from Assumption \ref{asmp: ACPW behavior}:
$$ |D_3(\pi_D)| \lesssim \|\widehat\omega - \omega\|_{L_2} \|\mu - \widehat{\mu}\|_{L_2} = O_p(n^{-\alpha_2}) \times O_p(n^{-\alpha_1}) = O_p(n^{-(\alpha_1+\alpha_2)}). $$
Since $\alpha_1 + \alpha_2 > 1/2$, it follows that $D_3(\pi_D) = o_p(n^{-1/2})$.

We have shown that $\widehat{\mathcal S}_{\mathrm{ACPW}}(\pi_D) = \bar{\mathcal S}_{\mathrm{ACPW}}(\pi_D) + o_p(n^{-1/2})$.
Observing the definition of the Oracle estimator $\bar{\mathcal S}_{\mathrm{ACPW}}(\pi_D)$, we see that it is exactly the sample average of the Efficient Influence Functions plus the true parameter:
$$ \bar{\mathcal S}_{\mathrm{ACPW}}(\pi_D) = \mathcal{S}(\pi_D) + \frac{1}{n} \sum_{i=1}^n \psi^{\pi_D}(\mathcal{D}_i) $$
where $\psi^{\pi_D}(\mathcal{D}_i)$ are i.i.d. with mean zero and variance $\Sigma(\pi_D) = \text{Var}(\psi^{\pi_D}(\mathcal{D}))$. By the Central Limit Theorem, $\sqrt{n}(\bar{\mathcal S}_{\mathrm{ACPW}}(\pi_D) - \mathcal{S}(\pi_D)) \xrightarrow{d} \mathcal{N}(0, \Sigma(\pi_D))$.
By Slutsky's theorem, the empirical estimator $\widehat{\mathcal S}_{\mathrm{ACPW}}(\pi_D)$ shares the same asymptotic distribution:
$$ \sqrt{n}(\widehat{\mathcal S}_{\mathrm{ACPW}}(\pi_D) - \mathcal{S}(\pi_D)) \xrightarrow{d} \mathcal{N}(0, \Sigma(\pi_D)). $$
\end{proof}

\subsection{Proof for DM for Behavioral Policy}

\begin{proof}
    The direct method estimator $\widehat S_{DM}(\pi_D)$ is obtained by \begin{gather*}
        \widehat S_{DM}(\pi_D)=\frac{1}{n} \sum_{i=1}^n \widehat g(X_i,P_i),
    \end{gather*} where 
\begin{gather*}
        \widehat g(x,p)=  \int_{p}^\infty \widehat \mu(x, z) dz.
    \end{gather*}

 It follows that
\begin{align*}
       & \widehat S_{DM}(\pi_D)= \frac{1}{n} \sum_{i=1}^n \widehat g(X_i,P_i)-\frac{1}{n} \sum_{i=1}^n g(X_i,P_i)+\frac{1}{n} \sum_{i=1}^n  g(X_i,P_i)\\
     = & \, \underbrace{\frac{1}{n} \sum_{i=1}^n \widehat g(X_i,P_i)-\frac{1}{n} \sum_{i=1}^n g(X_i,P_i)-\E\left[ \widehat g(X,P)- g(X,P)\right]}_{o_p(n^{-1/2}) \mbox{ by Lemma 19.24 in \citet{van2000asymptotic}}}+\E\left[ \widehat g(X,P)- g(X,P)\right]\\
      & \,+\frac{1}{n} \sum_{i=1}^n  g(X_i,P_i)\\
     = & \, \E\left[ \widehat g(X,P)- g(X,P)\right]+\frac{1}{n} \sum_{i=1}^n  g(X_i,P_i)+o_p(n^{-1/2})\\
     = & \, \E\left[ \omega(X,P) \left(\widehat \mu(X,P)- \mu(X,P)\right)\right]+\frac{1}{n} \sum_{i=1}^n  g(X_i,P_i)+o_p(n^{-1/2}), 
    \end{align*} where the last equation simply follows  Equation \eqref{eq:balance} in Lemma \ref{lemma: balancing}, by letting $\pi=\pi_D$. It then follows that 
    \begin{align*}
       & \E\left[ \omega(X,P) \left(\widehat \mu(X,P)- \mu(X,P)\right)\right]+\frac{1}{n} \sum_{i=1}^n  g(X_i,P_i)+o_p(n^{-1/2})\\
     = & \, \underbrace{\E\left[ \omega(X,P) \left(\widehat \mu(X,P)- \mu(X,P)\right)\right]-\frac{1}{n} \sum_{i=1}^n \omega(X_i,P_i) \left(\widehat \mu(X_i,P_i)- \mu(X_i,P_i)\right)}_{o_p(n^{-1/2}) \mbox{ by Lemma 19.24 in \citet{van2000asymptotic}}}\\
     & + \frac{1}{n} \sum_{i=1}^n \omega(X_i,P_i) \left(\widehat \mu(X_i,P_i)- \mu(X_i,P_i)\right) +\frac{1}{n} \sum_{i=1}^n  g(X_i,P_i)+o_p(n^{-1/2})\\
     = & \, \frac{1}{n} \sum_{i=1}^n \omega(X_i,P_i) \left(\widehat \mu(X_i,P_i)- \mu(X_i,P_i)\right) +\frac{1}{n} \sum_{i=1}^n  g(X_i,P_i)+o_p(n^{-1/2})\\
     = & \, \frac{1}{n} \sum_{i=1}^n \omega(X_i,P_i) \left(\widehat \mu(X_i,P_i)- Y_i +\epsilon_i \right) +\frac{1}{n} \sum_{i=1}^n  g(X_i,P_i)+o_p(n^{-1/2})\\
     = & \, \underbrace{\frac{1}{n} \sum_{i=1}^n \omega(X_i,P_i) \left(\widehat \mu(X_i,P_i)- Y_i \right)}_{o_p(n^{-1/2}) \mbox{ by Assumption \ref{asmp: dm}}} + \underbrace{\frac{1}{n} \sum_{i=1}^n  \omega(X_i,P_i) \epsilon_i + \frac{1}{n} \sum_{i=1}^n  g(X_i,P_i)}_{\frac{1}{n} \sum_{i=1}^n \psi^\pi({\cal D}_i)+S(\pi_D)}+o_p(n^{-1/2}).
    \end{align*}

Thus we have \begin{gather*}
    \widehat S_{DM}(\pi_D)= \frac{1}{n} \sum_{i=1}^n \psi^\pi({\cal D}_i)+o_p(n^{-1/2}).
\end{gather*}

Since the Efficient Influence Functions $\psi^\pi(\mathcal{D}_i)$ are i.i.d. with mean zero and finite variance, the Central Limit Theorem implies that this leading term converges in distribution to a normal random variable with variance $\text{Var}(\psi^{\pi}(\mathcal{D}))$. This completes the proof. 

\end{proof}

\begin{corollary} \label{thm:dr rate behavior difference} Suppose that the assumptions in Theorem \ref{thm:eif aware} hold, we have the following results: \\
(i) Under Assumption \ref{asmp: dm}, the DM estimator $\widehat \Delta_{DM}(\pi)$ attains the semiparametric efficiency bound, 
    \begin{gather*}
     \sqrt{n} \left( \widehat \Delta_{DM}(\pi)-\Delta(\pi)\right)\rightarrow \mathcal{N}\left(0,\mbox{Var}[\psi^{\Delta}(\cal {D})] \right),
    \end{gather*} where $\psi^{\Delta}(\cal {D})=\psi^{\pi}(\cal {D}) - \psi^{\pi_D}(\cal {D})$.

  (ii)  Suppose Assumption \ref{asmp: CPW} holds, and further assume that the number of basis functions $L$ satisfies $L \gg n^{d/2s}$, then the CPW estimator $\widehat \Delta_{CPW}(\pi)$ attains the semiparametric efficiency bound, 
    \begin{gather*}
     \sqrt{n} \left( \widehat \Delta_{CPW}(\pi)-\Delta(\pi)\right)\rightarrow\mathcal{N}\left(0,\mbox{Var}[\psi^{\Delta}(\cal {D})] \right).
    \end{gather*}

 (iii)   Under Assumption \ref{asmp: ACPW}, the ACPW estimator $\widehat \Delta_{ACPW}(\pi)$ attains the semiparametric efficiency bound: 
    \begin{gather*}
     \sqrt{n} \left( \widehat \Delta_{ACPW}(\pi)-\Delta(\pi)\right)\rightarrow \mathcal{N}\left(0,\mbox{Var}[\psi^{\Delta}(\cal {D})] \right).
    \end{gather*}
\end{corollary}

This follows directly from the linearity of the efficient influence function and the asymptotic normality established in Theorems \ref{thm:dr rate} and \ref{thm:dr rate behavior}. 

\section{Proof of Proposition \ref{prop:variance} : Consistency of Variance Estimators}
\label{sec:variance_consistency}

\begin{proof}
    Recall that the three variance estimators share a similar structure, \begin{align*}
   &\widehat \Sigma_{CPW}(\pi)=   \frac{1}{n} \sum_{i=1}^n  \left[\int_0^\infty\pi(p | X_i) \int_{p}^\infty \widehat \mu(X_i,z) dzdp  + \frac{F^\pi(P_i | X_i)}{\widehat\pi_D (P_i | X_i)}(Y_i - \widehat \mu(X_i,P_i))-\widehat \calS_{CPW}(\pi)\right]^2,\\
   &\widehat \Sigma_{ACPW}(\pi)=   \frac{1}{n} \sum_{i=1}^n  \left[\int_0^\infty\pi(p | X_i) \int_{p}^\infty \widehat \mu^{-k(i)}(X_i,z) dzdp  + \frac{F^\pi(P_i | X_i)}{\widehat\pi_D^{-k(i)} (P_i | X_i)}(Y_i - \widehat \mu^{-k(i)}(X_i,P_i))-\widehat \calS_{ACPW}(\pi)\right]^2,\\
   & \widehat \Sigma_{DM}(\pi)=   \frac{1}{n} \sum_{i=1}^n  \left[\int_0^\infty\pi(p | X_i) \int_{p}^\infty \widehat \mu(X_i,z) dzdp  + \frac{F^\pi(P_i | X_i)}{\widehat\pi_D (P_i | X_i)}(Y_i - \widehat \mu(X_i,P_i))-\widehat \calS_{DM}(\pi)\right]^2.
\end{align*}

Thus, all three estimators can be represented as \begin{gather*}
    \widehat{\Sigma}_j(\pi) = \frac{1}{n} \sum_{i=1}^n \left[\widehat h_j(X_i)+\widehat \omega_j(X_i,P_i)(Y_i-\widehat \mu_j(X_i,P_i)) - \widehat \calS_j(\pi) \right]^2,
\end{gather*} for $j \in \{\text{CPW, ACPW, DM}\}$.

To show $\widehat \Sigma_j(\pi)$ is a consistent estimator, it suffices to show that $\widehat \Sigma_j(\pi)-\bar\Sigma(\pi)=o_p(1)$, where $\bar\Sigma(\pi)$ is the oracle estimator:
\begin{gather*}
    \bar\Sigma(\pi) = \frac{1}{n} \sum_{i=1}^n \left[ h(X_i)+ \omega(X_i,P_i)(Y_i- \mu(X_i,P_i)) - \calS(\pi) \right]^2.
\end{gather*}

To establish the consistency of $\widehat \Sigma_j(\pi)$, we analyze the difference $\widehat \Sigma_j(\pi) - \bar \Sigma(\pi)$:
\begin{align*}
    &\widehat \Sigma_j(\pi) - \bar \Sigma(\pi) = \frac{1}{n} \sum_{i=1}^n \Bigg( \left[\widehat h_j(X_i)+\widehat \omega_j(X_i,P_i)(Y_i-\widehat \mu_j(X_i,P_i)) - \widehat \calS_j(\pi) \right]^2 \\
    &\qquad - \left[ h(X_i)+ \omega(X_i,P_i)(Y_i- \mu(X_i,P_i)) - \calS(\pi) \right]^2 \Bigg).
\end{align*}
Using the identity $a^2 - b^2 = (a-b)^2 + 2b(a-b)$, we let the difference between the estimated and oracle terms for observation $i$ be:
\begin{align*}
    D_{j,i} &= \left[\widehat h_j(X_i) - h(X_i)\right] + \left[\widehat \omega_j(X_i,P_i) - \omega(X_i,P_i)\right](Y_i - \mu(X_i,P_i)) \\
    &\quad - \widehat \omega_j(X_i,P_i)(\widehat \mu_j(X_i,P_i) - \mu(X_i,P_i)) - (\widehat \calS_j(\pi) - \calS(\pi)).
\end{align*}
Then the difference in variance estimators becomes:
\begin{align*}
    \widehat \Sigma_j(\pi) - \bar \Sigma(\pi) &= \frac{1}{n} \sum_{i=1}^n D_{j,i}^2 + \frac{2}{n} \sum_{i=1}^n \left[ h(X_i)+ \omega(X_i,P_i)(Y_i- \mu(X_i,P_i)) - \calS(\pi) \right] D_{j,i}.
\end{align*}
By the Cauchy-Schwarz inequality, the second term is bounded by:
\begin{equation*}
    2 \sqrt{\bar \Sigma(\pi)} \sqrt{\frac{1}{n} \sum_{i=1}^n D_{j,i}^2}.
\end{equation*}

It thus suffices to show that $\frac{1}{n} \sum_{i=1}^n D_{j,i}^2=o_p(1).$

Under the assumptions in Theorem 
\ref{thm:dr rate}, all
the nuisance estimators are $L_2$-consistent, and the resulting estimator $\widehat \calS_j(\pi)$ is consistent for $\calS(\pi)$. Now we apply the Cauchy-Schwarz inequality $(a+b+c+d)^2 \leq 4(a^2+b^2+c^2+d^2)$ to obtain:
\begin{align*}
    \frac{1}{n} \sum_{i=1}^n D_{j,i}^2 &\leq \frac{4}{n} \sum_{i=1}^n (\widehat h_j(X_i) - h(X_i))^2 \\
    &\quad + \frac{4}{n} \sum_{i=1}^n (\widehat \omega_j(X_i,P_i) - \omega(X_i,P_i))^2(Y_i - \mu(X_i,P_i))^2 \\
    &\quad + \frac{4}{n} \sum_{i=1}^n \widehat \omega_j^2(X_i,P_i)(\widehat \mu_j(X_i,P_i) - \mu(X_i,P_i))^2 \\
    &\quad + 4(\widehat \calS_j(\pi) - \calS(\pi))^2\\
    & \lesssim \frac{1}{n} \sum_{i=1}^n (\widehat h_j(X_i) - h(X_i))^2 + \frac{1}{n} \sum_{i=1}^n (\widehat \omega_j(X_i,P_i) - \omega(X_i,P_i))^2\\
    &\quad + \frac{1}{n} \sum_{i=1}^n (\widehat \mu_j(X_i,P_i) - \mu(X_i,P_i))^2 +\underbrace{(\widehat \calS_j(\pi) - \calS(\pi))^2}_{o_p(1) \text{ by Theorem \ref{thm:dr rate}}}.
\end{align*}

For $j \in \{\text{CPW, DM}\}$, because $ \widehat{\omega}_j,$ and $\widehat{\mu}_j$ belong to a Donsker class, by Lemma 19.24 in \citet{van2000asymptotic} and the Slutsky's theorem, we have:
\begin{align*}
    \frac{1}{n} \sum_{i=1}^n (\widehat \mu_j(X_i,P_i) - \mu(X_i,P_i))^2 = \underbrace{\E[(\widehat \mu_j(X,P) - \mu(X,P))^2 ]}_{o_p(1)} + o_p(n^{-1/2}) = o_p(1), \\
    \frac{1}{n} \sum_{i=1}^n (\widehat \omega_j(X_i,P_i) - \omega(X_i,P_i))^2 = \underbrace{\E[(\widehat \omega_j(X,P) - \omega(X,P))^2 ]}_{o_p(1)} + o_p(n^{-1/2}) = o_p(1).
\end{align*}
By Lemma \ref{lem:error transform},  we have \begin{align*}
    \frac{1}{n} \sum_{i=1}^n (\widehat h_j(X_i) - h(X_i))^2 = \underbrace{\E[(\widehat h_j(X) - g(X,P))^2 ]}_{o_p(1)} + o_p(n^{-1/2}) = o_p(1).
\end{align*} Thus, we have $$\frac{1}{n} \sum_{i=1}^n D_{j,i}^2=o_p(1).$$

For $j = \text{ACPW}$, the proof follows a procedure similar to that of Theorem \ref{thm:dr rate}; we omit the details here for brevity. This thus completes the proof.

\end{proof}

\section{Proof of Theorem \ref{thm: aware rate}: Asymptotic Normality for Inequality-Aware Surplus}
\label{sec:normality_unequal}

\begin{proof}
It suffices to show that \begin{align*}
 \frac{1}{n}\sum_{i=1}^n \Bigg[ r \frac{(Y_i-\widehat \mu^{-k(i)}(X_i,P_i))F^\pi(P_i|X_i)}{\widehat{\pi}^{-k(i)}_D(P_i|X_i)} \left(\int_0^\infty \pi(p | X_i) \int_{p}^\infty \widehat \mu^{-k(i)}(X_i, z)dzdp\right)^{r-1} \\
 + \left(\int_0^\infty\pi(p | X_i) \int_{p}^\infty \widehat \mu^{-k(i)}(X_i, z)dzdp \right)^r \Bigg] \\
 - \frac{1}{n}\sum_{i=1}^n \Bigg[ r \frac{(Y_i- \mu(X_i,P_i))F^\pi(P_i|X_i)}{{\pi}_D(P_i|X_i)} \left(\int_0^\infty \pi(p | X_i) \int_{p}^\infty  \mu(X_i, z)dzdp\right)^{r-1} \\
 + \left(\int_0^\infty\pi(p | X_i) \int_{p}^\infty  \mu(X_i, z)dzdp \right)^r \Bigg]= o_p(n^{-1/2}),
\end{align*}

Recall that \begin{gather*}
 \widehat h (X_i)=  \int_0^\infty\pi(p | X_i) \int_{p}^\infty \widehat \mu(X_i, z)dzdp, \mbox{ and } \widehat\omega(X_i,P_i) = \frac{F^\pi(P_i|X_i)}{\widehat{\pi}_D(P_i|X_i)} 
\end{gather*}

In what follows, we let $\mu_i$, $\omega_i$ and $h_i$ denote $\mu(X_i, P_i)$, $\omega(X_i, P_i)$, and $h(X_i)$ respectively. Where the context is clear, these symbols also apply to their corresponding estimators.

It follows that \begin{align*}
 &\widehat \calS^r(\pi)=
 \frac{1}{n}\sum_{i=1}^n \Bigg[ r \widehat\omega^{-k(i)}(X_i,P_i) (Y_i-\widehat \mu^{-k(i)}(X_i,P_i))   \widehat h^{-k(i)} (X_i)^{r-1} 
 + \widehat h^{-k(i)} (X_i)^r \Bigg] \\
 = & \frac{1}{n}\sum_{i=1}^n \Bigg[ r (\widehat \omega^{-k(i)}-\omega_i+\omega_i) (\mu_i+\varepsilon_i-\widehat \mu^{-k(i)}) \widehat h^{-k(i)} (X_i)^{r-1} 
 + \widehat h^{-k(i)} (X_i)^r \Bigg]\\
 = & \underbrace{\frac{1}{n}\sum_{i=1}^n \Bigg[ r (\widehat\omega^{-k(i)}-\omega_i) (\mu_i-\widehat \mu^{-k(i)})   \widehat h^{-k(i)} (X_i)^{r-1} \Bigg]}_{J_1}+ \underbrace{\frac{1}{n}\sum_{i=1}^n \Bigg[ r (\widehat \omega^{-k(i)}-\omega_i) \varepsilon_i \widehat h^{-k(i)} (X_i)^{r-1} \Bigg]}_{J_2}\\
  & +  \underbrace{\frac{1}{n}\sum_{i=1}^n \Bigg[ r \omega_i \varepsilon_i \widehat h^{-k(i)} (X_i)^{r-1} \Bigg]}_{J_3}+ \underbrace{\frac{1}{n}\sum_{i=1}^n \Bigg[ r \omega_i (\mu_i-\widehat \mu^{-k(i)})   \widehat h^{-k(i)} (X_i)^{r-1} \Bigg]+\frac{1}{n}\sum_{i=1}^n \widehat h^{-k(i)} (X_i)^r}_{J_4}.
\end{align*}

Here, $J_1$ represents the product error term, since $\widehat h^{-k(i)}$ is uniformly bounded, $J_1$ can be bounded using an argument similar to that used for $D_3$ in the proof of Theorem \ref{thm:dr rate}. 

As for $J_2$, it is mean zero. Also note that the term $\widehat h^{-k(i)}$ is uniformly bounded, thus $J_2$ can be bounded following the logic applied to $D_1$ in the same proof. It follows that both $J_1$ and $J_2$ are of order $o_p(n^{-1/2})$.

We now deal with $J_3$. By applying the mean value theorem to $\widehat h^{-k(i)} (X_i)^{r-1}$, we have \begin{gather*}
    \widehat h^{-k(i)} (X_i)^{r-1}=h_i^{r-1}+(r-1)\dot h_i ^{r-2} ( \widehat h^{-k(i)} (X_i)-h_i),
\end{gather*} where $\dot h_i$ is the intermediate value, i.e., $\dot h_i=t\widehat h^{-k(i)} (X_i)+(1-t)h_i$, for some $0<t<1$.

Thus we have \begin{align*}
    &J_3=\frac{1}{n}\sum_{i=1}^n  r \omega_i \varepsilon_i \widehat h^{-k(i)} (X_i)^{r-1}\\
    = & \frac{1}{n}\sum_{i=1}^n  r \omega_i \varepsilon_i h_i (X_i)^{r-1}+\frac{1}{n}\sum_{i=1}^n  r(r-1) \omega_i \varepsilon_i \dot h_i ^{r-2} ( \widehat h^{-k(i)} (X_i)-h_i).
\end{align*}  Since $\widehat h^{-k(i)} (X_i)$ and $h_i$ are uniformly bounded, $\bar h_i$ is also bounded. Thus, the second term in the above equation is $o_p(n^{-1/2})$, since it has mean zero and $\widehat h^{-k(i)}(X_i)$ has $L_2$ consistency (the argument parallels the bound for $J_2$ above). Thus we have shown that \begin{align*}
    J_3= \frac{1}{n}\sum_{i=1}^n  r \omega_i \varepsilon_i h_i (X_i)^{r-1}+o_p(n^{-1/2}).
\end{align*}

We finally deal with $J_4$. Recall that \begin{gather*}
    J_4=\frac{1}{n}\sum_{i=1}^n \Bigg[ r \omega_i (\mu_i-\widehat \mu^{-k(i)})   \widehat h^{-k(i)} (X_i)^{r-1} \Bigg]+\frac{1}{n}\sum_{i=1}^n \widehat h^{-k(i)} (X_i)^r.
\end{gather*}

We first apply the mean value theorem to $\widehat h^{-k(i)} (X_i)^{r}$,  \begin{gather} \label{eq: mvt 1}
    \widehat h^{-k(i)} (X_i)^{r}=h_i^{r}+r\bar h_i ^{r-1} ( \widehat h^{-k(i)} (X_i)-h_i),
\end{gather} where $\bar h_i$ is the intermediate value.

Then we apply the mean value theorem again to $\widehat h^{-k(i)} (X_i)^{r-1}$, we have \begin{gather}\label{eq: mvt 2}
    \widehat h^{-k(i)} (X_i)^{r-1}=\bar h_i^{r-1}+(r-1)\widetilde h_i ^{r-2} ( \widehat h^{-k(i)} (X_i)-\bar h_i),
\end{gather} where $\widetilde h_i$ is intermediate value between $\widehat h^{-k(i)} (X_i)$ and $\bar h_i$.

Now plug Equations \eqref{eq: mvt 1} and \eqref{eq: mvt 2} into term $J_4$, we have 
    \begin{align*}
    & J_4=\frac{1}{n}\sum_{i=1}^n \Bigg[ r \omega_i (\mu_i-\widehat \mu^{-k(i)})   \widehat h^{-k(i)} (X_i)^{r-1} \Bigg]+\frac{1}{n}\sum_{i=1}^n \widehat h^{-k(i)} (X_i)^r\\
    =\; & \frac{1}{n}\sum_{i=1}^n \Bigg[ r \omega_i (\mu_i-\widehat \mu^{-k(i)})   (\bar h_i^{r-1}+(r-1)\widetilde h_i ^{r-2} ( \widehat h^{-k(i)} (X_i)-\bar h_i)) \Bigg]\\
    & +\frac{1}{n}\sum_{i=1}^n r\bar h_i ^{r-1} ( \widehat h^{-k(i)} (X_i)-h_i)+\frac{1}{n}\sum_{i=1}^n h_i^{r}\\
    =\; & \underbrace{\frac{1}{n}\sum_{i=1}^n  r \omega_i (\mu_i-\widehat \mu^{-k(i)})   \bar h_i^{r-1}}_{I_1}+\underbrace{\frac{1}{n}\sum_{i=1}^n \Bigg[ r(r-1) \omega_i (\mu_i-\widehat \mu^{-k(i)})   \widetilde h_i ^{r-2} ( \widehat h^{-k(i)} (X_i)-\bar h_i) \Bigg]}_{I_2}\\
    & +\underbrace{\frac{1}{n}\sum_{i=1}^n r\bar h_i ^{r-1} ( \widehat h^{-k(i)} (X_i)-h_i)}_{I_3}+\frac{1}{n}\sum_{i=1}^n h_i^{r}.
\end{align*}

We first deal with the second term $I_2$, since $\omega_i$ and $\widetilde h_i$ are uniformly bounded, by applying the Cauchy-Schwarz, we have 
\begin{align*}
    I_2 & \leq C \sqrt{\frac{1}{n}\sum_{i=1}^n  (\mu_i-\widehat \mu^{-k(i)})^2}  \sqrt{\frac{1}{n}\sum_{i=1}^n   ( \widehat h^{-k(i)} (X_i)-\bar h_i)^2} \\
    & \leq C \sqrt{\frac{1}{n}\sum_{i=1}^n  (\mu_i-\widehat \mu^{-k(i)})^2}  \sqrt{\frac{1}{n}\sum_{i=1}^n   ( \widehat h^{-k(i)} (X_i)-h_i)^2} =O_p(n^{-2\alpha_1})=o_p(n^{-1/2}),
\end{align*} since $\alpha_1>1/4.$

It remains to show that $I_1+I_3=o_p(\sqrt{1/n}).$ By the balancing property in Lemma \ref{lemma: balancing}, $I_1+I_3$ has mean zero. It thus suffices to show that \begin{gather*}
  \E\left[  r \omega_i (\mu_i-\widehat \mu^{-k(i)})   \bar h_i^{r-1}\right]^2=o(1), \mbox{ and } 
  \E\left[r\bar h_i ^{r-1} ( \widehat h^{-k(i)} (X_i)-h_i)\right]^2=o(1).
\end{gather*}
Similar to bound $J_2$ before, because  $\widehat \mu^{-k(i)}$ and $\widehat h^{-k(i)}$ are both $L_2$ consistent (Lemma \ref{lem:error transform}), the above two equations hold. Thus we have $J_4=\frac{1}{n}\sum_{i=1}^n h_i^{r}+o_p(\sqrt{1/n})$.

To summarize, we have shown that \begin{gather*}
    \widehat \calS^r(\pi)=J_3+J_4+o_p(n^{-1/2})=\frac{1}{n}\sum_{i=1}^n \left[ \underbrace{h_i^{r}+r \omega_i \varepsilon_i h_i (X_i)^{r-1}}_{\text{uncentered EIF for }\calS^r(\pi)}\right]+o_p(n^{-1/2}).
\end{gather*} This thus completes the proof.
\end{proof}

\section{Asymptotic Normality for Inequality-Aware Surplus for the Behavior Policy}
\label{sec:normality_unequal_behavioral}

We next present the result for inequality-aware surplus for the behavior policy.

\begin{assumption} \label{asmp: aware behavior}
Assume $\pi_D(p \given x)>c$, for all $p \in \calP$, and every $x$, for some constant $c$. In addition, suppose that the estimators for the demand function and the behavior policy are constructed using the cross-fitting procedure, and that they achieve the following convergence rate: 
    \begin{align} \label{eq: product rate}
        \sqrt{\E[(\widehat \mu(X,P)-\mu(X,P))^2]} =O_p(n^{-\alpha_1}), \mbox{ and } \; \sqrt{\E[(\widehat \pi_D(P|X)-\pi_D(P|X))^2]}=O_p(n^{-\alpha_2}),
    \end{align} with $\alpha_1>1/4$, and  $\alpha_2>1/4$.
\end{assumption}

\begin{theorem} \label{thm:ia behavior}
    Suppose that Assumptions \ref{assumption:ignorability}, \ref{assumption:overlap}, and \ref{asmp: aware behavior} hold,  then
    \begin{align*}
       \sqrt{n} (\widehat \calS^r(\pi_D)- \calS^r(\pi_D))\rightarrow \mathcal{N}(0,\Sigma^r({\pi_D})),
    \end{align*} where $\Sigma^r({\pi_D})$ is the variance of the EIF of $\calS^r(\pi_D)$. 
\end{theorem}

\begin{proof}
To establish the asymptotic normality of $\widehat{\mathcal{S}}^r(\pi_D)$, we show that the estimator is equivalent to the sample average of its efficient influence function (EIF) up to an $o_p(n^{-1/2})$ remainder. 

Recall that \begin{align*}
       \nonumber   \widehat \calS^r&(\pi_D) = \frac{1}{n}\sum_{i=1}^n \Bigg[r  \left(\frac{(Y_i-\widehat \mu^{-k(i)}(X_i,P_i))\widehat F^{\pi_D}(P_i|X_i)}{\widehat\pi_D(P_i|X_i)} + \int_{P_i} \widehat \mu^{-k(i)}(X_i,z)dz \right)\\
& \times \left(\int_0^\infty \widehat \pi_D(p | X_i) \int_{p}^\infty \widehat \mu^{-k(i)}(X_i, z)dzdp \right)^{r-1} 
        + (1-r)\left( \int_0^\infty \widehat\pi_D(p | X_i) \int_{p}^\infty \widehat \mu^{-k(i)}(X_i, z)dzdp \right)^r\Bigg].
    \end{align*}

Let \begin{gather*}
    \widehat g_i=\int_{P_i} \widehat \mu^{-k(i)}(X_i,z)dz,\\
    \widehat h_i= \int_0^\infty \widehat\pi_D(p | X_i) \int_{p}^\infty \widehat \mu^{-k(i)}(X_i, z)dzdp, \\
    \widetilde h_i= \int_0^\infty \pi_D(p | X_i) \int_{p}^\infty \widehat \mu^{-k(i)}(X_i, z)dzdp.
\end{gather*}

Note that $|\widehat h-\widetilde h|_{L_2}=|\widehat \pi_D-\pi_D|_{L_2}=O_p(n^{-\alpha_2})$, thus $|\widehat h_i- h_i|_{L_2}=O_p(n^{-\alpha_1}+n^{-\alpha_2})$

To simplify notation, we suppress the $-k(i)$ subscript, noting that nuisance functions are always trained on out-of-sample observations.

Thus, we have 
\begin{align*}
       \widehat \calS^r(\pi_D) =& \frac{1}{n}\sum_{i=1}^n \Bigg[r  \left((Y_i-\widehat \mu_i)\widehat \omega_i + \widehat g_i \right) \widehat h_i^{r-1} + (1-r)\widehat h_i^r\Bigg]\\
       =&\frac{1}{n}\sum_{i=1}^n \Bigg[r  \left((\varepsilon_i+\mu_i-\widehat \mu_i)(\widehat \omega_i-\omega_i+\omega_i) + \widehat g_i \right) \widehat h_i^{r-1} + (1-r)\widehat h_i^r\Bigg]\\
       =&\underbrace{\frac{1}{n}\sum_{i=1}^n r  \varepsilon_i(\widehat \omega_i-\omega_i)\widehat h_i^{r-1}}_{o_p(n^{-1/2})} +\underbrace{\frac{1}{n}\sum_{i=1}^n r  (\mu_i-\widehat\mu_i)(\widehat \omega_i-\omega_i)\widehat h_i^{r-1}}_{o_p(n^{-1/2})}+\underbrace{\frac{1}{n}\sum_{i=1}^n r \varepsilon_i\omega_i\widehat h_i^{r-1}}_{\frac{1}{n}\sum_{i=1}^n r \varepsilon_i\omega_i h_i^{r-1}+o_p(n^{-1/2})} \\
       &+\frac{1}{n}\sum_{i=1}^n r \omega_i ( \mu_i-\widehat\mu_i) \widehat h_i^{r-1}+\frac{1}{n}\sum_{i=1}^n r \widehat g_i \widehat h_i^{r-1}+\frac{1}{n}\sum_{i=1}^n (1-r) \widehat h_i^{r}.
    \end{align*}

The first three terms of the above equation can be bounded by applying the exact techniques used for terms $J_1$, $J_2$ and $J_3$ in the proof of Theorem \ref{thm: aware rate}.

It follows that
\begin{align*}
\widehat \calS^r(\pi_D) =\frac{1}{n}\sum_{i=1}^n r \varepsilon_i\omega_i h_i^{r-1} +\underbrace{\frac{1}{n}\sum_{i=1}^n r \omega_i ( \mu_i-\widehat\mu_i) \widehat h_i^{r-1}+\frac{1}{n}\sum_{i=1}^n r \widehat g_i \widehat h_i^{r-1}+\frac{1}{n}\sum_{i=1}^n (1-r) \widehat h_i^{r}}_{Q_1}+o_p(n^{-1/2}).
\end{align*}

We now analyze $Q_1.$

\begin{align*}
Q_1 &=\frac{1}{n}\sum_{i=1}^n r \omega_i ( \mu_i-\widehat\mu_i) \widehat h_i^{r-1}+\frac{1}{n}\sum_{i=1}^n r \widehat g_i \widehat h_i^{r-1}+\frac{1}{n}\sum_{i=1}^n (1-r) \widehat h_i^{r}\\
=& \underbrace{\frac{1}{n}\sum_{i=1}^n r \omega_i ( \mu_i-\widehat\mu_i) \widehat h_i^{r-1}+\frac{1}{n}\sum_{i=1}^n r (\widehat g_i-g_i) \widehat h_i^{r-1}}_{E_1} +\underbrace{\frac{1}{n}\sum_{i=1}^n r  g_i \widehat h_i^{r-1}+\frac{1}{n}\sum_{i=1}^n (1-r) \widehat h_i^{r}}_{E_2}\\ 
\end{align*}

We next show the first term $E_1=o_p(n^{-1/2})$. Note that $E_1$ has zero mean because:
\begin{equation*}
\mathbb{E}[\omega(\mu - \hat{\mu})] = \mathbb{E}[h - \widetilde{h}] = \mathbb{E}[g - \hat{g}],
\end{equation*}
where the first equality follows from Lemma \ref{lemma: balancing} and the second arises from the two different representations of demand. By Chebyshev's inequality, it thus suffices to show that 
\begin{gather*}
    \E\left[r\omega ( \mu-\widehat\mu) \widehat h^{r-1}+r (\widehat g-g) \widehat h^{r-1} \right]^2=o(1).
\end{gather*}
Since $(a+b)^2 \leq 2(a^2+b^2)$,
 it is therefore sufficient to show that
\begin{gather*}
    \E\left[r\omega ( \mu-\widehat\mu) \widehat h^{r-1}\right]^2=o(1), \mbox{ and } \E\left[ r (\widehat g-g) \widehat h^{r-1}\right]^2=o(1).
\end{gather*} Since $r$, $\omega$, and $\widehat h$ are bounded, and $\widehat \mu$ and $\widehat g$ are $L_2$ consistent, the above two equations hold. Thus, we have \begin{gather*}
    Q_1=E_2+o_p(n^{-1/2}).
\end{gather*}

We now deal with $E_2$. It follows that \begin{align*}
   E_2=& \frac{1}{n}\sum_{i=1}^n r  g_i \widehat h_i^{r-1}+\frac{1}{n}\sum_{i=1}^n (1-r) \widehat h_i^{r}\\
   =& \frac{1}{n}\sum_{i=1}^n r  g_i \left[h_i^{r-1} +(r-1)\dot h_i^{r-2}(\widehat h_i-h_i) \right]+\frac{1}{n}\sum_{i=1}^n (1-r) \left[ h_i^{r} +r\bar h_i^{r-1}(\widehat h_i-h_i) \right],
\end{align*} where both $\dot h_i$ and $\bar h_i$ are the intermediate values between $\widehat h_i$ and $h_i$. It follows that \begin{align*}
   E_2=\underbrace{\frac{1}{n}\sum_{i=1}^n r  g_i h_i^{r-1} +
   \frac{1}{n}\sum_{i=1}^n (1-r)  h_i^{r}}_{\text{part of the EIF of} \calS^r(\pi_D)} +r(r-1) \left[\underbrace{\frac{1}{n}\sum_{i=1}^n   g_i \dot h_i^{r-2}(\widehat h_i-h_i) -\frac{1}{n}\sum_{i=1}^n  \bar h_i^{r-1}(\widehat h_i-h_i)}_{G_1}\right],
\end{align*}

It remains to show that $G_1=o_p(n^{-1/2})$. It follows that \begin{align*}
    G_1=&\frac{1}{n}\sum_{i=1}^n   (\widehat h_i-h_i) \left(g_i \dot h_i^{r-2}-\bar h_i^{r-1} \right) \\
    =&\frac{1}{n}\sum_{i=1}^n   (\widehat h_i-h_i) \left[g_i \left(h_i^{r-2} +(r-2) \ddot h_i^{r-3}(\dot h_i-h_i)\right) -\left(h_i^{r-1}+(r-1)\check h_i^{r-2}(\bar h_i-h_i) \right) \right],
\end{align*} where again $\ddot h_i$ and $\check h_i$ are the intermediate values. Then we have 
\begin{align*}
    G_1=&\underbrace{\frac{1}{n}\sum_{i=1}^n   (\widehat h_i-h_i)   (g_i h_i^{r-2}-h_i^{r-1})}_{H_1} +\underbrace{\frac{1}{n}\sum_{i=1}^n  (r-2) g_i\ddot h_i^{r-3} (\widehat h_i-h_i)  (\dot h_i-h_i)}_{H_2}\\
    & +\underbrace{\frac{1}{n}\sum_{i=1}^n  (r-1)\check h_i^{r-2}(\widehat h_i-h_i) (h_i-\bar h_i)}_{H_3}
\end{align*}

Here, $H_1$ is zero mean, because $\E(g|X)=h(X)$. Thus, we have $$\E[(\widehat h_i-h_i)^2(g_i h_i^{r-2}-h_i^{r-1})^2]\leq C \E[(\widehat h_i-h_i)^2(g_i h_i^{r-2}-h_i^{r-1})^2]=o(1),$$ and by the Chebechev's inequality, $H_1=o_p(n^{-1/2})$ 

Then, we have \begin{align*}
    H_2=&\frac{1}{n}\sum_{i=1}^n  (r-2) g_i\ddot h_i^{r-3} (\widehat h_i-h_i)  (\dot h_i-h_i) \\
    \lesssim & \sqrt{\frac{1}{n}\sum_{i=1}^n  (\widehat h_i-h_i)^2  }\sqrt{\frac{1}{n}\sum_{i=1}^n    (\dot h_i-h_i)^2}\\
    \lesssim & \frac{1}{n}\sum_{i=1}^n    (\widehat h_i-h_i)^2=O_p\left(n^{-2\alpha_1}\right)+O_p\left(n^{-2\alpha_2}\right).
\end{align*} Here, $|\widehat h_i- h_i|_{L_2}=O_p(n^{-\alpha_1}+n^{-\alpha_2})$, since the estimation of $h_i$ also requires the estimation of $\pi_D$. By Assumption \ref{asmp: aware behavior}, $\alpha_1>1/4$, and $\alpha_2>1/4$. Thus, $H_2=o_p(n^{-1/2}).$

Similarly, we have \begin{align*}
H_3=O_p\left(n^{-2\alpha_1}\right)+O_p\left(n^{-2\alpha_2}\right)=o_p(n^{-1/2}).
\end{align*}

To summarize, we have \begin{align*}
\widehat \calS^r(\pi_D)  =\underbrace{\frac{1}{n}\sum_{i=1}^n r \varepsilon_i\omega_i h_i^{r-1} +\frac{1}{n}\sum_{i=1}^n r  g_i h_i^{r-1} +
   \frac{1}{n}\sum_{i=1}^n (1-r)  h_i^{r}+\calS^r(\pi_D)}_{\text{EIF for } \calS^r(\pi_D) }+o_p(n^{-1/2})
\end{align*}

By the Central Limit Theorem, and the  Slutsky's theorem, we have 
$$ \sqrt{n}(\widehat{\mathcal S}^r(\pi_D) - \mathcal{S}^r(\pi_D)) \xrightarrow{d} \mathcal{N}(0, \Sigma^r(\pi_D)). $$

\end{proof}

\section{Partial Identification}
\label{app:partial_identification}

In practice, the overlap assumption (Assumption \ref{assumption:overlap}), which requires that every price we wish to evaluate has a positive probability of being assigned under the historical policy, may fail when the firm is unwilling to conduct extensive price experimentation. This may occur when regularly updating prices is logistically challenging (such as in brick-and-mortar retail), in scenarios where changing prices risks customer backlash, or when firms do not want to sacrifice short-term profit.
In this case, the surplus is not point-identified—recall Definition \ref{def:partial} for the formal definition. An alternative choice is to establish partial identification bounds for the surplus estimation.  For example, if historical data only contains prices between \$10 and \$20, but the target policy we wish to evaluate includes prices at \$5 or \$25, the demand in these unobserved regions is unknown. Partial identification provides a credible range (i.e., a minimum and maximum possible value) for the true surplus, rather than a single, unreliable point estimate. 
To tighten the identification region, we impose a mild regularity condition on the valuation distribution.

\begin{assumption}[Log-concavity]\label{ass: log-concavity}
    For all $x$, the demand function $\mu(x,\cdot)$ is log-concave. 
\end{assumption}

The log-concavity assumption implies that the hazard rate is monotone, which is a common assumption in the pricing literature \citep{cole2015sample,huang2018making,allouah2021revenue}. It encompasses a broad range of valuation distributions, including normal, exponential, and uniform \citep{bagnoli2005log}. Importantly, it rules out pathological ``thick tailed'' beliefs that would yield implausibly large surplus in the unobserved price regions. We also note that equation (\ref{model: true model}) directly implies that the purchase probabilities are monotonically decreasing, which further restricts the set of feasible demand functions. Given these conditions, we next present how to construct the partial identification bounds when the overlap assumption is violated.

\subsection{Partial Identification Bound}

We begin by briefly outlining our main idea: consider the purchase probability curve $\mu(x,z)$. 
In regions where the firm experimented, we estimate this curve nonparametrically.  
In regions without experimentation, the curve is unknown but must lie within the smallest region consistent with Assumption~\ref{ass: log-concavity} while interpolating the observed points.

\subsubsection{Lower Envelope} 
We denote by $z_1$ and $z_2$ the closest observed prices to the left and right of $z$ for covariate value $x$. If no observation lies to the left of $z$, we invoke the boundary condition $z_1 = 0$, implying $\mu(x,0)=1$ for all $x$. Conversely, if there is no observation to the right of $z$, we set $z_2 = V_{\max}$, where $V_{\max}$ is the smallest price at which $Y=0$ almost surely, so that $\mu(x, V_{\max}) = 0$, for all $x$.
Because $\log \mu(x,\cdot)$ is concave, any chord connecting $(z_1,\log \mu(x, z_1))$ and $(z_2,\mu(x, z_2))$ lies below the graph. Taking the maximum of that chord with the last observed point to the left yields the tightest feasible lower bound at $z$. %
Formally, Lemma~\ref{lower_bound_partial} establishes this lower bound.

\begin{lemma}[Lower Bound]
\label{lower_bound_partial}
 Under the binary purchasing model in Equation \eqref{model: true model}, Assumptions \ref{assumption:ignorability},  and \ref{ass: log-concavity}, we have $F_{l}(z, z_1,z_2,x) \equiv e^{l(z, z_1,z_2,x)} \leq \mu(x, z)$, for all $x$, where 
\begin{align}
    l(z, z_1,z_2,x) = \frac{z-z_2}{z_1-z_2}\log(\mu(x, z_1)) + \frac{z_1-z}{z_1-z_2}\log(\mu(x, z_2)).
\end{align}
\end{lemma}

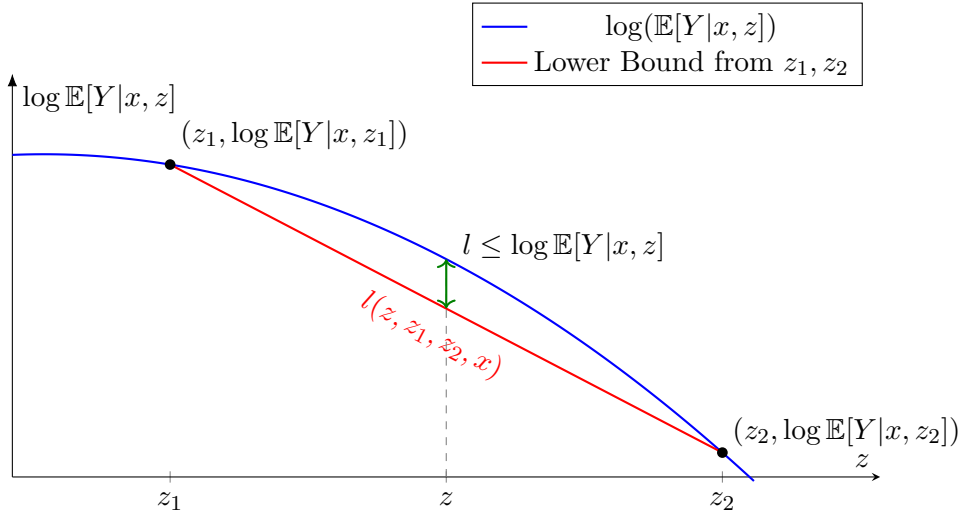
\begin{figure}
    \centering
\begin{tikzpicture}
    \def\myfuncLower(#1){4 - 0.2*((#1)-0.2)^2}

    \begin{axis}[
        axis x line=middle, 
        axis y line=none,
        xlabel=$z$,
        ytick=\empty,
        xtick={1, 2.75, 4.5},
        xticklabels={$z_1$, $z$, $z_2$},
        xmin=0, xmax=5.5,
        ymin=0, ymax=6,
        width=13cm,
        height=8cm,
        legend style={at={(0.98,0.98)}, anchor=north east},
        clip=false
    ]
    \draw[-{Latex}] (axis cs:0, 0) -- (axis cs:0, 5) node[below right] {$\log \E[Y|x,z]$};

    \addplot[
        domain=0:4.7,
        samples=100,
        thick,
        blue,
    ] {\myfuncLower(x)};
    \addlegendentry{$\log(\E[Y|x,z])$}

    \coordinate (P1) at (axis cs:1, {\myfuncLower(1)});
    \coordinate (P2) at (axis cs:4.5, {\myfuncLower(4.5)});

    \addplot[thick, red, mark=none] coordinates {
        (1, 3.872) 
        (4.5, 0.302)
    } node[midway, below, sloped, yshift=-2pt] {$l(z, z_1, z_2, x)$};
    \addlegendentry{Lower Bound from $z_1,z_2$}

    \fill[black] (P1) circle (2pt);
    \node[above=2pt, anchor=south west] at (P1) {$(z_1, \log \E[Y|x,z_1])$};

    \fill[black] (P2) circle (2pt);
    \node[above=8pt, anchor=south east,right] at (P2) {$(z_2, \log \E[Y|x,z_2])$};

    \def\zVal{2.75}
    \coordinate (PointOnCurve) at (axis cs:\zVal, {\myfuncLower(\zVal)});

    \pgfmathsetmacro{\yChord}{3.875 + (0.302 - 3.875)/(4.5 - 1) * (\zVal - 1)}
    \coordinate (PointOnChord) at (axis cs:\zVal, \yChord);
    \draw[gray, dashed] (axis cs:\zVal, 0) -- (PointOnCurve);
    \draw[<->, thick, green!50!black] (PointOnChord) -- (PointOnCurve)
        node[midway, right, text=black, xshift=2pt,yshift=14pt] {$l \leq \log \E[Y|x,z]$};
    \end{axis}
\end{tikzpicture}
    \caption{Lower Bound}
    \label{fig:lower_bound}
\end{figure}

This is proven in Appendix \ref{sec:proof_lower_bound_partial}. Intuitively, the bound lets the demand curve fall as steeply as the concavity constraint permits while still passing through the neighbouring data points. The graphical illustration of the lower bound is shown in \ref{fig:lower_bound}. 

\subsubsection{Upper envelope}  The upper bound combines two concavity constraints, one interpolating the closest pair on the left $(z_1,z_2)$ and another interpolating the closest pair on the right $(z_3,z_4)$. The bound at $z$ is the minimum of these two extrapolations as shown in Lemma~\ref{lemma:upper}. When only one side is observed, we fall back on the boundary conditions at $0$ or $V_{\max}$ to close the gap. The resulting envelope prevents the unobserved segment of the curve from bending downwards too slowly, which would otherwise generate unrealistically high surplus.

Now let $z_1\leq z_2\leq z \leq z_3 \leq z_4$, if there is no empirical observed point smaller (larger) than $z$, then $z_1=\text{NA}, z_2=0$ ($z_3=V_\text{max}, z_4=\text{NA} $). 
    If there is only one point $z'$ smaller (bigger) than $z$, then we can set $z_1=0,z_2=z'$ ($z_3=z', z_4=V_\text{max}$). 
    Define 
\begin{align*}
    u_1(z_1,z_2,x) = \log( \mu(x, z_2)) + \frac{z - z_2}{z_2 - z_1}(\log(\mu(x, z_2)) - \log(\mu(x, z_1))),\\
    u_2(z_3,z_4,x) = \log(\mu(x, z_3)) - \frac{z_3 - z}{z_4 - z_3}(\log(\mu(x, z_4)) - \log(\mu(x, z_3))).
\end{align*} We formally state the upper bound in the subsequent lemma.

\begin{lemma}[Upper Bound]\label{lemma:upper}
 Under the assumptions in Lemma \ref{lower_bound_partial},  we have
$ \mu(x,z) \leq F_u(z_1,z_2,z_3,z_4,x)$, for all $x$, where 
    \begin{align*}
    F_u(z_1,z_2,z_3,z_4, x) = 
    \begin{cases}
        \min(e^{u_1},e^{u_2})& z_1, z_2, z_3, z_4 \neq \text{NA}\\
        \min(e^{u_1}, \mu(x, z_2)) & z_1, z_2, z_3 \neq \text{NA}, z_4 = \text{NA}\\ 
        \min(e^{u_2}, \mu(x, z_2)) & z_2, z_3, z_4 \neq \text{NA}, z_1 = \text{NA}\\ 
        1 & \text{otherwise}
    \end{cases}.
\end{align*}
\end{lemma}

\subsubsection{Combined Bound}
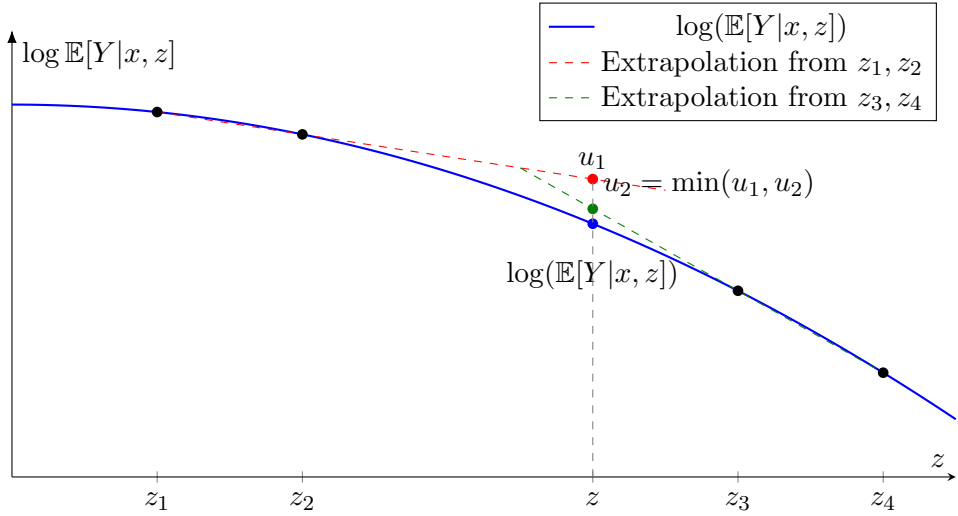
\begin{figure}
    \centering
\begin{tikzpicture}
\def\myfunc(#1){10 - 0.2*((#1)-0)^2}
\begin{axis}[
        axis x line=middle, 
        axis y line=none,
    xlabel=$z$,
    xtick={1, 2, 4, 5, 6},
    xticklabels={$z_1$, $z_2$, $z$, $z_3$, $z_4$},
    ytick=\empty,
    xmin=0, xmax=6.5,
    ymin=0, ymax=13, %
        width=14cm,
        height=8cm,
    legend style={at={(0.98,0.98)}, anchor=north east},
    clip=false
    ]
    \draw[-{Latex}] (axis cs:0, 0) -- (axis cs:0, 12) node[below right] {$\log \E[Y|x,z]$};

    \addplot[
        domain=0:6.5,
        samples=100,
        thick,
        blue,
    ] {\myfunc(x)};
    \addlegendentry{$\log(\E[Y|x,z])$}

    \coordinate (z1) at (1, {\myfunc(1)});
    \coordinate (z2) at (2, {\myfunc(2)});
    \coordinate (z_actual)  at (4, {\myfunc(4)});
    \coordinate (z3) at (5, {\myfunc(5)});
    \coordinate (z4) at (6, {\myfunc(6)});

    \addplot[
        domain=1:4.5,
        samples=2,
        dashed,
        red,
    ] {-0.6*x + 10.4};
    \addlegendentry{Extrapolation from $z_1, z_2$}
    
    \coordinate (u1) at (4, 8);

    \addplot[
        domain=3.5:6,
        samples=2,
        dashed,
        green!50!black,
   ] {-2.2*x + 16};
    \addlegendentry{Extrapolation from $z_3, z_4$}
    
    \coordinate (u2) at (4, 7.2);
    
    \fill (z1) circle (2pt);
    \fill (z2) circle (2pt);
    \fill (z3) circle (2pt);
    \fill (z4) circle (2pt);
    
    \fill[blue] (z_actual) circle (2pt) node[below=10pt, black] {$\log(\E[Y|x,z])$};
    \fill[red] (u1) circle (2pt) node[above, black] {$u_1$};
    \fill[green!50!black] (u2) circle (2pt) node[above right, black] {$u_2=\min(u_1, u_2)$};
    \draw[gray, dashed] (axis cs:4,0) -- (u1);

\end{axis}
\end{tikzpicture}
    \caption{Upper Bound}
    \label{fig:upper_bound}
\end{figure}

Given Lemmas \ref{lower_bound_partial} and \ref{lemma:upper}, the lower   and upper bounds then can be estimated by 
$\widehat{\cal S}_{-}$ and $\widehat{\cal S}_{+}$, respectively, as defined below, 
\begin{align*}
    & \widehat {\calS}_{-} = \frac{1}{n} \sum_{i=1}^n \int_{z=0}^{\infty} 
     \Big\{\mathbb{I}[\pi_D(z|X_i)\neq 0]\hat{\E}(Y|P=z,X_i) F^\pi(z|X_i)  + \mathbb{I}[\pi_D(z|X_i)= 0] \widehat{F}_l(z_2(z),z_3(z),X_i)F^\pi(z|X_i)\Big\}dz, 
\end{align*} 
\begin{align*}
    \widehat {\calS}_{+} = \frac{1}{n} \sum_{i=1}^n \int_{z=0}^{\infty} 
     \Big\{&\mathbb{I}[\pi_D(z|X_i)\neq 0]\hat{\E}(Y|P=z,X_i)F^\pi(z|X_i) \\ 
     &+ \mathbb{I}[\pi_D(z|X_i)= 0]\widehat{F}_{u}(z_1(z),z_2(z),z_3(z),z_4(z),X_i)F^\pi(z|X_i)\Big\}dz
\end{align*}
where $z_1(z)\leq z_2(z)\leq z \leq z_3(z) \leq z_4(z)$ are the closest empirical observations around $z$ given $x$. 
Here $\widehat{F}_l$ and $\widehat{F}_u$ are using the estimated demand function $\widehat{\mu}$ compared to $F_l$ and $F_u$. 
If there is no empirical observation, we can use 0 and $V_\text{max}$ as $z_2$ and $z_3$ and set $z_1, z_4$ as NA. We can get the similar CPW and ACPW estimators by replacing the first part of the estimator when $\pi_D(z|X)\neq 0$. The graphical illustration of the upper bound is shown in \ref{fig:upper_bound}. 

\subsection{Statistical Properties}

Next we provide the estimation error for the estimated partial identification bounds. Let $z_1 = \max_{\pi_D(\tilde{z})>0, \tilde{z}\leq z} \tilde{z}$ and $z_2 = \min_{\pi_D(\tilde{z})>0, \tilde{z}\geq z} \tilde{z}$, also define 
\begin{align*}
 {\calS}_{-}^* = \mathbb{E} \int_{z=0}^{\infty} 
     \Big\{\mathbb{I}[\pi_D(z|X)\neq 0]\E[Y\mid P=z,X]F^\pi(z|X)  + \mathbb{I}[\pi_D(z|X)= 0]{F}_{l}(z_1,z_2F^\pi(z|X)\Big\}dz,
\end{align*}

\begin{align*}    
     {\calS}_{+}^* = & \mathbb{E} \int_{z=0}^{\infty} 
     \Big\{\mathbb{I}[\pi_D(z|X)\neq 0]\E[Y\mid P=z,X]F^\pi(z|X)  + \mathbb{I}[\pi_D(z|X)= 0]{F}_{u}(z_1,z_2,z_3,z_4)F^\pi(z|X)\Big\}dz.
\end{align*} We then have the following result.

\begin{theorem}
\label{thm:partial_lower}
Suppose that the conditions in Lemma \ref{lower_bound_partial} hold. Furthermore, assume the following convergence rates for the nuisance parameters:
\begin{enumerate}
\item $\sqrt{\mathbb{E}[(\widehat{\mu}(x,p) - \mu(x,p))^2]} = O(n^{-\alpha_1})$,
\item $\sqrt{\mathbb{E}[(\widehat{F}_{l}(z_1,z_2,x) - F_{l}(z_1,z_2,x))^2]} = O(n^{-\alpha_2})$, and 
$\sqrt{\mathbb{E}[(\widehat{F}_{u}(z_1,z_2,z_3,z_4, x) - F_{u}(z_1,z_2,z_3,z_4,x))^2]} = O(n^{-\alpha_2})$, 
\item The matching discrepancy satisfies $\sqrt{\mathbb{E}[(F_{l}(z_1,z_2,x) - F_{l}(z_1(z),z_2(z),x))^2]} = O(n^{-\alpha_3})$, and $\sqrt{\mathbb{E}[(F_{u}(z_1,z_2,z_3,z_4,x) - F_{u}(z_1(z),z_2(z),z_3(z),z_4(z),x))^2]} = O(n^{-\alpha_3})$ .
\end{enumerate}
Then, we have:
\begin{align*}
(i); |\widehat{\mathcal{S}}_{-} - \mathcal{S}_{-}| &= O_p(n^{-\min \{\alpha_1,\alpha_2,\alpha_3,1/2\}}); \
(ii) ; |\widehat{\mathcal{S}}_{+} - \mathcal{S}_{+}| = O_p(n^{-\min \{\alpha_1,\alpha_2,\alpha_3,1/2\}}).
\end{align*}
\end{theorem}

Theorem \ref{thm:partial_lower} establishes the consistency of our proposed estimators for the upper and lower bounds in regions where the overlap assumption is violated and is proven in Appendix \ref{sec:proof_partial_lower}. Practically, this result is crucial for managers seeking to assess the impact of future pricing policies involving price points that have not been historically tested.However, a key distinction arises in the convergence properties. While the ACPW estimator for the point-identified region can achieve the parametric rate of $O_p(n^{-1/2})$ via orthogonality, the partial identification bounds in Theorem \ref{thm:partial_lower} rely on the direct method. In the absence of overlap, we cannot leverage propensity scores to debias the estimate. Consequently, the convergence rate of the bound estimators is dominated by the estimation error of the underlying nuisance functions: the demand learner $\widehat{\mu}$, the bound learner $\widehat{F}_l$, and the matching discrepancy (data samples close to the boundary). Specifically, the error rate is $O_p(n^{-\min(\alpha_1, \alpha_2, \alpha_3,1/2)})$, where these alphas represent the convergence rates of the nuisance components. As outlined in Assumption \ref{asmp: ACPW}, flexible nonparametric machine learning models often achieve rates slower than $n^{-1/2}$. Therefore, the resulting bounds will generally converge at a slower, nonparametric rate compared to the point estimates in the overlap region.

\subsection{Experiments for Partial Identification} \label{app:exp_partial}

To evaluate our partial identification method, we designed a simulation where the overlap assumption is intentionally violated. In this setup, we generate a feature vector $X\sim U\{0,1\}^{d}, \beta \sim U[-1,1]^{d}, d=10,  V=100+300\beta^T X + \epsilon, \epsilon\sim U[0,10], Y=\mathbb{I}[V>P]$. $U\{0,1\}^{d}$ represents randomly sample $d$ bernoulli variables with a probability of 0.5. The direct model is correctly specified using a linear model. 

Crucially, the price $P$ is drawn from a distribution with a gap in its support, $P \sim U([9, 9.5] \cup [10, 10.5])$, creating regions of non-overlap $[9.5, 10]$. The target pricing policy is a uniform policy sampled from $[9.1, 9.425, 9.75, 10.075, 10.4]$, therefore $9.75$ would fall into the non-overlap region. The purchase decision is then given by $Y = \mathbb{I}[V > P]$. Our estimation correctly assumes the linear functional form of the valuation model. We report the results with 50 runs. 

We compare our proposed method against two baselines for context:
\begin{itemize}
    \item The \textbf{Naive} baseline imputes demand in non-overlapping regions with extreme values (the lower bound is 0 and the upper bound is 1).  
    \item The \textbf{Oracle} baseline serves as a theoretical benchmark by using the true, known demand function.
\end{itemize}

\begin{figure}
     \begin{subfigure}[b]{0.5\textwidth}
         \centering
         \includegraphics[width=\linewidth]{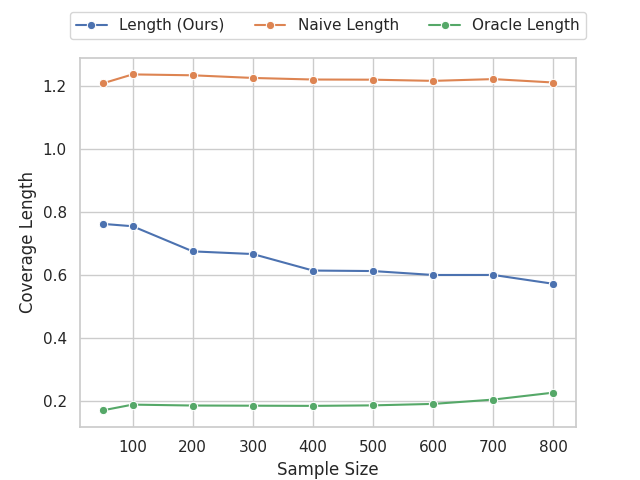}
         \caption{Bound Length}
         \label{fig:partial_length}
     \end{subfigure}
     \hfill
     \begin{subfigure}[b]{0.5\textwidth}
         \centering
         \includegraphics[width=\linewidth]{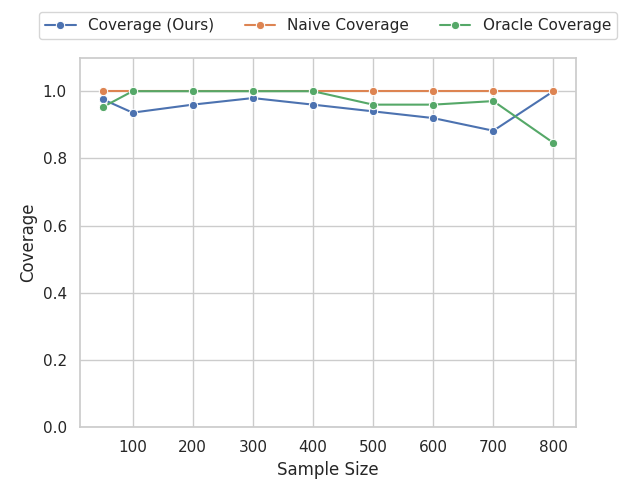}
         \caption{Bound Coverage}
         \label{fig:partial_cover}
     \end{subfigure}
     \caption{Partial Identification Bounds.}
     \label{fig:partial}
\end{figure}

We report the coverage length, which is the length of the partial identification interval for each method. 
The results are presented in \ref{fig:partial}. 
The Naive Length (orange line) remains high and constant regardless of sample size. This is expected. Without shape constraints, the naive baseline simply uses 0 and 1 as the demand lower and upper bound, resulting in a wide, uninformative interval.

The results show that our proposed bounds (blue line) are substantially tighter than the Naive approach and achieve near-perfect empirical coverage rates, confirming the robustness of our method. The oracle method (green) uses the ground-truth demand knowledge and is not affected by sample size. Empirically, our method and oracle coverage both achieve near-perfect coverage, however, we note that miscoverage may happen due to the randomness with finite samples.

\subsection{Proof of Lemma \ref{lower_bound_partial}}
\label{sec:proof_lower_bound_partial}

\proof{}
We omit the conditioning on $x$ part for simplicity. 
 Assumption \ref{ass: log-concavity} implies 
\begin{align}
    \log(\bar F(\theta x + (1-\theta))y) \geq \theta \log(\bar F(x)) + (1-\theta)\log(\bar F(y))
\end{align}
Let $\theta = \frac{z-z_2}{z_1-z_2}$, then we have 
\begin{align}
    \log(\Bar F_V(z)) \geq \frac{z-z_2}{z_1-z_2}\log(\bar F(z_1 \mid x)) + \frac{z_1-z}{z_1-z_2}\log(\bar F(z_2 \mid x)).
\end{align}

The proof is complete by utilizing the monotonicity property. 
\endproof

\subsection{Proof of Theorem \ref{thm:partial_lower}}
\label{sec:proof_partial_lower}

\begin{proof}
    
Here we use $\theta$ to represent $S$ in the main paper. 
Define
\begin{align}
    \tilde {\theta}_{-} = \frac{1}{n} \sum_{i=1}^n \int_{z=0}^{\infty} 
     \Big\{\mathbb{I}[\pi_D(z|X)\neq 0]\mu(z)F^\pi(z|X)  + \mathbb{I}[\pi_D(z|X)= 0]F_l(z,z_1,z_2,X) F^\pi(z|X)\Big\}dz
\end{align}
\begin{align}
        \mathbb{E}(\widehat{\theta}_{-} - {\theta}_{-}^*)^2 = \mathbb{E}(\widehat{\theta}_{-} - \tilde{\theta}_{-} + \tilde{\theta}_{-} - {\theta}_{-}^*)^2 \leq \mathbb{E}\underbrace{2(\widehat{\theta}_{-} - \tilde{\theta}_{-})^2}_{\text{(i)}} + \mathbb{E}\underbrace{2(\tilde{\theta}_{-} - {\theta}_{-}^*)^2}_{\text{(ii)}}
\end{align}

We abbreviate $F_l(z,z_1,z_2, X)$ as $F_l$, $\widehat{F}_l( z,z_1(z),z_2(z),X)$ as $\widehat{F}_l^o$, and $\widehat{F}_l(z,z_1,z_2,X)$ as $\widehat{F}_l$. $z_{1,2}(z)$ is the empirical observation and $z_{1,2}$ is the closet point in the population. 
\begin{align}
    \text{(i)} &\leq \frac{2}{n} \sum_{i=1}^n \mathbb{E} \Big( \int_{z=0}^{\infty} \mathbb{I}[\pi_D(z|X)\neq 0](\mu(z)-\widehat \mu(z))F^\pi(z|X)  + \mathbb{I}[\pi_D(z|X)= 0](F_l-\widehat{F}_l^o)F^\pi(z|X)\Big)^2 \\ 
    &\leq \frac{4}{n}\sum_{i=1}^n  \mathbb{E}\Big( \int_{z=0}^{\infty} \mathbb{I}[\pi_D(z|X)\neq 0](\mu(z)-\widehat \mu(z))F^\pi(z|X)\Big)^2 \\
    & \qquad \qquad + \frac{4}{n} \sum_{i=1}^n \mathbb{E}\Big( \int_{z=0}^{\infty}\mathbb{I}[\pi_D(z|X)= 0](F_l-\widehat{F}_l^o)F^\pi(z|X)\Big)^2 \nonumber 
\end{align}

The first and second inequalities are from Cauchy-Schwartz inequality. 
Denote the first term as (iii) and the second term as (iv). 
\begin{align}
    \text{(iii)} & \leq \frac{4}{n} \sum_{i=1}^n  \mathbb{E}\Big( \int_{z=0}^{P_\text{max}^D} (\mu(z)-\widehat \mu(z))^2\Big) \int_{z=0}^{P_\text{max}^D}(\mathbb{I}[\pi_D(z|X)\neq 0]F^\pi(z|X))^2 \\ 
    & \leq 4P_\text{max}^D P_\text{max}^D \sup_z \mathbb{E}(\Bar F_V - \widehat \mu)^2 = C_1\epsilon_n
\end{align}

We write $\sup_z \mathbb{E}(\Bar F_V - \widehat{\Bar F}_V)^2 = \epsilon_n$ and $C_1 = 4P_\text{max}^D P_\text{max}^D$. 
\begin{align}
    \text{(iv)} & =  \frac{4}{n} \sum_{i=1}^n \mathbb{E}\Big( \int_{z=0}^{\infty}\mathbb{I}[\pi_D(z|X)= 0](\Bar F_l-{\Bar F}_l^o + {\Bar F}_l^o - \widehat{\Bar F}_l^o)F^\pi(z|X)\Big)^2 \\ 
    & \leq \frac{8}{n} \sum_{i=1}^n\mathbb{E}\Big( \int_{z=0}^{\infty}\mathbb{I}[\pi_D(z|X)= 0](\Bar F_l-{\Bar F}_l^o)F^\pi(z|X)\Big)^2 
    \\
    & \qquad \qquad  + \frac{8}{n} \sum_{i=1}^n\mathbb{E}\Big( \int_{z=0}^{\infty}\mathbb{I}[\pi_D(z|X)= 0]({\Bar F}_l^o - \widehat{\Bar F}_l^o)F^\pi(z|X)\Big)^2 \nonumber 
\end{align}

Denote the first term as (v) and the second term as (vi). 
\begin{align}
    \text{(v)} & \leq \frac{8}{n}\sum_{i=1}^n\mathbb{E}\Big( \int_{z=0}^{P_\text{max}^D}({\Bar F_l} - {\Bar F}_l^o)\Big)^2 \Big( \int_{z=0}^{P_\text{max}^D}(\mathbb{I}[\pi_D(z|X)= 0]F^\pi(z|X))^2\Big) \\
    & \leq {8P_\text{max}^D}^2 \mathbb{E} \sup_{z}(({\Bar F_l} - {\Bar F}_l^o))^2 = 2C_1 \xi_n
\end{align}

Assume $\mathbb{E} \sup_{z}(({\Bar F_l} - {\Bar F}_l^o))^2\leq \xi_n$, 
\begin{align}
    \text{(vi)} & \leq \frac{8}{n} \sum_{i=1}^n\mathbb{E}\Big( \int_{z=0}^{P_\text{max}^D}({\Bar F_l}^o - \widehat{\Bar F}_l^o)\Big)^2 \Big( \int_{z=0}^{P_\text{max}^D}(\mathbb{I}[\pi_D(z|X)= 0]F^\pi(z|X))^2\Big) \\
    & \leq {8P_\text{max}^D}^2 \sup_z \mathbb{E}({\Bar F}_l^o - \widehat{\Bar F}_l^o)^2 = 2C_1\delta_n.
\end{align}

We write $\sup_z \mathbb{E}(\Bar F_l^0 - \widehat{\Bar F}_l^o)^2 = \delta_n$.
\begin{align}
    \text{(ii)} = \frac{1}{n} 2\text{Var}\left(\int_{z=0}^{\infty} 
     \Big\{\mathbb{I}[\pi_D(z|X)\neq 0]\mu(z)F^\pi(z|X)  + \mathbb{I}[\pi_D(z|X)= 0]F_l(\Bar{F}_V,z_1(z),z_2(z))F^\pi(z|X)\Big\}dz\right)
\end{align}

By Popoviciu's inequality, we have 
\begin{align}
    \text{(ii)} \leq \frac{2}{n}.
\end{align}

Then 
\begin{align}
    \mathbb{E}(\widehat{\theta}_{-} - {\theta}_{-}^*)^2 \leq 2C_1(\epsilon_n + \xi_n + \delta_n) + \frac{2}{n}.
\end{align}

Given the assumptions, we have 
\begin{align*}
    |\widehat{\theta}_{-} - {\theta}_{-}^*| =  O_p\left( n^{-\alpha_1} + n^{-\alpha_2} + n^{-\alpha_3} + n^{-1/2} \right) = O_p(n^{-\min \{\alpha_1,\alpha_2,\alpha_3,1/2\}})
\end{align*}

The upper bound's proof can be constructed similarly. 
\end{proof}

\section{Implementation of the Personalized Pricing Policy}
\label{app:personalized_policy}

The personalized pricing policy is implemented as follows.
We first fit a demand function $\widehat{d}(x,p) = \widehat{Pr}(Y=1|x,p)$, then calculate the estimated reward as $\widehat \mu(x,p) = p\widehat{d}(x,p)$. The price is selected using $P \sim \text{softmax}_{p} \gamma \widehat \mu(x,p)$, where $\gamma$ is the temperature. In all experiments, we set $r=1$. When $r\rightarrow\infty$, this will correspond to a myopic personalized pricing policy that maximizes reward based on the current best demand estimation.

\end{document}